\documentclass{article}

\PassOptionsToPackage{numbers, compress}{natbib}

    \usepackage[final]{neurips_2025}

\usepackage[utf8]{inputenc} 
\usepackage[T1]{fontenc}    
\usepackage{hyperref}       
\usepackage{url}            
\usepackage{booktabs}       
\usepackage{amsfonts}       
\usepackage{multirow}       
\usepackage{graphicx}       
\usepackage{nicefrac}       
\usepackage{microtype}      
\usepackage{xcolor}         
\usepackage{amsmath}
\usepackage{algorithm}
\usepackage{algpseudocode}
\usepackage{threeparttable}

\newcommand{\vs}{\textit{vs.~}}
\newcommand{\modelname}{AsymGS}

\title{Robust Neural Rendering in the Wild with Asymmetric Dual 3D Gaussian Splatting}

\author{%
Chengqi Li$^{1}$ \quad Zhihao Shi$^{1}$ \quad Yangdi Lu$^1$ \quad Wenbo He$^1$ \quad Xiangyu Xu$^2$\thanks{Corresponding author} \\
$^1$McMaster University \quad $^2$Xi'an Jiaotong University\\
}

\begin{document}
\maketitle

\begin{abstract}
3D reconstruction from in-the-wild images remains a challenging task due to inconsistent lighting conditions and transient distractors. Existing methods typically rely on heuristic strategies to handle the low-quality training data, which often struggle to produce stable and consistent reconstructions, frequently resulting in visual artifacts.
%
In this work, we propose \modelname{}, a novel framework that leverages the stochastic nature of these artifacts: they tend to vary across different training runs due to minor randomness. 
Specifically, our method trains two 3D Gaussian Splatting (3DGS) models in parallel, enforcing a consistency constraint that encourages convergence on reliable scene geometry while suppressing inconsistent artifacts.
To prevent the two models from collapsing into similar failure modes due to confirmation bias, we introduce a divergent masking strategy that applies two complementary masks: a multi-cue adaptive mask and a self-supervised soft mask, which leads to an asymmetric training process of the two models, reducing shared error modes.
In addition, to improve the efficiency of model training, we introduce a lightweight variant called Dynamic EMA Proxy, which replaces one of the two models with a dynamically updated Exponential Moving Average (EMA) proxy, and employs an alternating masking strategy to preserve divergence.
Extensive experiments on challenging real-world datasets demonstrate that our method consistently outperforms existing approaches while achieving high efficiency. See the project website at \url{https://steveli88.github.io/AsymGS}.
\end{abstract}

\section{Introduction}


3D scene reconstruction from multiple views is a fundamental problem in computer vision.
Recent advances such as Neural Radiance Fields (NeRF)~\cite{vedaldi_nerf_2020} and 3D Gaussian Splatting (3DGS)~\cite{kerbl_3d_2023} have achieved impressive rendering quality by learning volumetric or point-based scene representations from posed images.
However, these methods typically assume that training images exhibit consistent illumination and minimal occlusion, which are rarely satisfied in real-world settings.

In-the-wild images are often captured under varying lighting conditions and contain transient distractors such as pedestrians or vehicles; 
these factors introduce substantial noise into the supervision signal, leading to degraded reconstruction quality and visual artifacts.
While several recent works have attempted to address these challenges~\cite{martin-brualla_nerfw_2021,sabour_robustnerf_2023,ren_nerfonthego_2024,zhang_gsw_2024,kulhanek_nerfbaselines_2024,sabour_spotlesssplats_2024,leonardis_swag_2025,lin_hybridgs_2025}, they largely rely on heuristic strategies to suppress the effects of corrupted supervision from low-quality training data, such as per-image appearance embeddings that are only weakly or indirectly supervised through photometric losses~\cite{martin-brualla_nerfw_2021,zhang_gsw_2024}, or hand-crafted rules to filter outlier training signals~\cite{sabour_robustnerf_2023}.
As a result, such approaches often lack stability and generalizability, which leads to artifact-prone reconstructions.

\begin{figure}[t]
\vspace{-8mm}
\centering
\begin{minipage}[ht]{0.63\linewidth}
    \centering
    {\includegraphics[width=\linewidth]{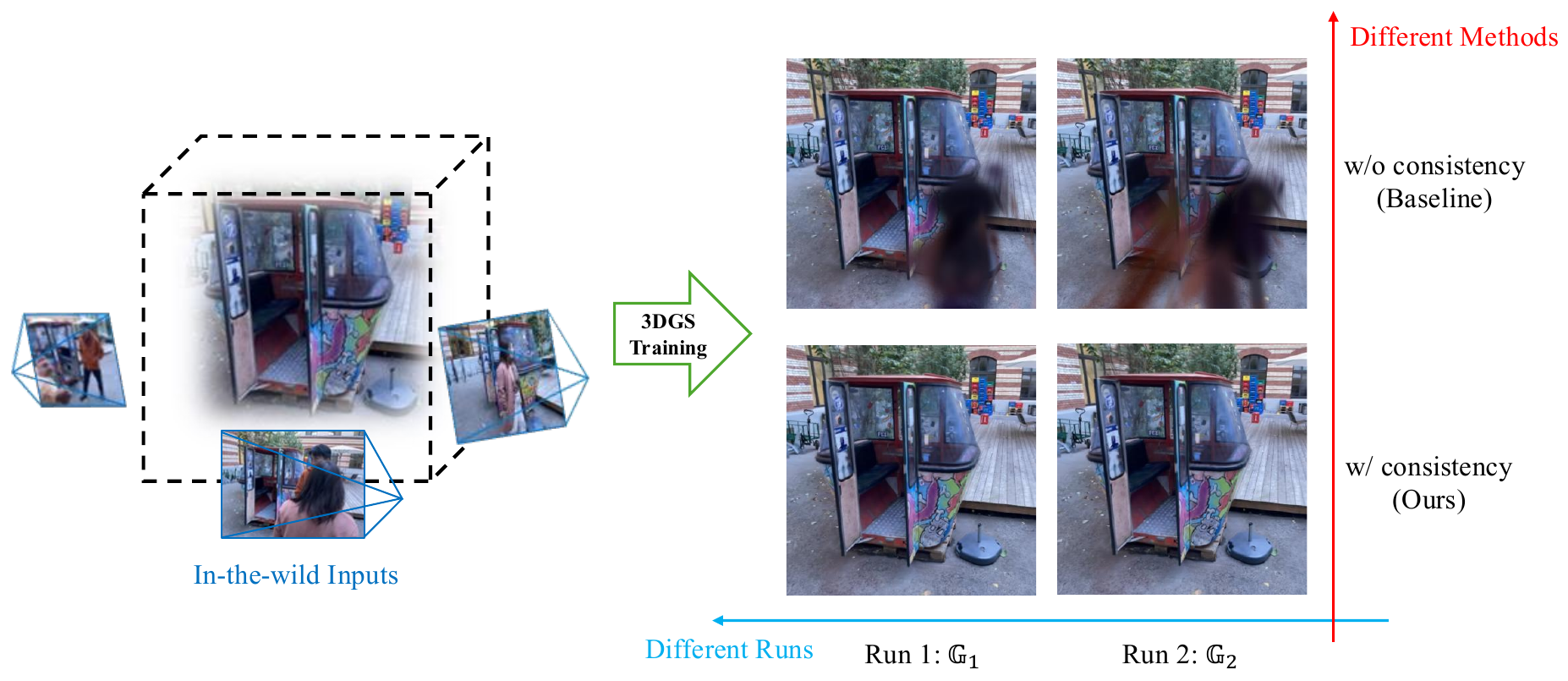}}
\end{minipage}
\hfill
\begin{minipage}[ht]{0.33\linewidth}
    \centering
    {\includegraphics[width=\linewidth]{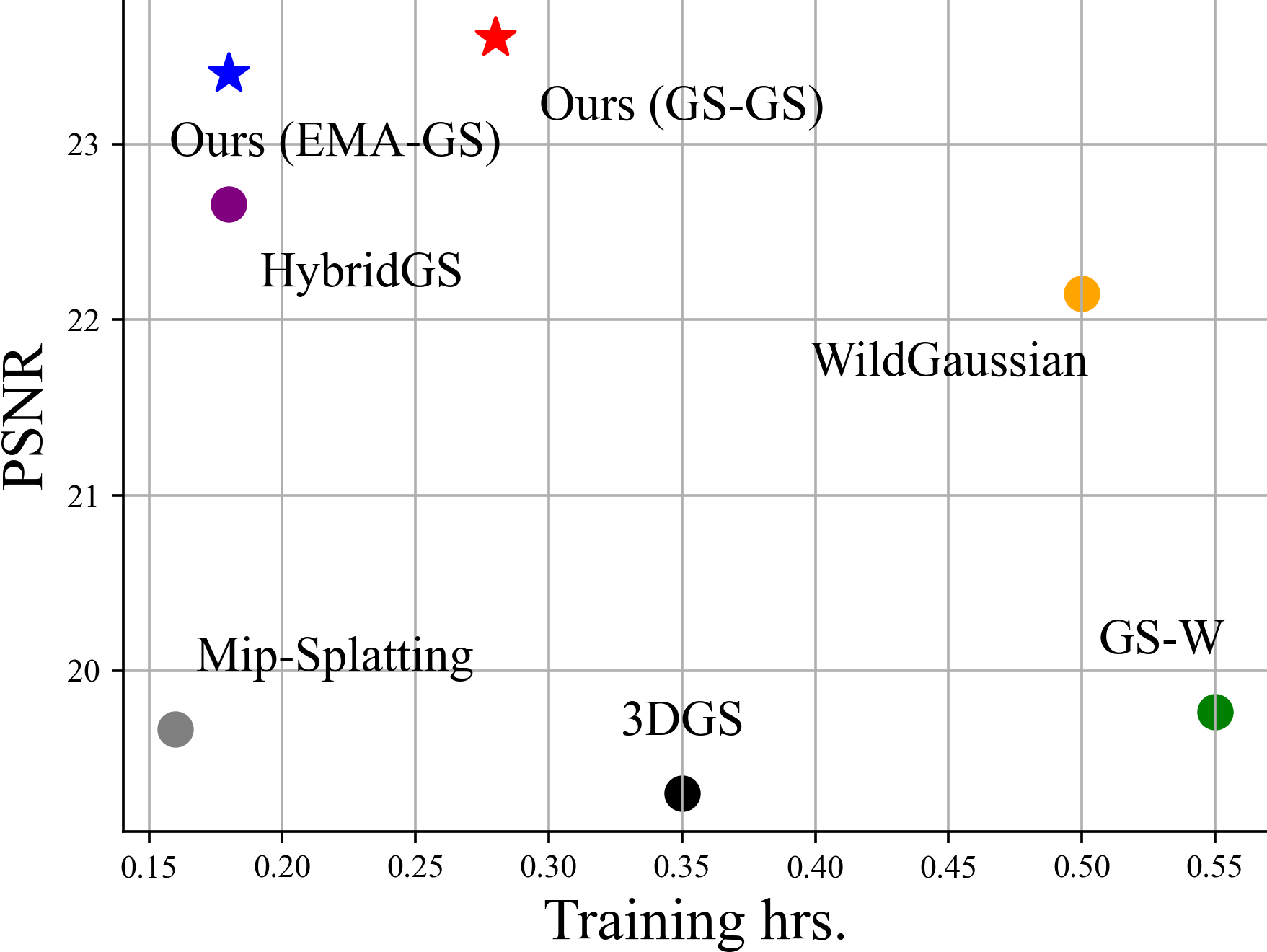}}
\end{minipage}
\vspace*{-2mm}
\caption{\textbf{Left:} The key insight of this work is that artifacts arising from low-quality in-the-wild inputs are typically stochastic across different runs of the same model (Baseline: Run 1 \textit{vs.} Run 2). This motivates the design of the Asymmetric Dual 3DGS framework, which enhances true scene structure  while suppressing errors through cross-model consistency (w/ consistency). \textbf{Right:} Our method compares favorably against the state-of-the-art approaches in terms of reconstruction quality while maintaining high training efficiency. Results are on the NeRF On-the-go dataset~\cite{ren_nerfonthego_2024}.}
\label{fig:figure1}
\vspace{-5mm}
\end{figure}

To bridge this gap, we propose a new framework, called \modelname{}, which is motivated by a key empirical observation: \emph{artifacts arising from low-quality in-the-wild training data are typically stochastic}. In other words, the artifacts vary randomly across different runs of the same model with only minor training perturbations, such as data order shuffling (see Figure~\ref{fig:figure1}-Left).
This suggests that enforcing consistency between independently trained 3DGS can help suppress unreliable or spurious signals in the in-the-wild training images.
To this end, we introduce a dual-model architecture where two 3DGS models are trained concurrently with a consistency constraint, following the intuition that true scene structure should be consistently reconstructed across different model runs, while the artifacts induced by low-quality data tend to diverge.

Nevertheless, naively training two models in parallel can lead to confirmation bias, where both models reinforce the same errors.
To encourage more divergent error modes and mitigate confirmation bias, we introduce a divergent masking strategy: applying distinct masks to each model that emphasize complementary factors for filtering out transient or distracting content.
One mask is learned in a self-supervised manner based on feature-level similarity between predicted and ground truth images, while the other, called multi-cue adaptive mask, uses stereo-based correspondence to identify likely distracting regions. These complementary filtering schemes encourage the two 3DGS models to focus on different static aspects of the scene.
Consequently, this asymmetric strategy leads to divergent and complementary optimization paths and reduces shared error modes.
The final reconstruction is then guided by the agreement between the two models, which reliably captures consistent and accurate scene structures while suppressing artifacts.

While the dual-model framework effectively improves the reconstruction quality, it introduces notable computational overhead in the training process. 
To mitigate this, we introduce a lightweight variant, called Dynamic EMA Proxy, which replaces the second 3DGS model with a dynamic, training-free Exponential Moving Average (EMA) copy of the primary model.
Unlike standard EMA~\cite{he_momentum_2020}, our Dynamic EMA proxy is specifically designed to track the evolving nature of 3DGS representations, accounting for Gaussian densification and pruning.
Since only one model is actively trained in this setup, which no longer allows independent masks for two models, we additionally design an alternating masking strategy that alternates between the two masks, maintaining divergent training signals and mitigating confirmation bias.

Our contributions are as follows: 
1) We propose a \modelname{} framework for in-the-wild 3D scene reconstruction. By enforcing consistency constraints between two 3DGS models with complementary masks, our framework significantly improves the robustness and accuracy of scene representations. 
2) We develop a divergent masking strategy by introducing different masking mechanisms for each 3DGS model, which handle various types of distractors and promote divergent optimization paths to mitigate confirmation bias. 3) To address the computational overhead of the dual-model framework, we introduce a Dynamic EMA Proxy, coupled with an alternating masking strategy, which effectively improves training efficiency. 4) We conduct extensive evaluations across a diverse set of in-the-wild 3D scene reconstruction datasets, demonstrating that our method consistently achieves state-of-the-art performance and efficiency, highlighting its robustness and generality.


\section{Related work}
\noindent \textbf{3D Scene Reconstruction.} Neural Radiance Fields (NeRF) \cite{vedaldi_nerf_2020} revolutionizes photorealistic novel view synthesis by modeling scenes as continuous functions that map 3D coordinates to color and density. More recently, 3D Gaussian Splatting (3DGS) \cite{kerbl_3d_2023, yu_mip-splatting_2024} has gained attention as a real-time alternative, representing scenes with optimizable Gaussian primitives. While effective, both methods assume static scenes to enforce multi-view consistency, an assumption often violated in in-the-wild settings due to varying illumination and transient objects, limiting their practical applicability.

\noindent \textbf{3D Scene Reconstruction in the Wild.} NeRF-W~\cite{martin-brualla_nerfw_2021} first addressed 3D scene reconstruction in the wild with a modular architecture combining a learnable appearance embedding and an uncertainty map to suppress distractors—an approach that has since become standard. NeRF On-the-go~\cite{ren_nerfonthego_2024} builds on this by using DINOv2 feature \cite{oquab_dinov2_2023} residuals to construct uncertainty maps. RobustNeRF~\cite{sabour_robustnerf_2023} tackles noisy training images through a robust loss function. 
GS-W~\cite{zhang_gsw_2024}, Wild-GS \cite{xu_wild-gs_2024} and WildGaussian~\cite{kulhanek_wildgaussians_2024} extend learnable embeddings with both global and per-Gaussian local embeddings for fine-grained appearance modeling. SpotlessSplats~\cite{sabour_spotlesssplats_2024} introduces a learnable mask based on thresholded and dilated residuals, while SWAG~\cite{leonardis_swag_2025} adds a view-dependent opacity term per Gaussian to identify transient distractors. HybridGS~\cite{lin_hybridgs_2025} employs a dual-model setup (3DGS for static content and 2DGS~\cite{Huang_2dgs_2024} for dynamic distractors) to learn an accurate uncertainty map iteratively.
Despite their contributions, these methods largely rely on heuristic strategies to suppress corrupted supervision signals from low-quality training data. 
For instance, the per-image appearance embeddings are only weakly or indirectly supervised through photometric losses~\cite{martin-brualla_nerfw_2021,zhang_gsw_2024}, and outlier filtering is often governed by hand-crafted rules~\cite{sabour_robustnerf_2023}. As a result, these approaches frequently suffer from instability and produce reconstructions with noticeable artifacts.
In this work, we propose a principled framework, \modelname{}, 
to reduce such artifacts and improve reconstruction stability, based on the key observation that many artifacts exhibit stochastic behavior and are not consistent across different training runs. Our method explicitly exploits this property to enhance robustness and achieve high-fidelity reconstructions.
\begin{figure}[t]
\centering
\resizebox{0.9\linewidth}{!}{
\includegraphics[width=\linewidth]{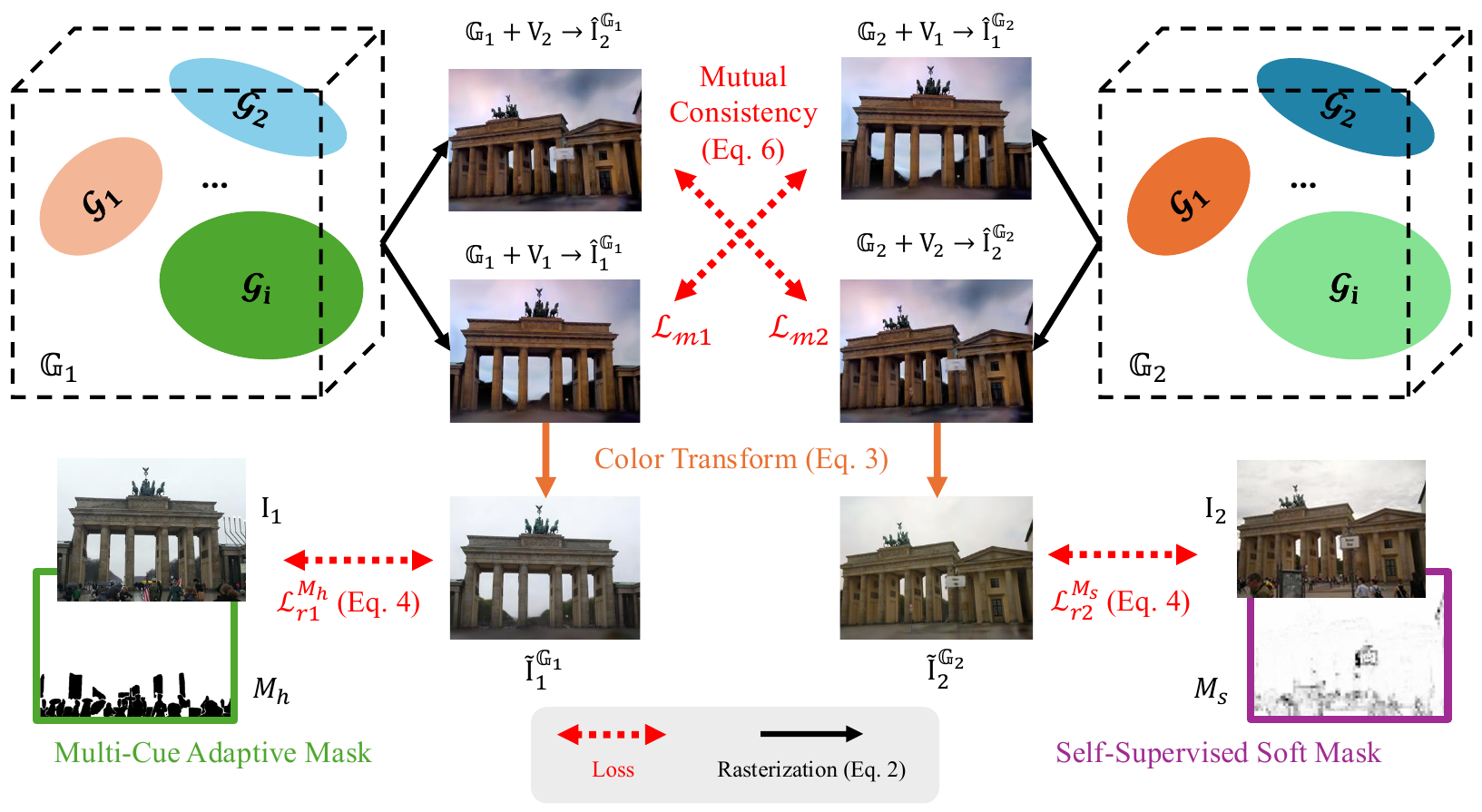}
}
\vspace*{-2mm}
\caption{Overview of the \modelname{} framework.
Two 3DGS models $\mathbb{G}_1$ and $\mathbb{G}_2$ are concurrently optimized with the reconstruction loss $\mathcal{L}_{r1}^{\mathbf{M}_h}$ and $\mathcal{L}_{r2}^{\mathbf{M}_s}$ (Eq.~\ref{eq:dualrecon1}), along with the mutual consistency loss $\mathcal{L}_{m1}$ and $\mathcal{L}_{m2}$ (Eq.~\ref{eq:dualmutual}). In addition, we apply a mask loss (Eq.~\ref{eq:mask}) for learning soft mask in a self-supervised manner.
For improved efficiency, we also propose an EMA version of our framework by replacing $\mathbb{G}_2$ with a dynamic EMA proxy.
Both the mask loss and the EMA proxy have been omitted here for clarity.
Note that the color transform in this figure is for illustration purpose, which undergoes a rasterization process in practice as introduced in Section~\ref{sec:preliminary}.
}
\label{fig:overview}
\vspace{-4mm}
\end{figure}

\section{Method}

An overview of the proposed algorithm is shown in Figure~\ref{fig:overview}. Please refer to the caption for details.


\subsection{Preliminaries}\label{sec:preliminary}
3D Gaussian Splatting (3DGS) \cite{kerbl_3d_2023} represents a 3D scene as a set of \(N\) 3D anisotropic Gaussians \(\mathbb{G} = \{\mathcal{G}_i\}_{i=1}^N\). Each Gaussian \(\mathcal{G}_i\) is parameterized by a centroid \(\mathbf{X}_i\),  a covariance matrix \(\mathbf{\Sigma}_i\), an opacity \(\alpha_i\), and a set of spherical harmonic (SH) coefficients \(\theta_i\) for view-dependent color representation. For rendering with a viewing camera \(\mathbf{V}_j\), we project both centroids of Gaussians and the covariance matrix onto the 2D image plane as \(\mathbf{x}_{ij}\) and \(\mathbf{\Sigma}_{ij}\), respectively. 
The projected opacity \(\alpha_{ij}\), which is a function of 2D image plane coordinate \(\mathbf{y}\), can then be defined as below:
\begin{equation}
\alpha_{ij}(\mathbf{y}) = \alpha_{i} \cdot \exp{\left[-\frac{1}{2} (\mathbf{y} - \mathbf{x}_{ij})^T {\mathbf{\Sigma}_{ij}}^{-1} (\mathbf{y} - \mathbf{x}_{ij})\right]} 
\label{eq:projopacity}  
\end{equation}
The color \(\mathbf{C}\) of a pixel located at \(\mathbf{y}\) for camera \(\mathbf{V}_j\) is computed using the \(\alpha\)-blending with the following formula:
\begin{gather}
\mathbf{C}(\mathbf{y},{\mathbf{V}_j}, \mathbb{G}) = \sum_{i=1}^{N} \mathbf{c}_{ij} \alpha_{ij}(\mathbf{y}) \prod_{k=1}^{i-1} \left(1 - \alpha_{kj}(\mathbf{y})\right), \quad \mathbf{c}_{ij} = \mathrm{SH} \left(\mathbf{r}_{ij}, \theta_i\right)  
\label{eq:blending}    
\end{gather}
where $\mathbf{r}_{ij}$ is the ray direction from the \(j\)-th camera center to the $i$-th Gaussian centroid, and $\mathbf{c}_{ij}$ is the corresponding color of the observed Gaussian primitive obtained using spherical harmonic (SH) function. 
Performing Eq.~\ref{eq:blending} for every pixel on the image plane constitutes the rasterization process, which results in the rendered image $\hat{\mathbf{I}}^{\mathbb{G}}_j = \mathrm{Rasterize}(\mathbb{G}, \mathbf{V}_j)$.

\noindent\textbf{View-dependent appearance modeling.}
To address appearance variations in in-the-wild data, we follow the approach of WildGaussian~\cite{kulhanek_wildgaussians_2024} to adaptively adjust the observed color of the Gaussian primitives to account for the view-dependent factors, such as the varying illumination across images captured at different times of day. 

This adjustment is conditioned on both the per-Gaussian appearance embedding \(\mathbf{p}_i\) and the per-view appearance embedding \(\mathbf{q}_j\). Specifically, \(\mathbf{p}_i\), \(\mathbf{q}_j\), and \(\mathbf{c}_{ij}\) are sent into an MLP \(f\) to predict affine transformation parameters:
\begin{equation}\label{eq:appearance modeling}
(a, b) = f(\mathbf{p}_i, \mathbf{q}_j, \mathbf{c}_{ij}), \quad \tilde{\mathbf{c}}_{ij} = a \cdot \mathbf{c}_{ij} + b,
\end{equation}
where \(a\) and \(b\) are three-dimensional outputs corresponding to RGB channels. The transformed color \(\tilde{\mathbf{c}}_{ij}\) is then used to replace \(\mathbf{c}_{ij}\) in the blending process described by Eq.~\ref{eq:blending}, and the resulting view-dependent image is denoted as $\tilde{\mathbf{I}}^{\mathbb{G}}_j = \mathrm{Rasterize}_\mathrm{dep}(\mathbb{G}, \mathbf{V}_j)$. During training, \(\mathbf{p}_i\), \(\mathbf{q}_j\), \(f\), and 3DGS parameters are jointly optimized. 


\subsection{Dual 3DGS}
A central insight of this work is that artifacts arising from in-the-wild training data are typically stochastic in nature. 
When two 3DGS models are trained on the same scene but with different view sampling orders, their static scene representations remain consistent, whereas their renderings could diverge in regions affected by outliers. An example is shown in Figure~\ref{fig:figure1}-Left.

Motivated by this observation, we introduce a framework with two 3DGS, where each model is trained with a different sampling order, and a consistency constraint is enforced between their renderings.  
Specifically, we maintain two sets of Gaussians, \(\mathbb{G}_1\) and \(\mathbb{G}_2\), to represent the same scene. In each training iteration, we independently sample two views from separate training view lists, yielding two viewing cameras \(\mathbf{V}_1\) and \(\mathbf{V}_2\), along with their corresponding ground-truth images, \(\mathbf{I}_1\) and \(\mathbf{I}_2\).

Similar to 3DGS~\cite{kerbl_3d_2023}, we train \(\mathbb{G}_1\) and \(\mathbb{G}_2\) with reconstruction objectives defined as:
\begin{gather}
\mathcal{L}_{r1}^{\mathbf{M}} = \mathcal{L}_\text{recon}(\tilde{\mathbf{I}}^{\mathbb{G}_1}_1, \mathbf{I}_1,{\mathbf{M}}), \quad
\mathcal{L}_{r2}^{\mathbf{M}} = \mathcal{L}_\text{recon}(\tilde{\mathbf{I}}^{\mathbb{G}_2}_2, \mathbf{I}_2,{\mathbf{M}}), \label{eq:dualrecon1}\\
\mathcal{L}_\text{recon}(\tilde{\mathbf{I}}, \mathbf{I}, {\mathbf{M}}) = \lambda \cdot \mathrm{DSSIM}(\mathbf{M} \odot \tilde{\mathbf{I}}, \mathbf{M} \odot \mathbf{I}) + (1 - \lambda) \cdot \|\mathbf{M} \odot \tilde{\mathbf{I}} - \mathbf{M} \odot \mathbf{I}\|_1,
\label{eq:dualrecon}
\end{gather}
where \(\tilde{\mathbf{I}}^{\mathbb{G}_n}_j = \mathrm{Rasterize}_\mathrm{dep} (\mathbb{G}_n,\mathbf{V}_j) \) is the rendered image for the $n$-th 3DGS model from viewpoint $\mathbf{V}_j$ using Eq.~\ref{eq:blending} and \ref{eq:appearance modeling}.
${\mathbf{M}}$ is a spatial mask to filter out transient distracting regions, such as pedestrians or moving vehicles, which will be detailed in Section~\ref{sec:asymmetric}.
\(\odot\) denotes element-wise multiplication.
DSSIM represents the structural dissimilarity index measure~\cite{wang2004image}.
$\lambda$ is a hyperparameter to balance the DSSIM and the $L_1$ terms.


\noindent\textbf{Mutual consistency.} 
Since \(\mathbb{G}_1\) and \(\mathbb{G}_2\) represent the same underlying scene, their renderings from the same camera viewpoint should stay close. This motivates us to define a mutual consistency regularization as:
\begin{gather}
\mathcal{L}_{m1} = \|\hat{\mathbf{I}}^{\mathbb{G}_2}_1 - \hat{\mathbf{I}}^{\mathbb{G}_1}_1\|_1, \quad
\mathcal{L}_{m2} = \|\hat{\mathbf{I}}^{\mathbb{G}_1}_2 - \hat{\mathbf{I}}^{\mathbb{G}_2}_2\|_1
\label{eq:dualmutual}
\end{gather}
where \(\hat{\mathbf{I}}^{\mathbb{G}_n}_j = \mathrm{Rasterize}(\mathbb{G}_n, \mathbf{V}_j)\) is the view-dependent rendering obtained via Eq.~\ref{eq:blending}.
We emphasize that this consistency constraint is performed over $\hat{\mathbf{I}}^{\mathbb{G}_n}_j$ instead of $\tilde{\mathbf{I}}^{\mathbb{G}_n}_j$, because $\hat{\mathbf{I}}^{\mathbb{G}_n}_j$ captures the intrinsic appearance of the 3D scene, whereas $\tilde{\mathbf{I}}^{\mathbb{G}_n}_j$ is affected by dynamic lighting. 
This strategy provides a principled way to preserve static structures while suppressing spurious signals, which enables more robust and reliable reconstruction.

Note that we only use the \(L_1\) loss for consistency regularization in Eq.~\ref{eq:dualmutual} as incorporating the DSSIM loss adversely affects performance in our experiments.
Furthermore, we empirically find that incorporating consistency regularization too early in training can hinder convergence, as both models may still be dominated by noise and unstable geometry. 
To address this, we adopt a progressive strategy: we first allow the two models to be trained independently for a number of warm-up iterations, during which they develop their own estimates of the static scene. Once their reconstructions become sufficiently stable, we introduce the consistency loss to encourage convergence on shared, reliable structures.

%



\subsection{Asymmetric Dual 3DGS}\label{sec:asymmetric}
While the above framework offers improved consistency through mutual supervision, its symmetric design, where both 3DGS models are trained in the same manner using the reconstruction loss in Eq.~\ref{eq:dualrecon}, poses a risk of confirmation bias: both models may converge toward the same reconstruction errors due to their similar optimization signals.
%
%

To address this issue, we propose an Asymmetric Dual 3DGS variant, where each model is trained with a distinct masking strategy that emphasizes complementary criteria for filtering out transient or distracting content. 
This encourages divergent error patterns, enhances robustness, and mitigates confirmation bias. 
%
Specifically, we use a Multi-Cue Adaptive Mask and a Self-Supervised Soft Mask.

\begin{figure}[t]
\centering
\begin{minipage}[b]{0.15\linewidth}
    \centering
    {\includegraphics[width=\linewidth]{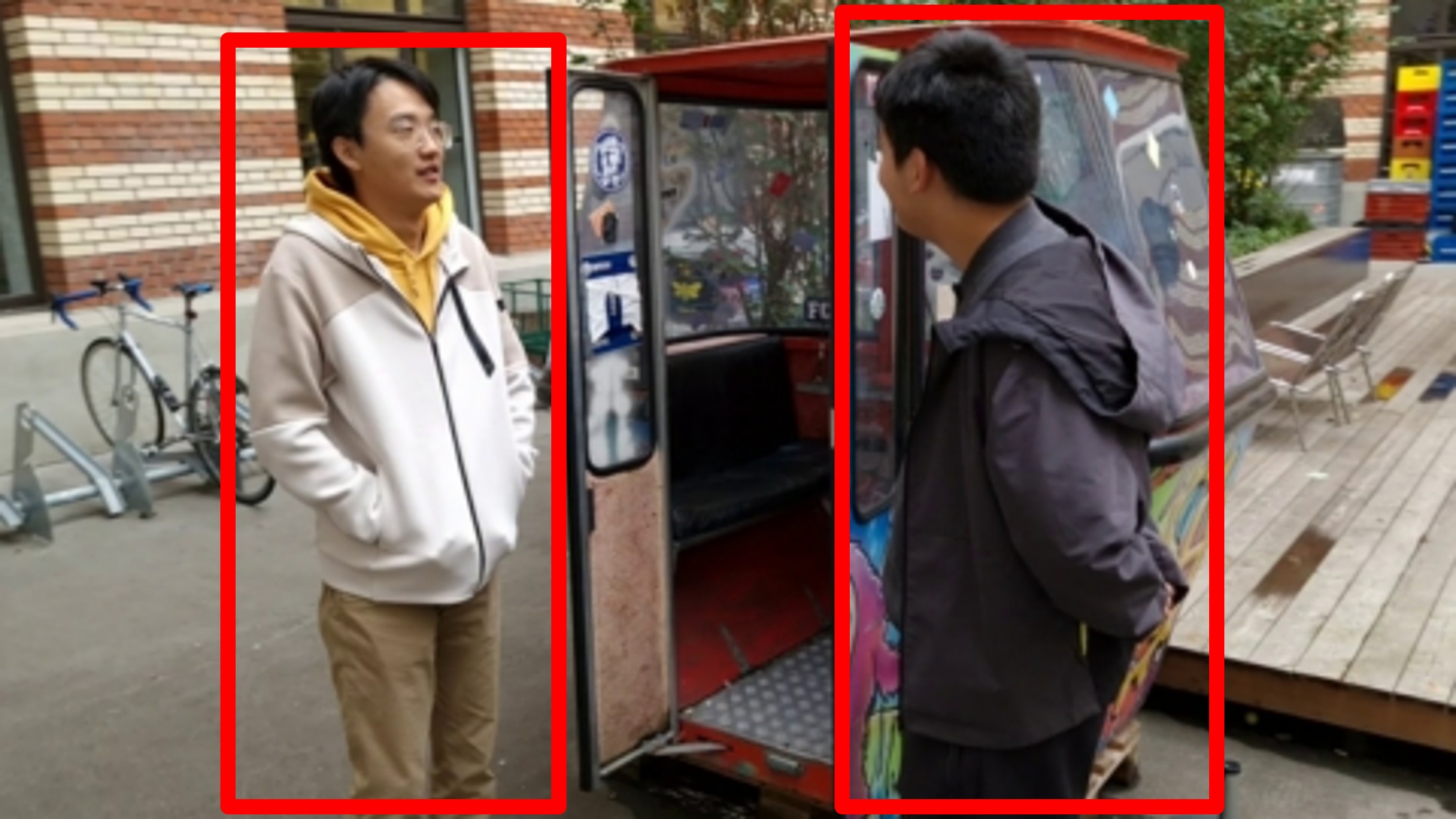}}
    {\includegraphics[width=\linewidth]{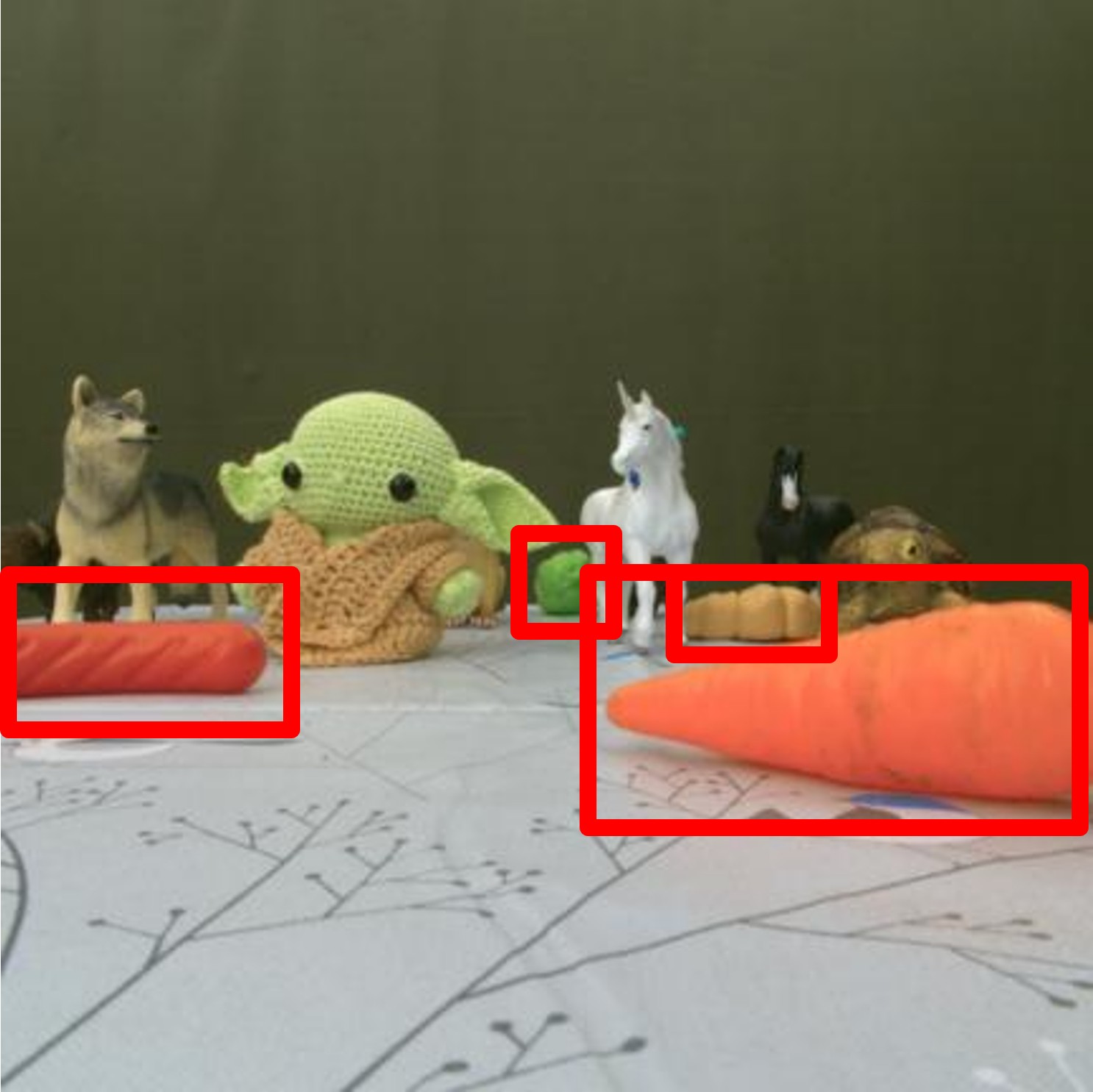}}
    \centerline{Input}
\end{minipage}
\hfill
\begin{minipage}[b]{0.15\linewidth}
    \centering
    {\includegraphics[width=\linewidth]{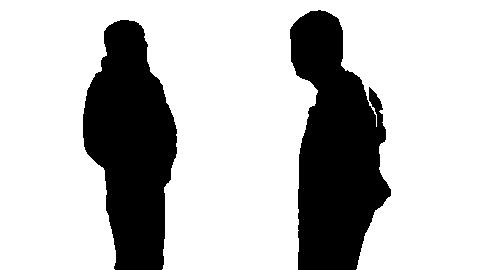}}
    {\includegraphics[width=\linewidth]{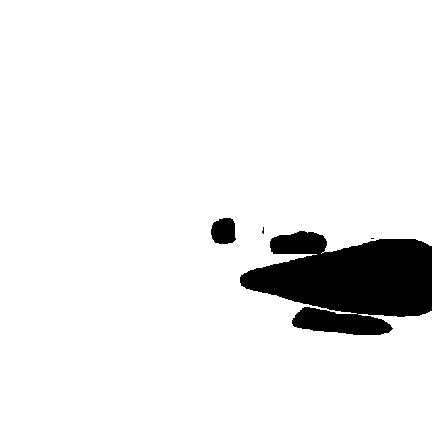}}
    \centerline{\(\mathbf{M}_h\)}
\end{minipage}
\hfill
\begin{minipage}[b]{0.15\linewidth}
    \centering
    {\includegraphics[width=\linewidth]{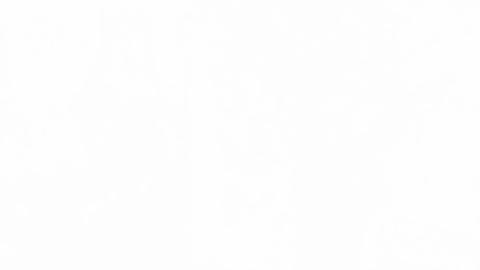}}
    {\includegraphics[width=\linewidth]{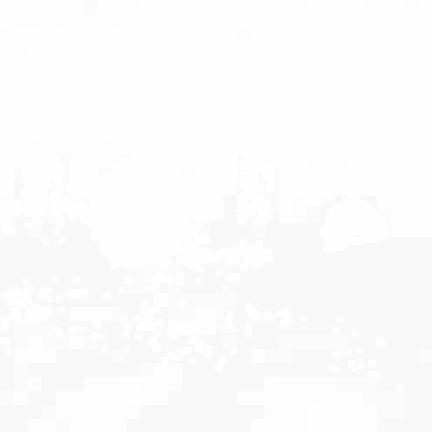}}
    \centerline{\(\mathbf{M}_s\) (Early)}
\end{minipage}
\hfill
\begin{minipage}[b]{0.15\linewidth}
    \centering
    {\includegraphics[width=\linewidth]{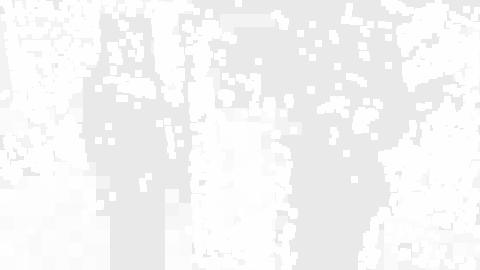}}
    {\includegraphics[width=\linewidth]{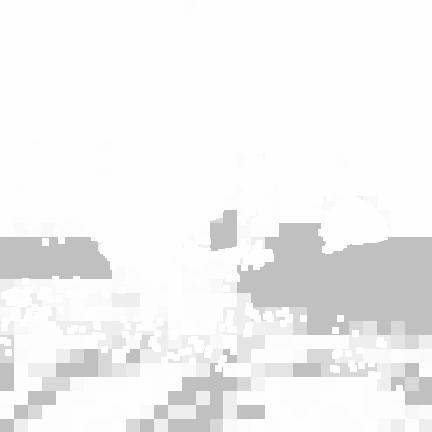}}
    \centerline{}
\end{minipage}
\hfill
\begin{minipage}[b]{0.15\linewidth}
    \centering
    {\includegraphics[width=\linewidth]{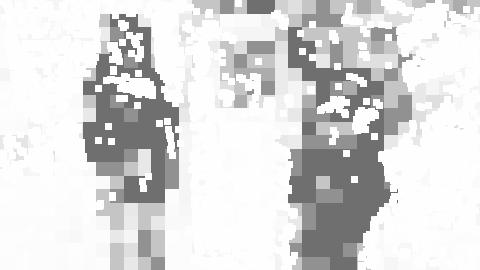}}
    {\includegraphics[width=\linewidth]{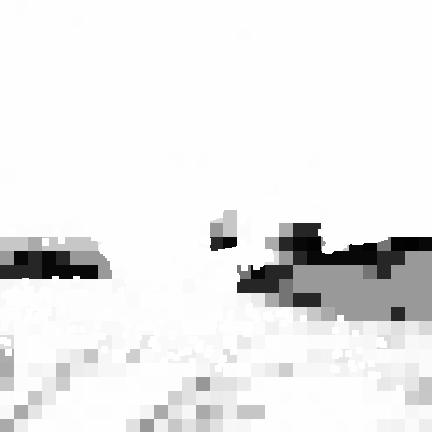}}
    \centerline{}
\end{minipage}
\hfill
\begin{minipage}[b]{0.15\linewidth}
    \centering
    {\includegraphics[width=\linewidth]{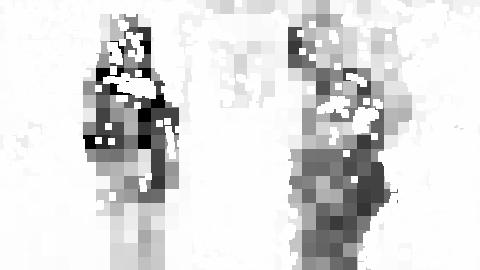}}
    {\includegraphics[width=\linewidth]{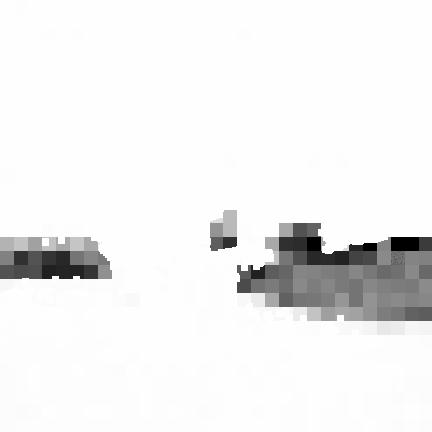}}
    \centerline{\(\mathbf{M}_s\) (Final)}
\end{minipage}
\hfill
\caption{Comparisons of hard and soft masks. Distractors are highlighted in red boxes in the input. The right four columns show the evolving of the self-supervised soft mask across different epochs. 
}
\label{fig:mask_compare}
\vspace{-4mm}
\end{figure}

\noindent\textbf{Multi-Cue Adaptive Mask (\(\mathbf{M}_h\)).} 
As illustrated in Figure~\ref{fig:mask_compare}, \(\mathbf{M}_h\) is a hard binary mask (1 indicates static regions and 0 indicates distractors) that identifies transient and distracting regions by integrating multiple cues, including semantic segmentation, stereo correspondence, pixel-level residuals, and feature-level residuals. 

We begin by applying the Segment Anything (SAM) model~\cite{Kirillov_sam_2023,li_segmentsam_2025} to partition each image into semantically coherent regions.
To detect static content, we perform multi-view stereo matching across the training images with COLMAP~\cite{schonberger2016structure}.
Semantic regions are considered static if they contain a sufficient number of valid correspondences (we empirically choose a threshold of 3 matches).
Among the remaining regions, we identify transient distractors by analyzing reconstruction residuals.
For each region, we compute pixel-level residuals, \textit{i.e.}, the $L_1$ error between the rendered and ground-truth images, and feature-level residuals, \textit{i.e.}, the cosine distance between DINOv2-encoded feature maps~\cite{oquab_dinov2_2023} of the rendered and ground-truth images.
Regions with above-average residuals in both metrics are classified as distractors and masked out during training.
This multi-cue approach offers higher robustness than prior methods that rely on single cues~\cite{sabour_robustnerf_2023, ren_nerfonthego_2024,martin-brualla_nerfw_2021,Rematas_urban_2022}, which generalizes more effectively across diverse in-the-wild scenes.
See Algorithm \ref{alg:MAM} in the supplementary material for full details of the Multi-Cue Adaptive Mask.

\noindent\textbf{Self-Supervised Soft Mask (\(\mathbf{M}_s\)).} 
To complement the rule-based hard mask \(\mathbf{M}_h\), we introduce a learnable soft mask \(\mathbf{M}_s\), whose values range between 0 and 1.
Unlike the static $\mathbf{M}_h$, this soft mask is optimized jointly with the model and adapts throughout training.
The objective for $\mathbf{M}_s$ is derived from the cosine similarity between DINOv2 feature maps of the ground-truth image $\mathbf{F}$ and the rendered image $\tilde{\mathbf{F}}$:
%
\begin{equation}\label{eq:mask}
\mathcal{L}_{\text{mask}} = \| \mathbf{M}_s - f_\text{interp}(\text{CosineSimilarity}(\mathbf{F}, \tilde{\mathbf{F}})) \|_1,
\end{equation}
where $f_\text{interp}$ denotes spatial interpolation to match the training image resolution. 
This formulation is self-supervised, requiring no ground-truth masks.
We initialize $\mathbf{M}_s$ as an all-one tensor, allowing the model to gradually refine the mask as training progresses (Figure~\ref{fig:mask_compare}).

As shown in Figure~\ref{fig:mask_compare}, the hard mask $\mathbf{M}_h$ is more definitive with clearer boundaries but can be overconfident, potentially missing certain transient objects.
In contrast, the soft mask $\mathbf{M}_s$ is often more sensitive to subtle distractors and better captures ambiguous regions, thus providing complementary information to $\mathbf{M}_h$.

By combining the loss terms in Eq.~\ref{eq:dualrecon}, \ref{eq:dualmutual}, \ref{eq:mask}, the overall objective for our \modelname{} can be written as:
\begin{equation}\label{eq:finalloss}
\mathcal{L} = \mathcal{L}_{r1}^{\mathbf{M}_h} + \mathcal{L}_{r2}^{\mathbf{M}_s} + \lambda_m (\mathcal{L}_{m1} + \mathcal{L}_{m2}) + \lambda_{\text{mask}} \mathcal{L}_{\text{mask}},
\end{equation}
where $\lambda_m$ and $\lambda_\text{mask}$ are hyperparameters for balancing loss terms.
Since $\mathbf{M}_s$ dynamically evolves during training and intrinsically differs from the fixed $\mathbf{M}_h$, training one model with $\mathcal{L}_{r1}^{\mathbf{M}_h}$ and the other with $\mathcal{L}_{r2}^{\mathbf{M}_s}$ in our \modelname{} framework introduces complementary inductive biases. This asymmetry promotes diverse learning dynamics, making it less likely for the two models to converge on the same reconstruction error, thereby reducing confirmation bias and enhancing overall robustness.


\subsection{Dynamic EMA proxy}\label{sec:dynamicEMA}

While the \modelname{} framework significantly improves robustness and reconstruction quality, it requires simultaneous training of two 3DGS models, which introduces considerable computational overhead and undermines the fast training advantage of 3DGS.
To address this issue, we propose a more efficient alternative by replacing one of the two models with a dynamic EMA proxy. Moreover, an alternating masking strategy is introduced to counteract confirmation bias. This design retains the benefits of dual-model regularization while significantly reducing computation cost.


Let \(\mathbb{G}_1\) denote the set of Gaussians actively optimized during training, and \(\mathbb{G}_{\text{EMA}}\) its EMA counterpart, updated at each training step by:
\begin{gather}
\mathbb{G}_{\text{EMA}}^{t} = \beta \cdot \mathbb{G}_{\text{EMA}}^{t-1} + (1 - \beta) \cdot \mathbb{G}_1^{t}, \quad \mathbb{G}_{\text{EMA}}^{0} = \mathbb{G}_1^{0},
\label{eq:gsema}
\end{gather}
where \(t\) and \(t-1\) denote the current and previous training iterations, respectively. 
Here, we slightly abuse set notation for simplicity: the weighted summation between \(\mathbb{G}_1\) and \(\mathbb{G}_{\text{EMA}}\) is performed element-wise over corresponding Gaussian attributes, such as the centroids, opacities, and SH coefficients. 
We then rewrite the consistency regularization in Eq.~\ref{eq:dualmutual} with the EMA proxy as follows:
\begin{equation}
\mathcal{L}_{me} = \|\hat{\mathbf{I}}^{\mathbb{G}_{\text{EMA}}}_1 - \hat{\mathbf{I}}^{\mathbb{G}_1}_1\|_1,
\end{equation}
where \(\hat{\mathbf{I}}^{\mathbb{G}_{\text{EMA}}}_1 = \mathrm{Rasterize}(\mathbb{G}_{\text{EMA}}, \mathbf{V}_1)\) is the rendering of the EMA Gaussians from view $\mathbf{V}_1$. 
Since only one 3DGS model requires gradient updates, and the EMA update is a simple weighted average, this approach greatly improves training efficiency while preserving the benefits of dual-model consistency.


\noindent\textbf{Dynamic update.} 
Standard EMA is primarily designed for neural networks, where the number of parameters is typically fixed throughout training~\cite{tarvainen_mean_2017, he_momentum_2020}.
However, applying it to 3DGS presents unique challenges, as the number of Gaussians dynamically changes during training due to operations such as cloning, splitting, and pruning~\cite{kerbl_3d_2023}.


To support this dynamic data structure, we develop a dynamic EMA mechanism by introducing the following rules: 1) \textbf{Cloning}: When a Gaussian is cloned, its EMA attributes are also cloned.
2) \textbf{Pruning}: When a Gaussian is pruned, its EMA counterpart is removed as well.
3) \textbf{Splitting}: When a Gaussian splits into two, attributes that undergo discontinuous changes, \textit{i.e.}, the centroids and variances, are reinitialized in the EMA according to the values of the split Gaussians.
    The remaining attributes (e.g., opacities and SH coefficients) are directly inherited from the original EMA representation.

\noindent\textbf{Alternating masking strategy.}
Since only one model is trainable in our EMA framework, the original asymmetric training strategy used in Eq.~\ref{eq:finalloss} (\textit{i.e.}, $\mathbf{M}_h$ and $\mathbf{M}_s$) is not directly applicable.
Instead, we propose an alternating masking strategy by switching between the hard mask $\mathbf{M}_h$ and the soft mask $\mathbf{M}_s$ for training $\mathbb{G}_1$, which retains the complementary advantages from both decisive, rule-based filtering and adaptive, learned filtering.
The final loss for our dynamic EMA framework can be written as:
\begin{equation}
\mathcal{L} = \mathcal{L}_{r1}^{\mathbf{M}_{h/s}} + \lambda_m \mathcal{L}_{me} + \lambda_{\text{mask}} \mathcal{L}_{\text{mask}},
\end{equation}
where $\mathbf{M}_{h/s}$ indicates alternating between masks.
This strategy essentially injects randomness into the EMA update process, promoting diversity in optimization and reducing overfitting to erroneous supervision signals.
Note that we also explored other forms of randomness, including randomly mixing up EMA renderings with ground truth and applying random dropout of Gaussian primitives. Nevertheless, we empirically find that alternating masking remains the most effective approach.


\noindent\textbf{Discussion.}
Our approach is related to prior works that also leverage EMA, such as \cite{tarvainen_mean_2017} and \cite{he_momentum_2020}, which apply EMA to neural networks for tasks like semi-supervised or unsupervised image classification.
However, our method diverges in key aspects: unlike these methods that operate in the context of neural networks, we apply EMA to 3DGS, a dynamic representation where the number of Gaussians evolves throughout training.
This necessitates our proposed dynamic EMA mechanism, which adapts EMA updates to structural changes in the learned scene representation.
Additionally, we introduce an alternating masking strategy to preserve the benefits of asymmetric training even with a single learnable model.
These innovations mark significant departures from conventional EMA usage and highlight the contributions of this work.

\begin{table}[t]
\caption{Quantitative results on the NeRF On-the-go dataset~\cite{ren_nerfonthego_2024}. Efficiency is reported in terms of average training hours per scene. The best and second-best results are highlighted in \textbf{bold} and \underline{underline}, respectively.}
\label{tab:onthego}
\centering
\resizebox{\linewidth}{!}{
\begin{tabular}{lllllllllll}
\toprule
Scene & \multicolumn{3}{c}{High Occlusion} & \multicolumn{3}{c}{Medium Occlusion} & \multicolumn{3}{c}{Low Occlusion} & \\
\midrule
Method & PSNR$\uparrow$ & SSIM$\uparrow$ & LPIPS$\downarrow$ & PSNR$\uparrow$ & SSIM$\uparrow$ & LPIPS$\downarrow$ & PSNR$\uparrow$ & SSIM$\uparrow$ & LPIPS$\downarrow$ & Hrs. \\
\midrule
RobustNeRF \cite{sabour_robustnerf_2023} & 20.60 & 0.602 & 0.379 & 21.72 & 0.741 & 0.248 & 16.60 & 0.407 & 0.480 & - \\
NeRF On-the-go \cite{ren_nerfonthego_2024} & 22.37 & 0.753 & 0.212 & 22.50 & 0.780 & 0.205 & 20.13 & 0.627 & 0.287 & 43 \\
3DGS \cite{kerbl_3d_2023} & 19.03 & 0.649 & 0.340 & 19.19 & 0.709 & 0.220 & 19.68 & 0.649 & 0.199 & 0.35 \\
Mip-Splatting \cite{yu_mip-splatting_2024}  & 19.25 & 0.664 & 0.333 & 19.73 & 0.684 & 0.279 & 20.03 & 0.661 & 0.195 & 0.16 \\
GS-W \cite{zhang_gsw_2024} & 18.52 & 0.645 & 0.335 & 21.04 & 0.737 & 0.208 & 19.75 & 0.660 & 0.287 & 0.55 \\
WildGaussian \cite{kulhanek_wildgaussians_2024}  & 23.03 & 0.771 & 0.172 & 22.80 & 0.811 & 0.092 & 20.62 & 0.658 & 0.235 & 0.50 \\
SLS-mlp \cite{sabour_spotlesssplats_2024} & 21.92 & 0.710 & 0.222 & 22.79 & 0.817 & 0.162 & 20.02 & 0.596 & 0.276 & - \\
HybridGS \cite{lin_hybridgs_2025} & 23.05 & 0.768 & 0.204 & 23.51 & 0.830 & 0.160 & 21.42 & 0.684 & 0.268 & 0.18 \\
\midrule
Ours (GS-GS) & \textbf{24.34} & \textbf{0.825} & \textbf{0.150} & \textbf{24.56} & \textbf{0.872} & \textbf{0.090} & \textbf{21.91} & \textbf{0.728} & \underline{0.189} & 0.28 \\
Ours (EMA-GS) & \underline{24.12} & \underline{0.818} & \underline{0.154} & \underline{24.32} & \underline{0.864} & \underline{0.090} & \underline{21.77} & \underline{0.722} & \textbf{0.162} & 0.18 \\
\bottomrule
\end{tabular}
}
\vspace{-2mm}
\end{table}

\begin{figure}[t]
\centering
\begin{minipage}[b]{0.195\linewidth}
    \centering
    {\includegraphics[width=\linewidth]{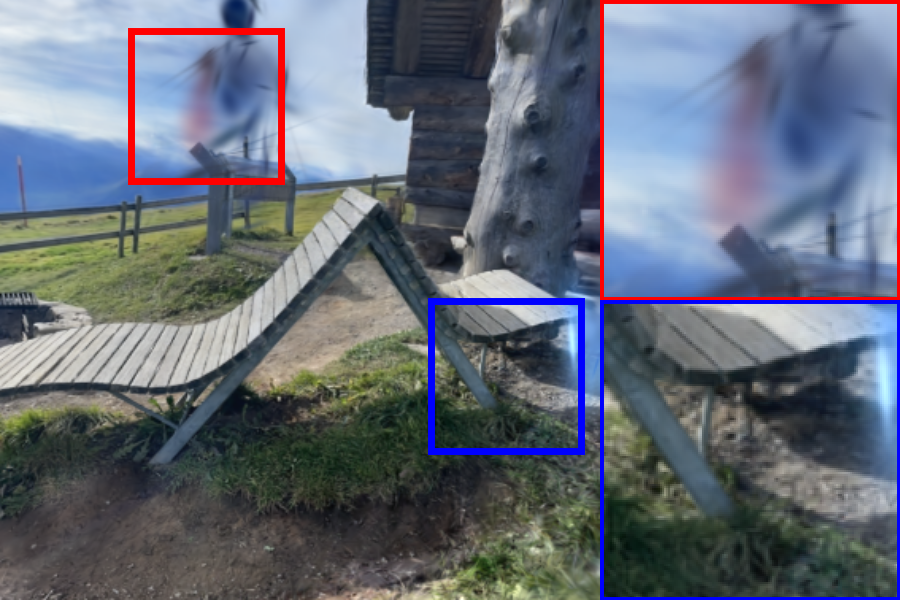}}
    {\includegraphics[width=\linewidth]{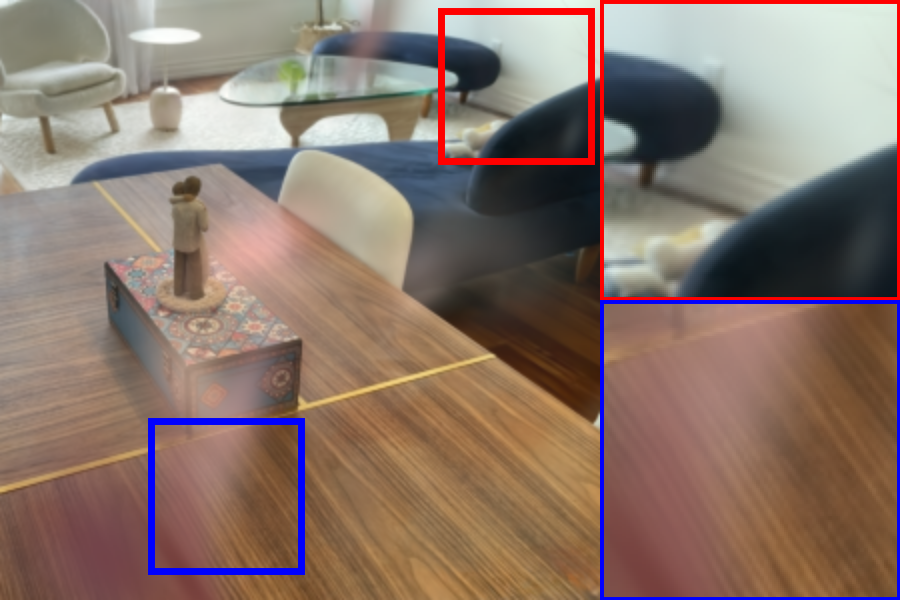}}
    \centerline{Mip-Splatting \cite{yu_mip-splatting_2024}}
\end{minipage}
\hfill
\begin{minipage}[b]{0.195\linewidth}
    \centering    
    {\includegraphics[width=\linewidth]{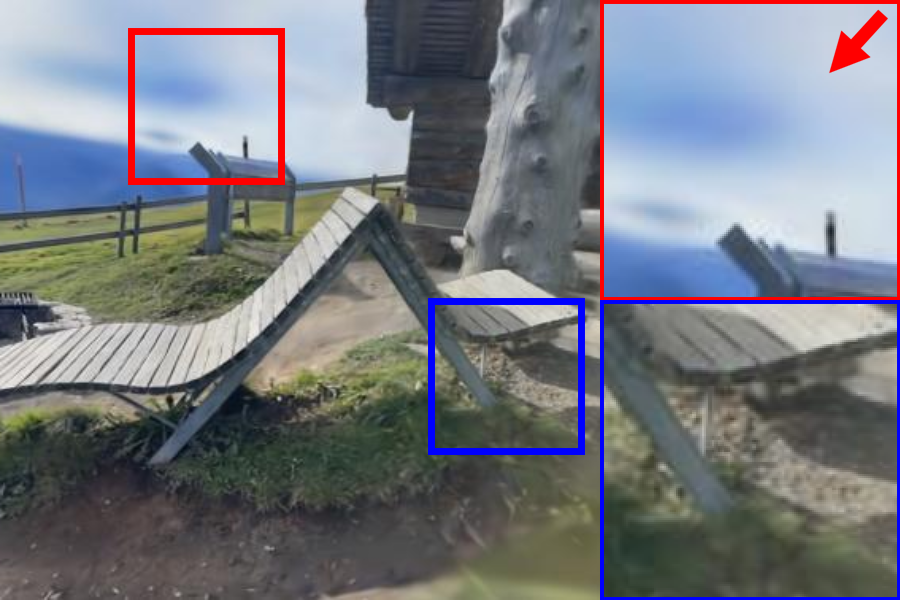}}
    {\includegraphics[width=\linewidth]{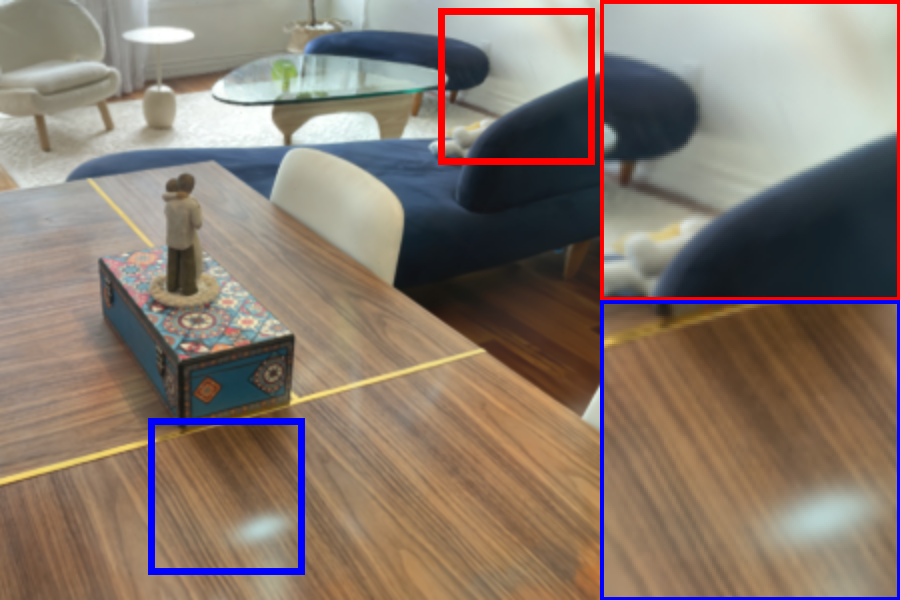}}
    \centerline{HybridGS \cite{lin_hybridgs_2025}}
\end{minipage}
\hfill
\begin{minipage}[b]{0.195\linewidth}
    \centering
    {\includegraphics[width=\linewidth]{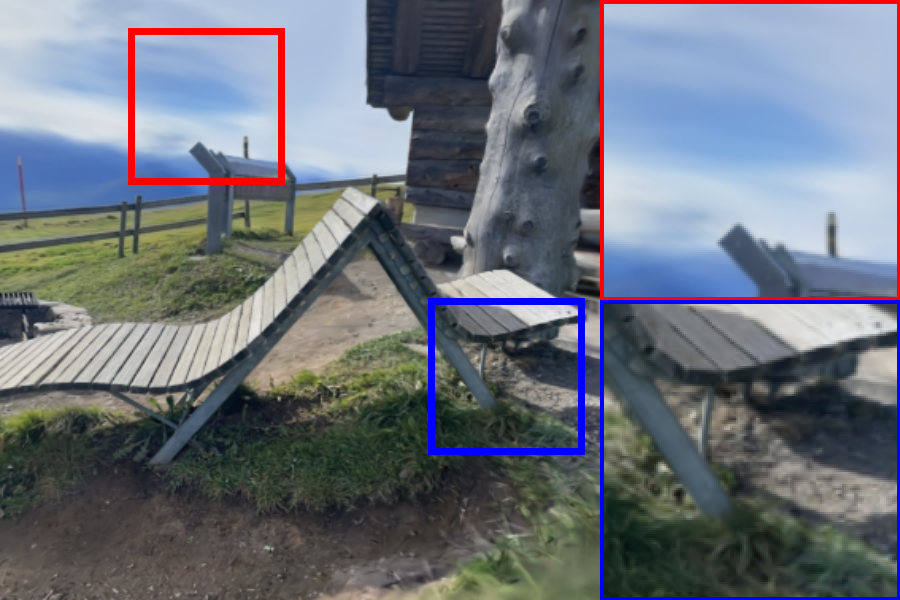}}
        {\includegraphics[width=\linewidth]{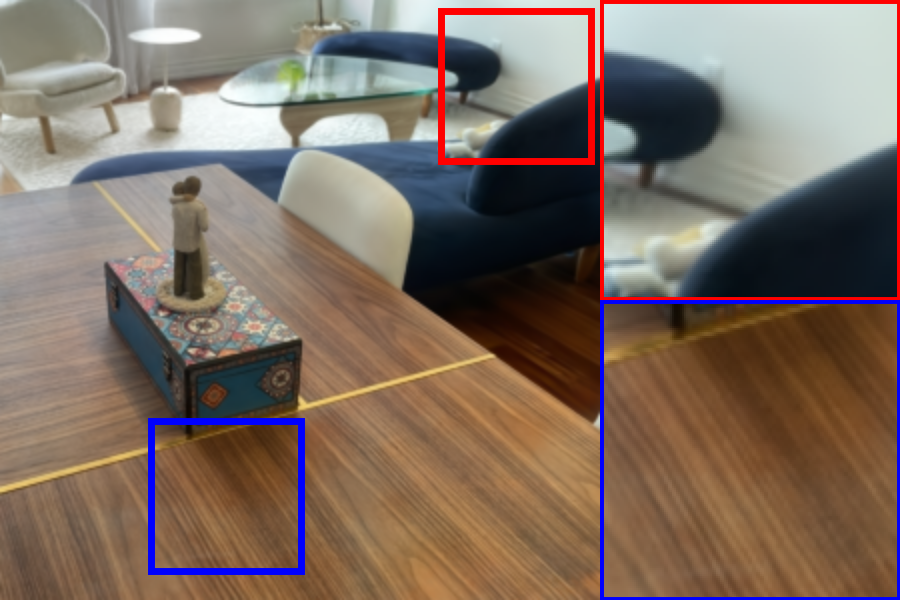}}
    \centerline{Our (EMA-GS)}
\end{minipage}
\hfill
\begin{minipage}[b]{0.195\linewidth}
    \centering
    {\includegraphics[width=\linewidth]{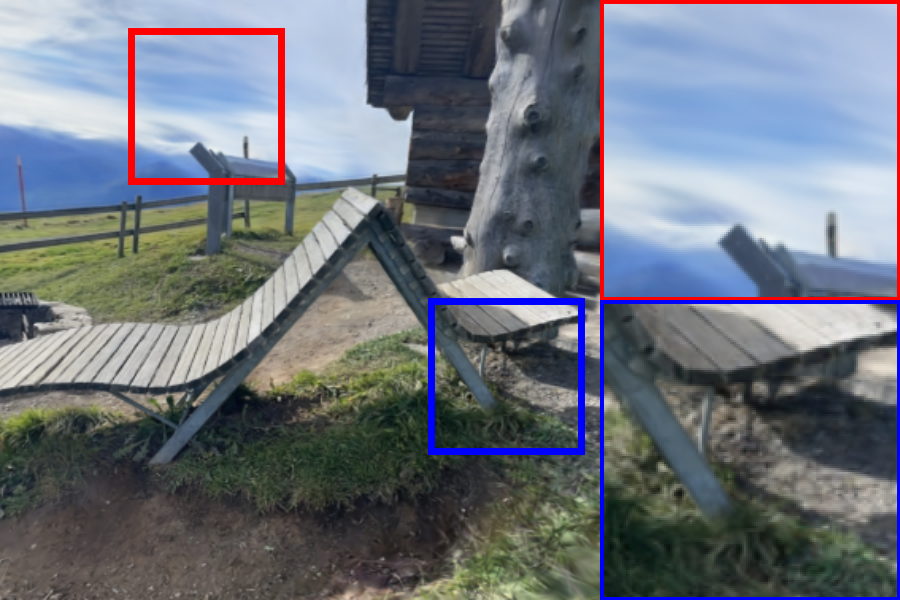}}
    {\includegraphics[width=\linewidth]{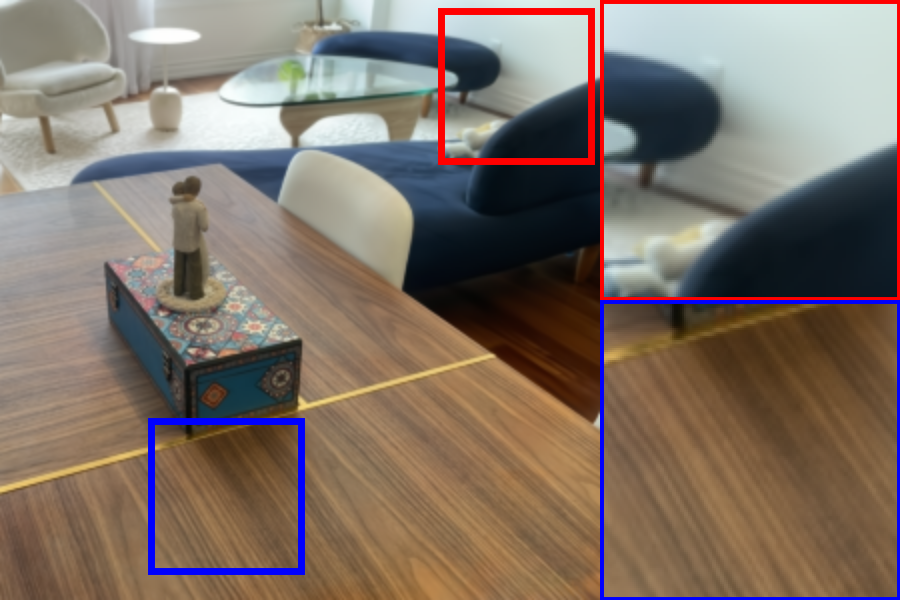}}
    \centerline{Ours (GS-GS)}
\end{minipage}
\hfill
\begin{minipage}[b]{0.195\linewidth}
    \centering
    {\includegraphics[width=\linewidth]{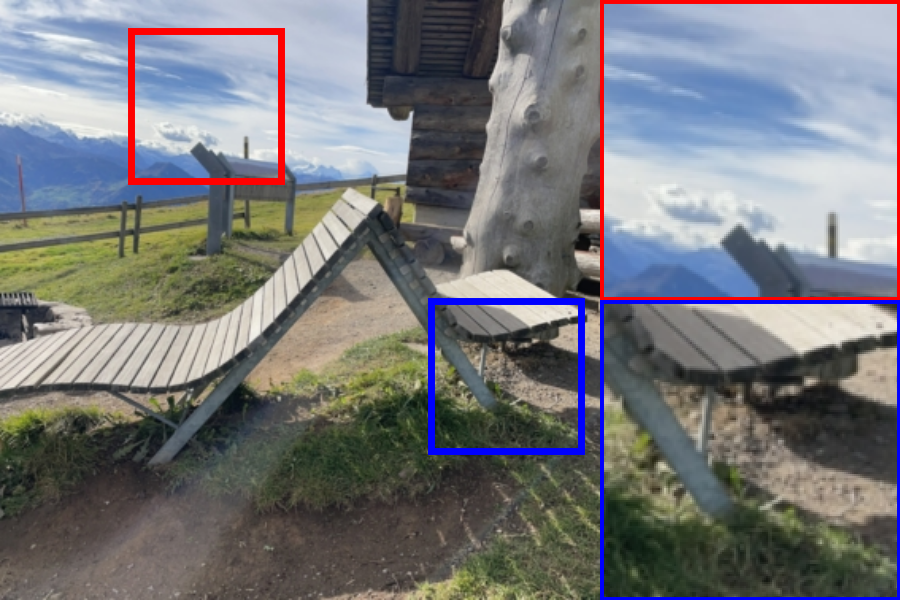}}
    {\includegraphics[width=\linewidth]{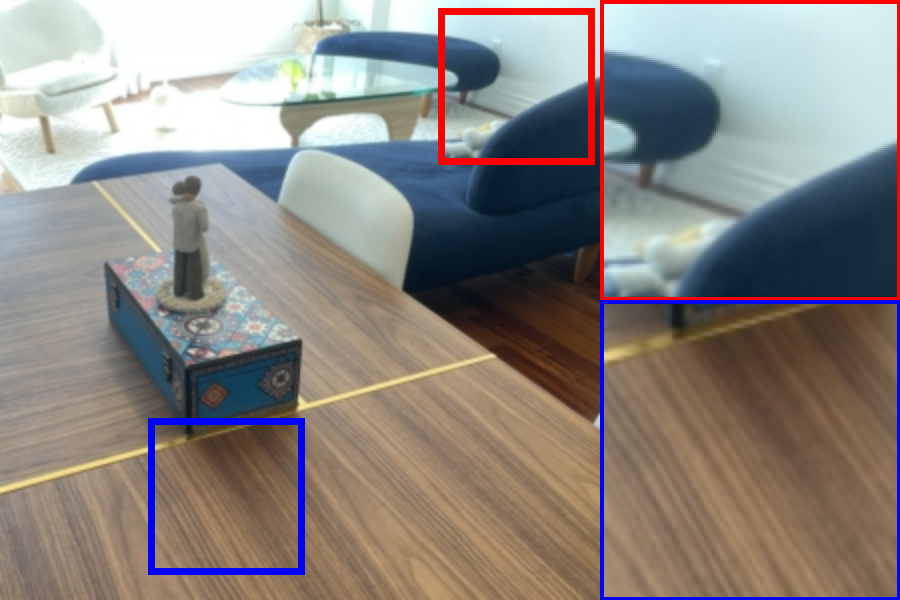}}
    \centerline{Ground Truth}
\end{minipage}
\hfill
\vspace*{-5mm}
\caption{Qualitative results on the NeRF On-the-go~\cite{ren_nerfonthego_2024} (top) and the RobustNeRF~\cite{sabour_robustnerf_2023} (bottom) datasets.}
\label{fig:onthegoandrobustnerf}
\end{figure}

\section{Experiments}
We evaluate our method on three in-the-wild datasets: the NeRF On-the-go dataset~\cite{ren_nerfonthego_2024}, the RobustNeRF dataset~\cite{sabour_robustnerf_2023}, and the PhotoTourism dataset~\cite{jin_phototourism_2020}. NeRF On-the-go and RobustNeRF mainly suffer from transient distractors. PhotoTourism contains both distractors and dynamic lighting.
We denote our full dual-model approach as GS-GS, and the efficient variant with the EMA proxy as EMA-GS. Implementation details can be found in Section \ref{sec:implement} of the supplementary material.

\subsection{Comparison with SOTA}
As shown in Tables~\ref{tab:onthego}-\ref{tab:phototourism}, the proposed \modelname{} framework, in both the GS-GS and EMA-GS setups, consistently outperforms all baselines by a significant margin across all evaluation datasets, highlighting the robustness and generality of our approach. 
Visual comparisons are provided in Figure~\ref{fig:onthegoandrobustnerf}.

In terms of training efficiency, our GS-GS model trains in an average of 30 minutes per scene on the NeRF On-the-go and RobustNeRF datasets, while the EMA proxy further reduces training time by one-third. For the PhotoTourism dataset, which features high-resolution imagery and dynamic lighting, EMA-GS cuts training time from 7.2 hours to 2.9 hours compared to the previous SOTA~\cite{kulhanek_wildgaussians_2024} while achieving better reconstruction quality.

\begin{table}[t]
\caption{Quantitative results on the RobustNeRF dataset~\cite{sabour_robustnerf_2023}. Efficiency is reported in terms of average training hours per scene. The best and second-best results are highlighted in \textbf{bold} and \underline{underline}, respectively.}
\label{tab:robustnerf}
\centering
\resizebox{\linewidth}{!}{
\begin{tabular}{llllllllllllll}
\toprule
Scene & \multicolumn{3}{c}{Statue} & \multicolumn{3}{c}{Android} & \multicolumn{3}{c}{Yoda} & \multicolumn{3}{c}{Crab} &  \\
\midrule
Method & PSNR$\uparrow$ & SSIM$\uparrow$ & LPIPS$\downarrow$ & PSNR$\uparrow$ & SSIM$\uparrow$ & LPIPS$\downarrow$ & PSNR$\uparrow$ & SSIM$\uparrow$ & LPIPS$\downarrow$ & PSNR$\uparrow$ & SSIM$\uparrow$ & LPIPS$\downarrow$ & Hrs. \\
\midrule
RobustNeRF \cite{sabour_robustnerf_2023}    & 20.60 & 0.760 & 0.150 & 23.28 & 0.750 & 0.130 & 29.78 & 0.820 & 0.150 & -     & -     & -     & -    \\
NeRF On-the-go \cite{ren_nerfonthego_2024} & 21.58 & 0.770 & 0.240 & 23.50 & 0.750 & 0.210 & 29.96 & 0.830 & 0.240 & -     & -     & -     & -    \\
3DGS \cite{kerbl_3d_2023}         & 21.02 & 0.810 & 0.160 & 23.11 & 0.810 & 0.130 & 26.33 & 0.910 & 0.140 & 29.74 & -     & -     & -    \\
Mip-Splatting \cite{yu_mip-splatting_2024} & 22.08 & 0.860 & 0.135 & 23.45 & 0.801 & 0.106 & 27.96 & 0.933 & 0.136 & 29.18 & 0.929 & 0.129 & 0.14 \\
GS-W \cite{yu_mip-splatting_2024}         & 21.99 & 0.862 & 0.102 & 24.23 & 0.824 & 0.090 & 32.74 & 0.957 & 0.084 & 33.22 & 0.952 & 0.088 & 0.37 \\
WildGaussian \cite{kulhanek_wildgaussians_2024}  & 23.25 & 0.886 & 0.105 & 24.57 & 0.827 & 0.085 & 32.84 & 0.956 & 0.091 & 32.81 & 0.952 & 0.092 & 0.82 \\
SLS-mlp \cite{sabour_spotlesssplats_2024}      & 22.54 & 0.840 & 0.130 & 25.05 & 0.850 & 0.090 & 33.66 & 0.960 & 0.100 & 34.43 & -     & -     & -    \\
HybridGS \cite{lin_hybridgs_2025}     & 22.93 & 0.870 & 0.100 & 25.15 & \underline{0.850} & \underline{0.070} & 35.32 & 0.960 & \textbf{0.070} & 35.17 & 0.960 & 0.080 & -   \\
\midrule
Ours (GS-GS)          & \underline{23.47}      & \textbf{0.894} & \underline{0.097} & \textbf{25.61} & \textbf{0.857} & 0.071 & \textbf{37.18} & \textbf{0.969} & \underline{0.074} & \textbf{36.18} & \textbf{0.964} & \textbf{0.078} & 0.31 \\
Ours (EMA-GS)           & \textbf{23.49} & \underline{0.890} & \textbf{0.096} & \underline{25.47} & 0.849 & \textbf{0.068} & \underline{36.50} & \underline{0.967} & 0.077 & \underline{35.60} & \underline{0.961} & \underline{0.080} & 0.21 \\
\bottomrule
\end{tabular}
}
\end{table}

\begin{figure}[t]
\centering
\begin{minipage}[b]{0.195\linewidth}
    \centering
    {\includegraphics[width=\linewidth]{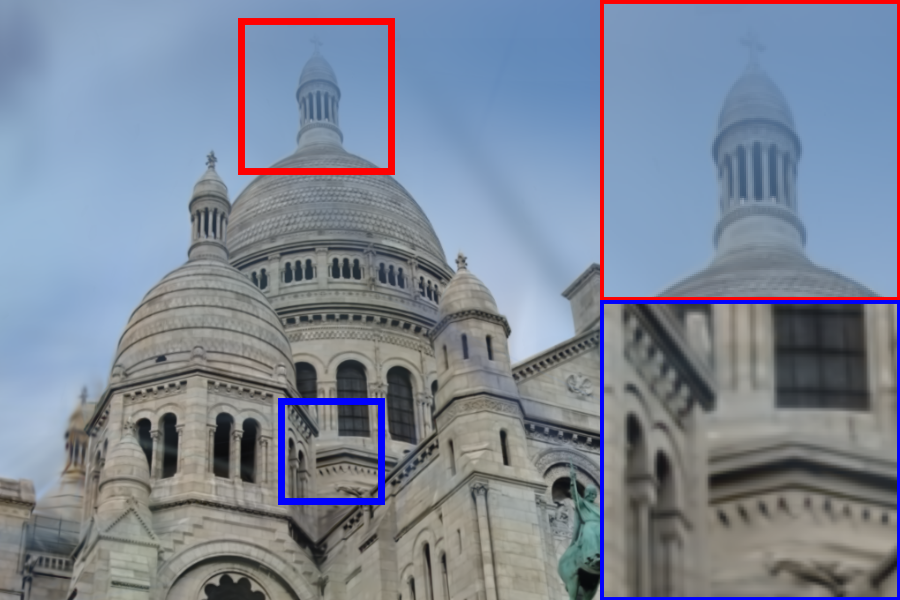}}
    \centerline{Mip-Splatting \cite{yu_mip-splatting_2024}}
\end{minipage}
\hfill
\begin{minipage}[b]{0.195\linewidth}
    \centering
    {\includegraphics[width=\linewidth]{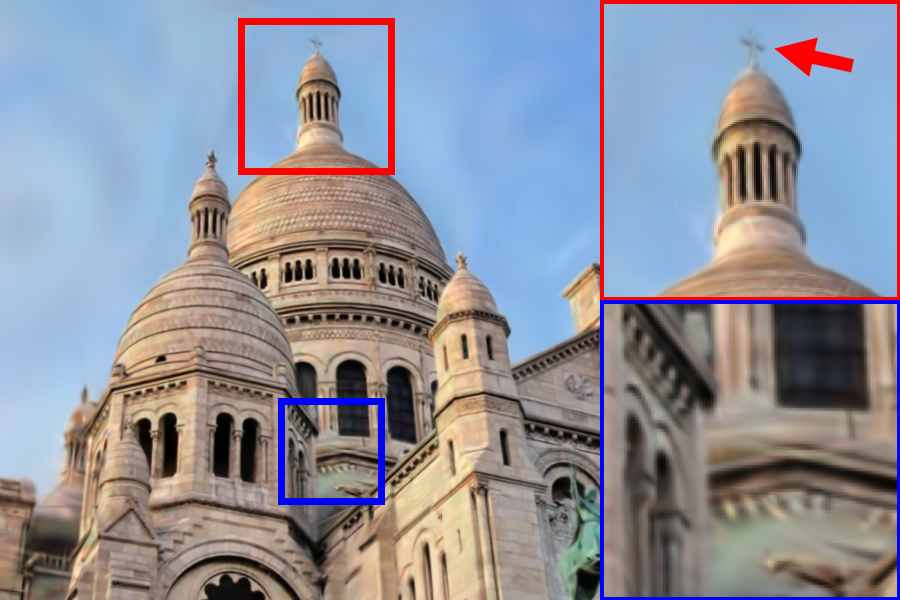}}
    \centerline{WildGaussian \cite{kulhanek_wildgaussians_2024}}
\end{minipage}
\hfill
\begin{minipage}[b]{0.195\linewidth}
    \centering
    {\includegraphics[width=\linewidth]{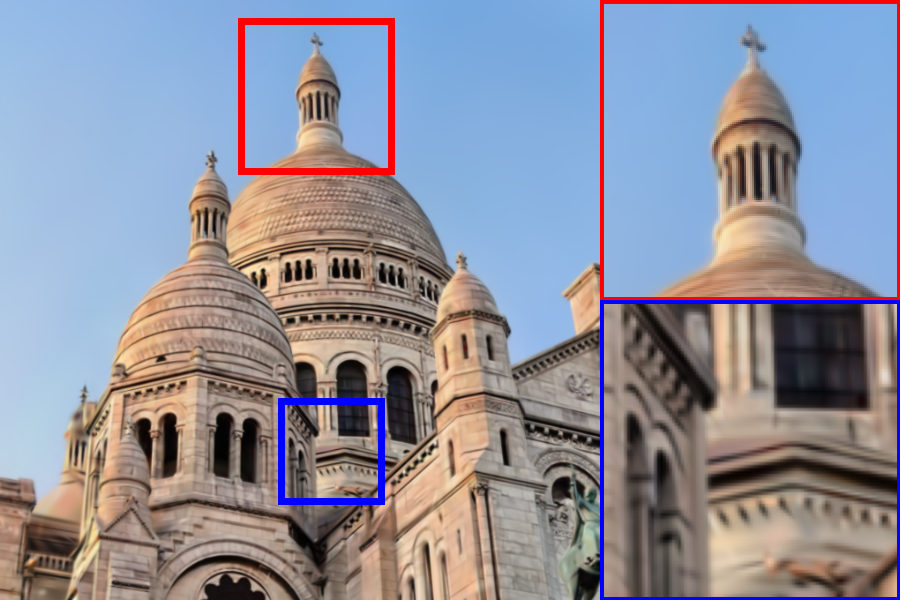}}
    \centerline{Our (EMA-GS)}
\end{minipage}
\hfill
\begin{minipage}[b]{0.195\linewidth}
    \centering
    {\includegraphics[width=\linewidth]{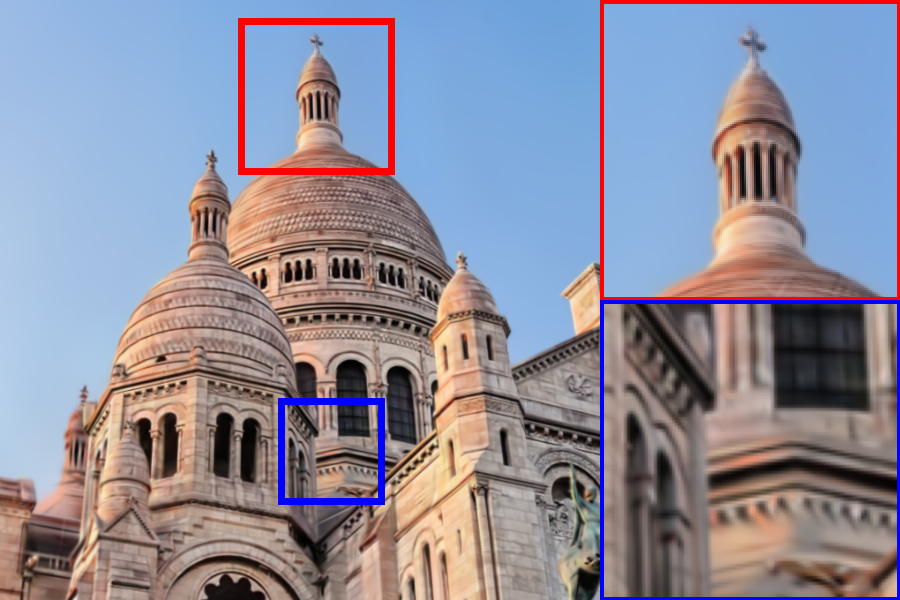}}
    \centerline{Ours (GS-GS)}
\end{minipage}
\hfill
\begin{minipage}[b]{0.195\linewidth}
    \centering
    {\includegraphics[width=\linewidth]{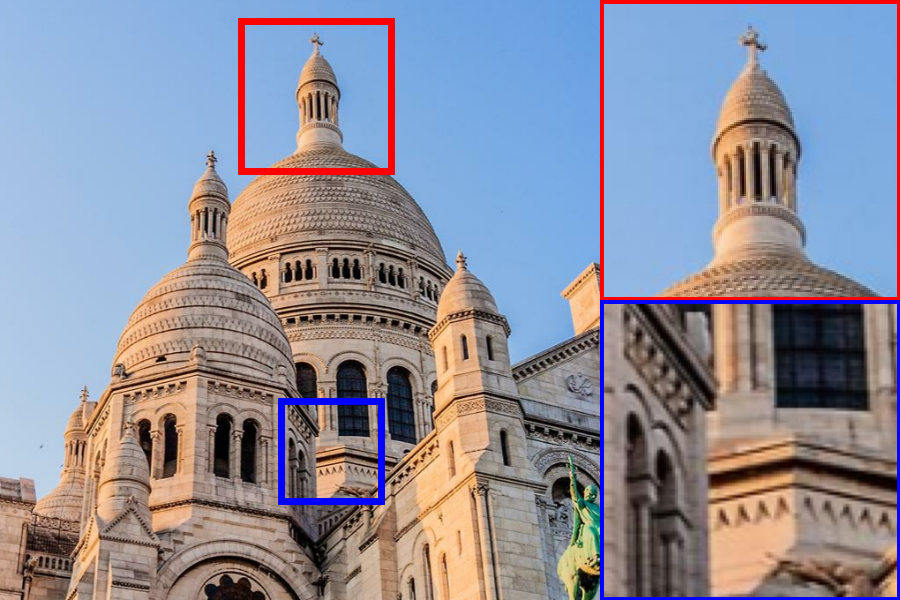}}
    \centerline{Ground Truth}
\end{minipage}
\hfill
\vspace*{-3mm}
\caption{Qualitative results on the PhotoTourism dataset~\cite{jin_phototourism_2020}. }
\vspace{-2mm}
\label{fig:phototourism}
\end{figure}

\begin{table}[t]
\caption{Quantitative results on the PhotoTourism dataset~\cite{jin_phototourism_2020}. Efficiency is reported in terms of average training hours per scene. The best and second-best results are highlighted in \textbf{bold} and \underline{underline}, respectively. We did not compare with HybridGS~\cite{lin_hybridgs_2025} on PhotoTourism, as it does not consider varying illumination, and cannot be customized or retrained due the absence of training code.}
\label{tab:phototourism}
\centering
\resizebox{\linewidth}{!}{
\begin{threeparttable}[b]
\begin{tabular}{lllllllllll}
\toprule
Scene & \multicolumn{3}{c}{Brandenburg Gate} & \multicolumn{3}{c}{Sacre Coeur} & \multicolumn{3}{c}{Trevi Fountain} &  \\
\midrule
Method & PSNR$\uparrow$ & SSIM$\uparrow$ & LPIPS$\downarrow$ & PSNR$\uparrow$ & SSIM$\uparrow$ & LPIPS$\downarrow$ & PSNR$\uparrow$ & SSIM$\uparrow$ & LPIPS$\downarrow$ & Hrs. \\
\midrule
NeRF \cite{vedaldi_nerf_2020}         & 18.90 & 0.882 & 0.138 & 15.60 & 0.846 & 0.163 & 16.14 & 0.696 & 0.282 &    - \\
3DGS \cite{kerbl_3d_2023}         & 19.37 & 0.880 & 0.141 & 17.44 & 0.835 & 0.204 & 17.58 & 0.709 & 0.266 & 2.2 \\
Mip-Splatting \cite{yu_mip-splatting_2024} & 20.01 & 0.877 & 0.166 & 17.54 & 0.831 & 0.203 & 17.36 & 0.684 & 0.319 & 2.3 \\
NeRF-W \cite{martin-brualla_nerfw_2021}       & 24.17 & 0.891 & 0.152 & 19.20 & 0.803 & 0.192 & 18.97 & 0.698 & 0.265 & 164 \\
Ha-NeRF \cite{Chen_hanerf_2022}      & 24.04 & 0.887 & 0.139 & 20.02 & 0.801 & 0.171 & 20.18 & 0.691 & 0.223 & 452 \\
K-Planes \cite{Fridovich-Keil_kplane_2023}     & 25.49 & 0.879 & 0.224 & 20.61 & 0.774 & 0.265 & 22.67 & 0.714 & 0.317 & 0.6 \\
RefinedFields \cite{kassab_refinedfields_2023} & 26.64 & 0.886 & - & 22.26 & 0.817 & - & 23.42 & 0.737 & - & 150 \\
GS-W \cite{zhang_gsw_2024}         & 23.51 & 0.897 & 0.166 & 19.39 & 0.825 & 0.211 & 20.06 & 0.723 & 0.274 & 1.2 \\
SWAG \cite{leonardis_swag_2025}         & 26.33 & 0.929 & 0.139 & 21.16 & 0.860 & 0.185 & 23.10 & \textbf{0.815} & \underline{0.208} & 0.8 \\
WildGaussian \cite{kulhanek_wildgaussians_2024} & 27.77 & 0.927 & 0.133 & 22.56 & 0.859 & 0.177 & 23.63 & 0.766 & 0.228 & 7.2 \\
\midrule
Ours (GS-GS) & \textbf{28.56} & \textbf{0.938} & \textbf{0.109} & \textbf{23.78} & \textbf{0.887} & \textbf{0.139} & \textbf{24.52} & \underline{0.790} & \textbf{0.202} & 5.3 \\
Ours (EMA-GS)    & \underline{28.50} & \underline{0.937} & \underline{0.115} & \underline{23.37} & \underline{0.882} & \underline{0.150} & \underline{23.85} & 0.775 & 0.242 & 2.9\\     
\bottomrule
\end{tabular}
\end{threeparttable}
}
\end{table}



\subsection{Ablation study}
\noindent \textbf{Dual 3DGS framework.}
As shown in Table \ref{tab:ablation}, the dual 3DGS framework, whether implemented directly or via an EMA proxy, consistently outperforms the single-model baseline across three datasets (L1 \textit{vs.} L5 \& L12 in Table \ref{tab:ablation}). Importantly, the mutual consistency loss is a crucial component. Removing it reduces the framework to a simple ensemble, which is shown to be ineffective, leading to an average drop of 0.5 dB in PSNR and consistent degradation in SSIM and LPIPS (L5 \textit{vs.} L9, L12 \textit{vs.} L16). 

Additionally, the GS-GS setup achieves better performance than its EMA proxy counterpart (L5 \textit{vs.} L12). We attribute this to two main factors: first, the GS-GS setup enables both models to be actively updated, effectively doubling the training iterations; second, the EMA proxy introduces confirmation bias due to its model accumulation nature, which limits its ability to correct erroneous predictions. 

\noindent \textbf{Distractor modeling.}
The results in Table~\ref{tab:ablation} further show that applying masks significantly improves reconstruction performance especially when distractors occupy a large portion of the input image. (L5 \textit{vs.} L6, L12 \textit{vs.} L13). 

Either \(\mathbf{M}_h\) or \(\mathbf{M}_s\) independently improves performance over the base model (for \(\mathbf{M}_h\), L2 \textit{vs.} L3, L6 \textit{vs.} L7, L13 \vs L14; for \(\mathbf{M}_s\), L2 \vs L4, L6 \vs L8, L13 \vs L15), indicating that both masks are effective, although they have different characteristics. 
This observation is further supported by the visual results in Figure~\ref{fig:mask_compare}, where the hard-selected mask performs well in simpler scenes with clearly defined regions (Figure~\ref{fig:mask_compare}-Top), while the self-supervised mask excels in more complex scenes containing multiple or ambiguous distractors (Figure~\ref{fig:mask_compare}-Bottom). Moreover, combining both masks leads to additional gains, confirming their complementary roles (L5 \vs L7 \& L8, L12 \vs L14 \& L15). 


\noindent \textbf{EMA proxy.}
The effectiveness of the EMA-GS setup heavily depends on the masking strategy (L5 \vs L7 \& L8). 
Using only a single type of mask often leads to performance similar to or even worse than the single-model baseline (L3 \vs L7, L4 \vs L8), indicating that confirmation bias can undermine robustness. This highlights the need for diverse masking signals to fully exploit the benefits of the EMA proxy. 
Moreover, as introduced in the Alternating Masking Strategy part of Section~\ref{sec:dynamicEMA}, we have also tried other forms of regularization to improve the performance of our EMA model, such as random mixup or dropout. 
However, as shown in L10 and L11 of Table \ref{tab:ablation}, these methods do work as good as our original approach of alternating between the two masking strategies.


\begin{table}[t]
\caption{Effectiveness of different modules. The first block presents results from a single base model using different mask strategies. The second and third blocks evaluate the EMA-GS and GS-GS setups, respectively. ``w/ $\mathbf{M}_{h/s}$'' indicates alternating between \(\mathbf{M}_h\) and \(\mathbf{M}_s\) at each training iteration. 
``w/o $\mathbf{M}$'' denotes that no mask is applied. The results are averaged across scenes within each dataset.}
\label{tab:ablation}
\centering
\resizebox{\linewidth}{!}{
\begin{threeparttable}[b]
\begin{tabular}{lllllllllll}
\toprule
& Dataset & \multicolumn{3}{c}{PhotoTourism} & \multicolumn{3}{c}{NeRF On-the-go} & \multicolumn{3}{c}{RobustNeRF} \\
\midrule
Line & Method & PSNR$\uparrow$ & SSIM$\uparrow$ & LPIPS$\downarrow$ & PSNR$\uparrow$ & SSIM$\uparrow$ & LPIPS$\downarrow$ & PSNR$\uparrow$ & SSIM$\uparrow$ & LPIPS$\downarrow$ \\
\midrule
1 & Single w/ \(\mathbf{M}_{h/s}\) & 24.76 & 0.864 & 0.167 & 22.97 & 0.798 & 0.133 & 29.62 & 0.914 & 0.083 \\
2 & Single w/o \(\mathbf{M}\) & 24.68 & 0.859 & 0.174 & 19.67 & 0.669 & 0.269 & 25.67 & 0.881 & 0.126 \\		
3 & Single w/o \(\mathbf{M}_s\) & 24.82 & 0.862 & 0.172 & 22.97 & 0.802 & 0.125 & 28.97 & 0.908 & 0.088 \\
4 & Single w/o \(\mathbf{M}_h\) & 24.83 & 0.862 & 0.171 & 22.24 & 0.783 & 0.153 & 29.38 & 0.912 & 0.086 \\
\midrule
5 & EMA-GS  & 25.24 & 0.864 & 0.169 & 23.40 & 0.801 & 0.135 & 30.27 & 0.917 & 0.080 \\
6 & EMA-GS w/o \(\mathbf{M}\)     & 24.70 & 0.862 & 0.170 & 21.68 & 0.766 & 0.172 & 28.42 & 0.905 & 0.092 \\
7 & EMA-GS w/o \(\mathbf{M}_s\) & 24.76 & 0.864 & 0.166 & 22.94 & 0.802 & 0.126 & 29.08 & 0.909 & 0.089 \\
8 & EMA-GS w/o \(\mathbf{M}_h\) & 24.84 & 0.863 & 0.169 & 22.15 & 0.780 & 0.159 & 29.29 & 0.911 & 0.087 \\
9 & EMA-GS w/o \(\mathcal{L}_{m}\) & 24.73 & 0.861 & 0.174 & 23.10 & 0.801 & 0.132 & 29.72 & 0.915 & 0.081 \\
10 & EMA-GS w/ Mixup & 25.13 & 0.866 & 0.162 & 23.39 & 0.802 & 0.132 & 30.18 & 0.915 & 0.081 \\
11 & EMA-GS w/ Dropout & 25.11 & 0.863 & 0.171 & 23.39 & 0.804 & 0.130 & 30.19 & 0.913 & 0.082\\
\midrule 
12 & GS-GS & 25.62 & 0.872 & 0.150 & 23.61 & 0.809 & 0.143 & 30.61 & 0.921 & 0.080 \\
13 & GS-GS w/o \(\mathbf{M}\)    & 25.09 & 0.867 & 0.160 & 22.39 & 0.785 & 0.164 & 29.45 & 0.914 & 0.087 \\
14 & GS-GS w/o \(\mathbf{M}_s\) & 25.24 & 0.867 & 0.159 & 23.32 & 0.812 & 0.132 & 30.38 & 0.920 & 0.079 \\
15 & GS-GS w/o \(\mathbf{M}_h\) & 25.11 & 0.867 & 0.161 & 22.76 & 0.794 & 0.157 & 30.17 & 0.919 & 0.082 \\
16 & GS-GS w/o \(\mathcal{L}_{m}\)      & 25.08 & 0.869 & 0.155 & 23.13 & 0.808 & 0.135 & 29.92 & 0.918 & 0.082 \\
\bottomrule
\end{tabular}
\end{threeparttable}
}
\vspace{-3mm}
\end{table}

\section{Conclusion}
In this work, we present \modelname{}, a robust and efficient framework for 3D scene reconstruction in unconstrained, in-the-wild environments. Our method employs two 3DGS models guided by distinct masking strategies to enforce cross-model consistency, effectively mitigating artifacts caused by low-quality observations. To further improve training efficiency, we introduce a dynamic EMA proxy that significantly reduces computational cost with minimal impact on performance. Extensive experiments on three challenging real-world datasets validate the effectiveness and generality of our approach.

\clearpage
\bibliographystyle{plainnat}
\bibliography{ref}

\clearpage
\section*{NeurIPS Paper Checklist}

\begin{enumerate}

\item {\bf Claims}
    \item[] Question: Do the main claims made in the abstract and introduction accurately reflect the paper's contributions and scope?
    \item[] Answer: \answerYes{} 
    \item[] Justification: The abstract and introduction clearly state the paper's focus on novel view synthesis with in-the-wild image collections. The key contributions (including the asymmetric dual-model framework, mutual consistency regularization, and the integration of complementary masking strategies) are explicitly outlined in the methodology and validated through extensive experiments.
    \item[] Guidelines:
    \begin{itemize}
        \item The answer NA means that the abstract and introduction do not include the claims made in the paper.
        \item The abstract and/or introduction should clearly State the claims made, including the contributions made in the paper and important assumptions and limitations. A No or NA answer to this question will not be perceived well by the reviewers. 
        \item The claims made should match theoretical and experimental results, and reflect how much the results can be expected to generalize to other settings. 
        \item It is fine to include aspirational goals as motivation as long as it is clear that these goals are not attained by the paper. 
    \end{itemize}

\item {\bf Limitations}
    \item[] Question: Does the paper discuss the limitations of the work performed by the authors?
    \item[] Answer: \answerYes{} 
    \item[] Justification: The limitations of our approach are discussed in the supplementary material.
    \item[] Guidelines:
    \begin{itemize}
        \item The answer NA means that the paper has no limitation while the answer No means that the paper has limitations, but those are not discussed in the paper. 
        \item The authors are encouraged to create a separate "Limitations" section in their paper.
        \item The paper should point out any strong assumptions and how robust the results are to violations of these assumptions (e.g., independence assumptions, noiseless settings, model well-specification, asymptotic approximations only holding locally). The authors should reflect on how these assumptions might be violated in practice and what the implications would be.
        \item The authors should reflect on the scope of the claims made, e.g., if the approach was only tested on a few datasets or with a few runs. In general, empirical results often depend on implicit assumptions, which should be articulated.
        \item The authors should reflect on the factors that influence the performance of the approach. For example, a facial recognition algorithm may perform poorly when image resolution is low or images are taken in low lighting. Or a speech-to-text system might not be used reliably to provide closed captions for online lectures because it fails to handle technical jargon.
        \item The authors should discuss the computational efficiency of the proposed algorithms and how they scale with dataset size.
        \item If applicable, the authors should discuss possible limitations of their approach to address problems of privacy and fairness.
        \item While the authors might fear that complete honesty about limitations might be used by reviewers as grounds for rejection, a worse outcome might be that reviewers discover limitations that aren't acknowledged in the paper. The authors should use their best judgment and recognize that individual actions in favor of transparency play an important role in developing norms that preserve the integrity of the community. Reviewers will be specifically instructed to not penalize honesty concerning limitations.
    \end{itemize}

\item {\bf Theory assumptions and proofs}
    \item[] Question: For each theoretical result, does the paper provide the full set of assumptions and a complete (and correct) proof?
    \item[] Answer: \answerNA{} 
    \item[] Justification: The paper presents a learning-based method and does not include theoretical results or formal proofs.
    \item[] Guidelines:
    \begin{itemize}
        \item The answer NA means that the paper does not include theoretical results. 
        \item All the theorems, formulas, and proofs in the paper should be numbered and cross-referenced.
        \item All assumptions should be clearly Stated or referenced in the Statement of any theorems.
        \item The proofs can either appear in the main paper or the supplemental material, but if they appear in the supplemental material, the authors are encouraged to provide a short proof sketch to provide intuition. 
        \item Inversely, any informal proof provided in the core of the paper should be complemented by formal proofs provided in appendix or supplemental material.
        \item Theorems and Lemmas that the proof relies upon should be properly referenced. 
    \end{itemize}

    \item {\bf Experimental result reproducibility}
    \item[] Question: Does the paper fully disclose all the information needed to reproduce the main experimental results of the paper to the extent that it affects the main claims and/or conclusions of the paper (regardless of whether the code and data are provided or not)?
    \item[] Answer: \answerYes{} 
    \item[] Justification: The paper provides comprehensive details of the proposed method, along with descriptions of the experimental setup, including datasets, training procedures, hyperparameters, and evaluation metrics, ensuring the main results can be reliably reproduced and validated.
    \item[] Guidelines:
    \begin{itemize}
        \item The answer NA means that the paper does not include experiments.
        \item If the paper includes experiments, a No answer to this question will not be perceived well by the reviewers: Making the paper reproducible is important, regardless of whether the code and data are provided or not.
        \item If the contribution is a dataset and/or model, the authors should describe the steps taken to make their results reproducible or verifiable. 
        \item Depending on the contribution, reproducibility can be accomplished in various ways. For example, if the contribution is a novel architecture, describing the architecture fully might suffice, or if the contribution is a specific model and empirical evaluation, it may be necessary to either make it possible for others to replicate the model with the same dataset, or provide access to the model. In general. releasing code and data is often one good way to accomplish this, but reproducibility can also be provided via detailed instructions for how to replicate the results, access to a hosted model (e.g., in the case of a large language model), releasing of a model checkpoint, or other means that are appropriate to the research performed.
        \item While NeurIPS does not require releasing code, the conference does require all submissions to provide some reasonable avenue for reproducibility, which may depend on the nature of the contribution. For example
        \begin{enumerate}
            \item If the contribution is primarily a new algorithm, the paper should make it clear how to reproduce that algorithm.
            \item If the contribution is primarily a new model architecture, the paper should describe the architecture clearly and fully.
            \item If the contribution is a new model (e.g., a large language model), then there should either be a way to access this model for reproducing the results or a way to reproduce the model (e.g., with an open-source dataset or instructions for how to construct the dataset).
            \item We recognize that reproducibility may be tricky in some cases, in which case authors are welcome to describe the particular way they provide for reproducibility. In the case of closed-source models, it may be that access to the model is limited in some way (e.g., to registered users), but it should be possible for other researchers to have some path to reproducing or verifying the results.
        \end{enumerate}
    \end{itemize}

\item {\bf Open access to data and code}
    \item[] Question: Does the paper provide open access to the data and code, with sufficient instructions to faithfully reproduce the main experimental results, as described in supplemental material?
    \item[] Answer: \answerYes{} 
    \item[] Justification: The datasets used are publicly available, and the code will be released upon publication.
    \item[] Guidelines:
    \begin{itemize}
        \item The answer NA means that paper does not include experiments requiring code.
        \item Please see the NeurIPS code and data submission guidelines (\url{https://nips.cc/public/guides/CodeSubmissionPolicy}) for more details.
        \item While we encourage the release of code and data, we understand that this might not be possible, so “No” is an acceptable answer. Papers cannot be rejected simply for not including code, unless this is central to the contribution (e.g., for a new open-source benchmark).
        \item The instructions should contain the exact command and environment needed to run to reproduce the results. See the NeurIPS code and data submission guidelines (\url{https://nips.cc/public/guides/CodeSubmissionPolicy}) for more details.
        \item The authors should provide instructions on data access and preparation, including how to access the raw data, preprocessed data, intermediate data, and generated data, etc.
        \item The authors should provide scripts to reproduce all experimental results for the new proposed method and baselines. If only a subset of experiments are reproducible, they should State which ones are omitted from the script and why.
        \item At submission time, to preserve anonymity, the authors should release anonymized versions (if applicable).
        \item Providing as much information as possible in supplemental material (appended to the paper) is recommended, but including URLs to data and code is permitted.
    \end{itemize}

\item {\bf Experimental setting/details}
    \item[] Question: Does the paper specify all the training and test details (e.g., data splits, hyperparameters, how they were chosen, type of optimizer, etc.) necessary to understand the results?
    \item[] Answer: \answerYes{} 
    \item[] Justification: The paper provides all necessary training and testing details, with method-specific settings described in the main paper and 3DGS-related heuristics detailed in the supplementary material.
    \item[] Guidelines:
    \begin{itemize}
        \item The answer NA means that the paper does not include experiments.
        \item The experimental setting should be presented in the core of the paper to a level of detail that is necessary to appreciate the results and make sense of them.
        \item The full details can be provided either with the code, in appendix, or as supplemental material.
    \end{itemize}

\item {\bf Experiment statistical significance}
    \item[] Question: Does the paper report error bars suitably and correctly defined or other appropriate information about the statistical significance of the experiments?
    \item[] Answer: \answerNo{} 
    \item[] Justification: Error bars are not reported, as repeated runs showed no significant variation under identical settings.
    \item[] Guidelines:
    \begin{itemize}
        \item The answer NA means that the paper does not include experiments.
        \item The authors should answer "Yes" if the results are accompanied by error bars, confidence intervals, or statistical significance tests, at least for the experiments that support the main claims of the paper.
        \item The factors of variability that the error bars are capturing should be clearly Stated (for example, train/test split, initialization, random drawing of some parameter, or overall run with given experimental conditions).
        \item The method for calculating the error bars should be explained (closed form formula, call to a library function, bootstrap, etc.)
        \item The assumptions made should be given (e.g., Normally distributed errors).
        \item It should be clear whether the error bar is the standard deviation or the standard error of the mean.
        \item It is OK to report 1-sigma error bars, but one should State it. The authors should preferably report a 2-sigma error bar than State that they have a 96\% CI, if the hypothesis of Normality of errors is not verified.
        \item For asymmetric distributions, the authors should be careful not to show in tables or figures symmetric error bars that would yield results that are out of range (e.g. negative error rates).
        \item If error bars are reported in tables or plots, The authors should explain in the text how they were calculated and reference the corresponding figures or tables in the text.
    \end{itemize}

\item {\bf Experiments compute resources}
    \item[] Question: For each experiment, does the paper provide sufficient information on the computer resources (type of compute workers, memory, time of execution) needed to reproduce the experiments?
    \item[] Answer: \answerYes{} 
    \item[] Justification: The paper specifies the computational resources used for each experimental setting. Experiments were performed on an RTX 4090. This information, along with training times, is provided in the supplementary.
    \item[] Guidelines:
    \begin{itemize}
        \item The answer NA means that the paper does not include experiments.
        \item The paper should indicate the type of compute workers CPU or GPU, internal cluster, or cloud provider, including relevant memory and storage.
        \item The paper should provide the amount of compute required for each of the individual experimental runs as well as estimate the total compute. 
        \item The paper should disclose whether the full research project required more compute than the experiments reported in the paper (e.g., preliminary or failed experiments that didn't make it into the paper). 
    \end{itemize}
    
\item {\bf Code of ethics}
    \item[] Question: Does the research conducted in the paper conform, in every respect, with the NeurIPS Code of Ethics \url{https://neurips.cc/public/EthicsGuidelines}?
    \item[] Answer: \answerYes{} 
    \item[] Justification: The research fully complies with the NeurIPS Code of Ethics.
    \item[] Guidelines:
    \begin{itemize}
        \item The answer NA means that the authors have not reviewed the NeurIPS Code of Ethics.
        \item If the authors answer No, they should explain the special circumstances that require a deviation from the Code of Ethics.
        \item The authors should make sure to preserve anonymity (e.g., if there is a special consideration due to laws or regulations in their jurisdiction).
    \end{itemize}

\item {\bf Broader impacts}
    \item[] Question: Does the paper discuss both potential positive societal impacts and negative societal impacts of the work performed?
    \item[] Answer: \answerYes{} 
    \item[] Justification: By improving robustness under in-the-wild data, we aim to advance 3DGS models toward practical real-world applications. Potential risks, such as misuse in privacy-sensitive environments, are acknowledged in the broader context of scene reconstruction technologies.
    \item[] Guidelines:
    \begin{itemize}
        \item The answer NA means that there is no societal impact of the work performed.
        \item If the authors answer NA or No, they should explain why their work has no societal impact or why the paper does not address societal impact.
        \item Examples of negative societal impacts include potential malicious or unintended uses (e.g., disinformation, generating fake profiles, surveillance), fairness considerations (e.g., deployment of technologies that could make decisions that unfairly impact specific groups), privacy considerations, and security considerations.
        \item The conference expects that many papers will be foundational research and not tied to particular applications, let alone deployments. However, if there is a direct path to any negative applications, the authors should point it out. For example, it is legitimate to point out that an improvement in the quality of generative models could be used to generate deepfakes for disinformation. On the other hand, it is not needed to point out that a generic algorithm for optimizing neural networks could enable people to train models that generate Deepfakes faster.
        \item The authors should consider possible harms that could arise when the technology is being used as intended and functioning correctly, harms that could arise when the technology is being used as intended but gives incorrect results, and harms following from (intentional or unintentional) misuse of the technology.
        \item If there are negative societal impacts, the authors could also discuss possible mitigation strategies (e.g., gated release of models, providing defenses in addition to attacks, mechanisms for monitoring misuse, mechanisms to monitor how a system learns from feedback over time, improving the efficiency and accessibility of ML).
    \end{itemize}
    
\item {\bf Safeguards}
    \item[] Question: Does the paper describe safeguards that have been put in place for responsible release of data or models that have a high risk for misuse (e.g., pretrained language models, image generators, or scraped datasets)?
    \item[] Answer: \answerNA{} 
    \item[] Justification: The paper does not involve the release of high-risk models or sensitive datasets. It uses publicly available 3D scene datasets and does not introduce models or data with significant potential for misuse.
    \item[] Guidelines:
    \begin{itemize}
        \item The answer NA means that the paper poses no such risks.
        \item Released models that have a high risk for misuse or dual-use should be released with necessary safeguards to allow for controlled use of the model, for example by requiring that users adhere to usage guidelines or restrictions to access the model or implementing safety filters. 
        \item Datasets that have been scraped from the Internet could pose safety risks. The authors should describe how they avoided releasing unsafe images.
        \item We recognize that providing effective safeguards is challenging, and many papers do not require this, but we encourage authors to take this into account and make a best faith effort.
    \end{itemize}

\item {\bf Licenses for existing assets}
    \item[] Question: Are the creators or original owners of assets (e.g., code, data, models), used in the paper, properly credited and are the license and terms of use explicitly mentioned and properly respected?
    \item[] Answer: \answerYes{} 
    \item[] Justification: All external assets, including datasets and baseline models, are properly credited in the paper.
    \item[] Guidelines:
    \begin{itemize}
        \item The answer NA means that the paper does not use existing assets.
        \item The authors should cite the original paper that produced the code package or dataset.
        \item The authors should State which version of the asset is used and, if possible, include a URL.
        \item The name of the license (e.g., CC-BY 4.0) should be included for each asset.
        \item For scraped data from a particular source (e.g., website), the copyright and terms of service of that source should be provided.
        \item If assets are released, the license, copyright information, and terms of use in the package should be provided. For popular datasets, \url{paperswithcode.com/datasets} has curated licenses for some datasets. Their licensing guide can help determine the license of a dataset.
        \item For existing datasets that are re-packaged, both the original license and the license of the derived asset (if it has changed) should be provided.
        \item If this information is not available online, the authors are encouraged to reach out to the asset's creators.
    \end{itemize}

\item {\bf New assets}
    \item[] Question: Are new assets introduced in the paper well documented and is the documentation provided alongside the assets?
    \item[] Answer: \answerYes{} 
    \item[] Justification: The paper introduces new assets that are clearly documented. Details of our method, datasets, and implementation are provided in the main paper, and any additional information is included in the supplementary material.
    \item[] Guidelines:
    \begin{itemize}
        \item The answer NA means that the paper does not release new assets.
        \item Researchers should communicate the details of the dataset/code/model as part of their submissions via structured templates. This includes details about training, license, limitations, etc. 
        \item The paper should discuss whether and how consent was obtained from people whose asset is used.
        \item At submission time, remember to anonymize your assets (if applicable). You can either create an anonymized URL or include an anonymized zip file.
    \end{itemize}

\item {\bf Crowdsourcing and research with human subjects}
    \item[] Question: For crowdsourcing experiments and research with human subjects, does the paper include the full text of instructions given to participants and screenshots, if applicable, as well as details about compensation (if any)? 
    \item[] Answer: \answerNA{} 
    \item[] Justification: The paper does not involve crowdsourcing or research with human subjects, and therefore no participant instructions, screenshots, or compensation details are applicable.
    \item[] Guidelines:
    \begin{itemize}
        \item The answer NA means that the paper does not involve crowdsourcing nor research with human subjects.
        \item Including this information in the supplemental material is fine, but if the main contribution of the paper involves human subjects, then as much detail as possible should be included in the main paper. 
        \item According to the NeurIPS Code of Ethics, workers involved in data collection, curation, or other labor should be paid at least the minimum wage in the country of the data collector. 
    \end{itemize}

\item {\bf Institutional review board (IRB) approvals or equivalent for research with human subjects}
    \item[] Question: Does the paper describe potential risks incurred by study participants, whether such risks were disclosed to the subjects, and whether Institutional Review Board (IRB) approvals (or an equivalent approval/review based on the requirements of your country or institution) were obtained?
    \item[] Answer: \answerNA{} 
    \item[] Justification: The paper does not involve crowdsourcing or research with human subjects.
    \item[] Guidelines:
    \begin{itemize}
        \item The answer NA means that the paper does not involve crowdsourcing nor research with human subjects.
        \item Depending on the country in which research is conducted, IRB approval (or equivalent) may be required for any human subjects research. If you obtained IRB approval, you should clearly State this in the paper. 
        \item We recognize that the procedures for this may vary significantly between institutions and locations, and we expect authors to adhere to the NeurIPS Code of Ethics and the guidelines for their institution. 
        \item For initial submissions, do not include any information that would break anonymity (if applicable), such as the institution conducting the review.
    \end{itemize}

\item {\bf Declaration of LLM usage}
    \item[] Question: Does the paper describe the usage of LLMs if it is an important, original, or non-standard component of the core methods in this research? Note that if the LLM is used only for writing, editing, or formatting purposes and does not impact the core methodology, scientific rigorousness, or originality of the research, declaration is not required.
    \item[] Answer: \answerNA{} 
    \item[] Justification: Our work focuses on computer vision, and LLMs are not used in our method.
    \item[] Guidelines:
    \begin{itemize}
        \item The answer NA means that the core method development in this research does not involve LLMs as any important, original, or non-standard components.
        \item Please refer to our LLM policy (\url{https://neurips.cc/Conferences/2025/LLM}) for what should or should not be described.
    \end{itemize}

\end{enumerate}

\clearpage
\appendix
\section{Random nature of artifacts}

\begin{figure}[t]
\centering
\begin{minipage}[b]{0.195\linewidth}
    \centering
    {\includegraphics[width=\linewidth]{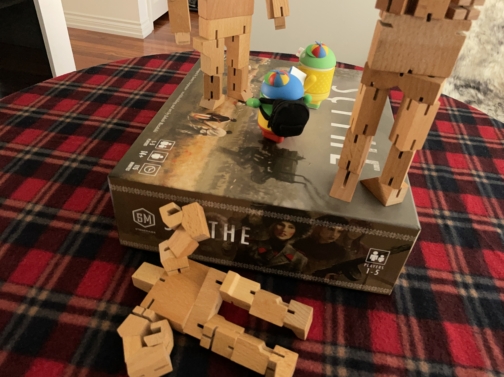}}
    {\includegraphics[width=\linewidth]{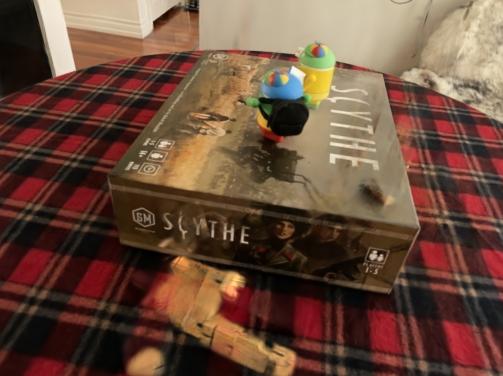}}
    {\includegraphics[width=\linewidth]{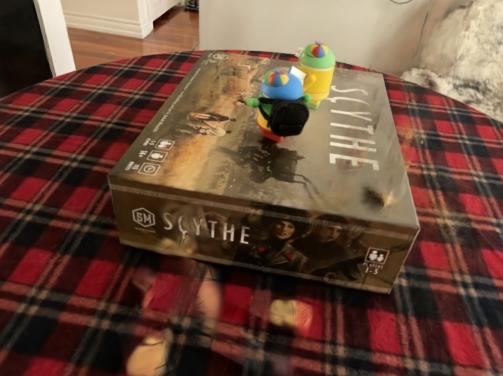}}
    {\includegraphics[width=\linewidth]{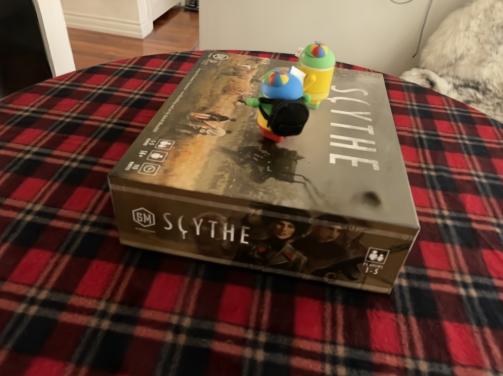}}
\end{minipage}
\hfill
\begin{minipage}[b]{0.195\linewidth}
    \centering
    {\includegraphics[width=\linewidth]{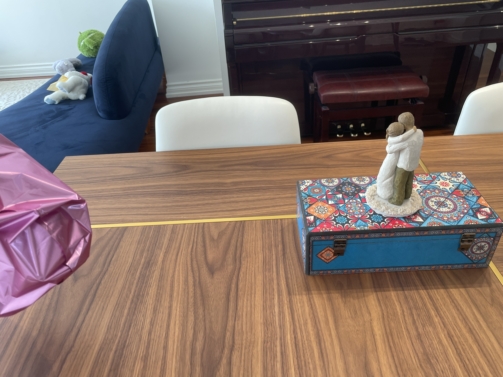}}
    {\includegraphics[width=\linewidth]{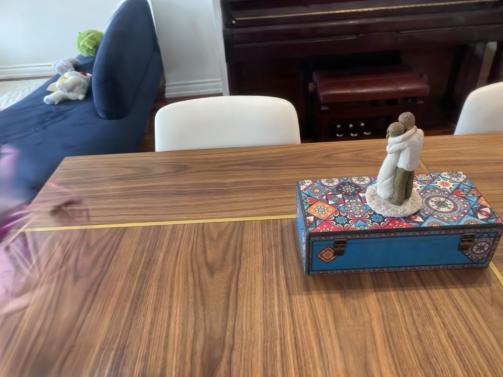}}
    {\includegraphics[width=\linewidth]{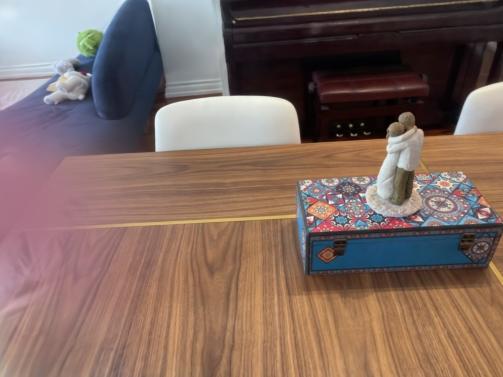}}
    {\includegraphics[width=\linewidth]{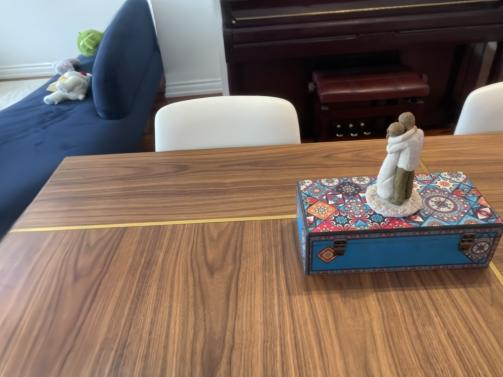}}
\end{minipage}
\hfill
\begin{minipage}[b]{0.195\linewidth}
    \centering
    {\includegraphics[width=\linewidth]{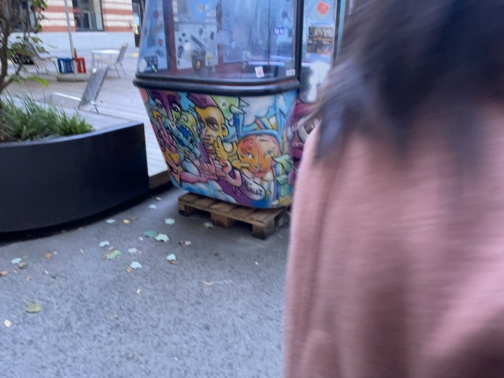}}
    {\includegraphics[width=\linewidth]{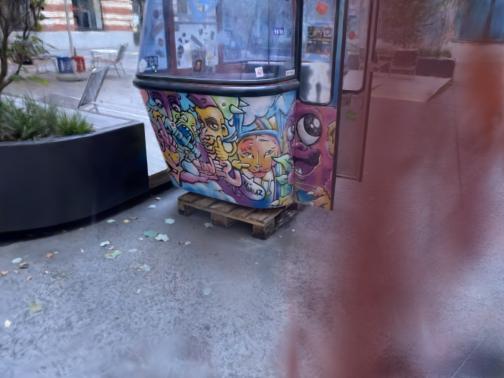}}
    {\includegraphics[width=\linewidth]{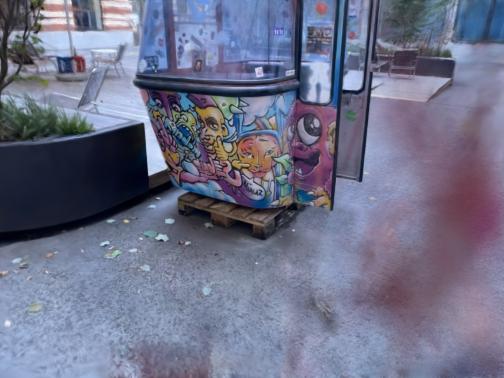}}
    {\includegraphics[width=\linewidth]{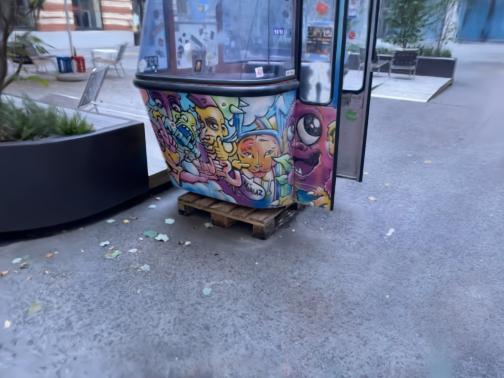}}
\end{minipage}
\hfill
\begin{minipage}[b]{0.195\linewidth}
    \centering
    {\includegraphics[width=\linewidth]{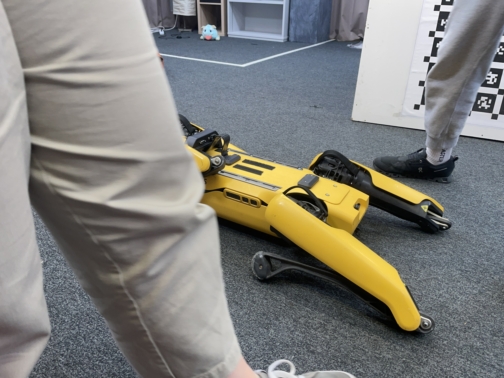}}
    {\includegraphics[width=\linewidth]{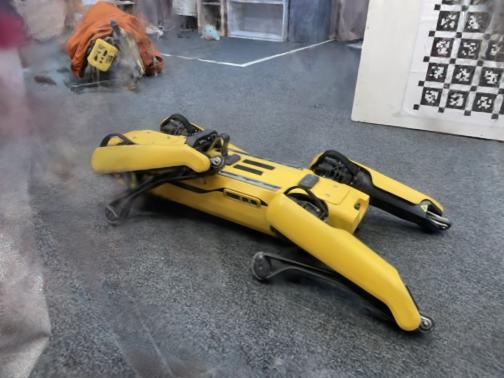}}
    {\includegraphics[width=\linewidth]{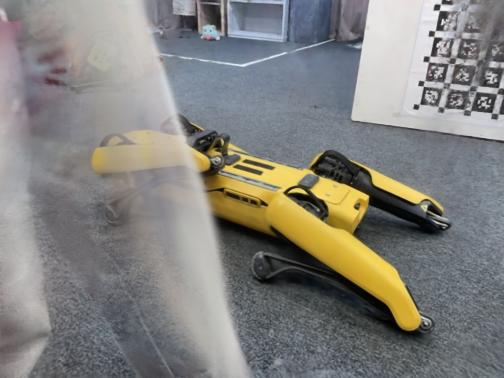}}
    {\includegraphics[width=\linewidth]{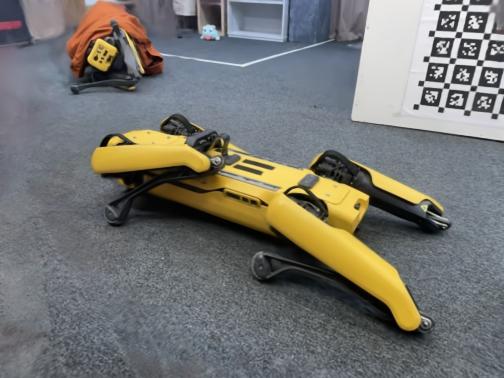}}
\end{minipage}
\hfill
\begin{minipage}[b]{0.195\linewidth}
    \centering
    {\includegraphics[width=\linewidth]{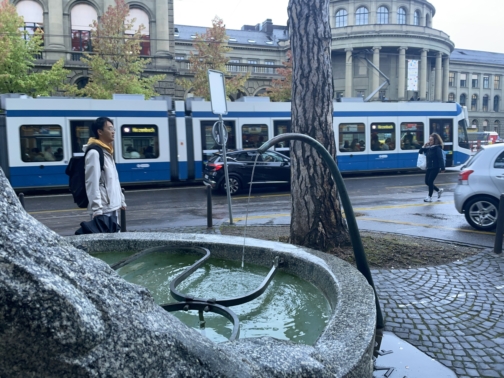}}
    {\includegraphics[width=\linewidth]{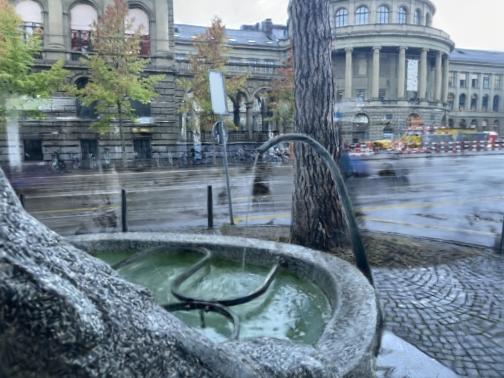}}
    {\includegraphics[width=\linewidth]{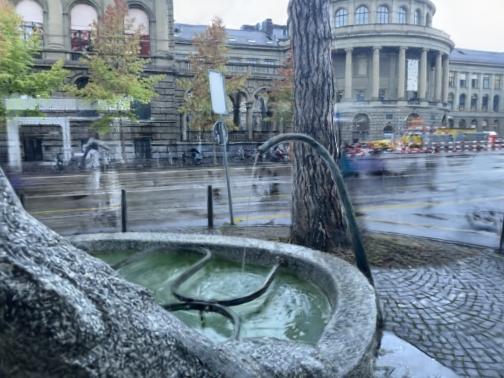}}
    {\includegraphics[width=\linewidth]{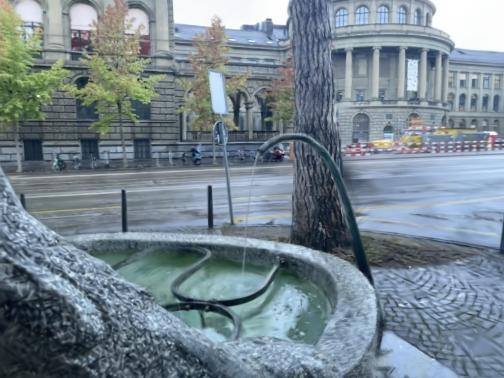}}
\end{minipage}
\hfill
\vspace*{-5mm}
\caption{Randomness of artifacts across training runs. Row 1 shows the target view (with the presence of distractors); Rows 2 and 3 present results from two independent runs of Mip-Splatting; Row 4 shows the result of our method with the mutual consistency regularization.}
\label{fig:artifacts}
\end{figure}

As shown in Figure \ref{fig:figure1} and \ref{fig:artifacts}, different runs of 3DGS on the same scene (with only the view order randomized) result in different artifacts, particularly in uncertain regions. The mutual consistency loss helps suppress these artifacts in both models. On one hand, the shared static regions remain consistent and act as a strong regularizer. On the other hand, the differing artifacts in uncertain areas provide complementary supervision signals, allowing regions affected by artifacts in one model to be recovered by the other.

For the effect of distinct masking strategies, Table \ref{tab:ablation} in the main paper presents a quantitative comparison. The performance degrades when both models use the same mask, for both GS-GS and EMA-GS settings. This supports our claim that using separate masks helps prevent convergence to the same erroneous reconstruction patterns.

\section{Comparison with HybridGS}
Our Asymmetric Dual 3DGS framework differs fundamentally from HybridGS \cite{lin_hybridgs_2025} in both design and training. HybridGS separates static and dynamic content using two models (3DGS for static, 2DGS for dynamic) and requires a staged training process with a learnable blending mask. In contrast, our method uses a dual 3DGS setup with mutual supervision to improve robustness against dynamic noise, all within the standard 3DGS training pipeline. While both methods use masking, HybridGS blends outputs based on 2DGS-derived uncertainty, whereas we apply two distinct masking strategies to reduce confirmation bias from a single, potentially inaccurate mask.

\section{Multi-cue adaptive mask}
Some prior works~\cite{sabour_robustnerf_2023, ren_nerfonthego_2024} use residuals between ground-truth and rendered images to detect distractors, assuming static regions are learned first. However, this can misclassify object boundaries and miss distractors resembling the background. Others~\cite{martin-brualla_nerfw_2021, Rematas_urban_2022} use pretrained semantic segmentation to mask known distractors, such as people or sky, but these methods rely on task-specific priors and lack generality across diverse scenes. We propose Multi-Cue Adaptive Masking to combine the strengths of residual-based and segmentation-based methods, while also providing a complementary hard mask that captures distinct error patterns compared to the self-supervised soft mask.

\begin{algorithm}
\caption{Multi-Cue Adaptive Masking}
\label{alg:MAM}
\begin{algorithmic}[1]
\Require Rendered image $\tilde{\mathbf{I}}$, ground-truth image $\mathbf{I}$, semantic masks $\{\mathbf{M}_k\}$ from SAM, stereo correspondence map $\mathbf{S}_{\ge 3}$ from COLMAP
\State $\mathbf{E}_{\text{pix}} = \|\tilde{\mathbf{I}} - \mathbf{I}\|_1$  \Comment{Pixel-level residual}
\State $\mathbf{F} = \text{DINOv2} (\mathbf{I}); \tilde{\mathbf{F}} = \text{DINOv2} (\tilde{\mathbf{I}})$ \Comment{DINOv2 features}
\State $\mathbf{E}_{\text{feat}} = 1 - \text{CosineSimilarity}(\tilde{\mathbf{F}}, \mathbf{F})$ \Comment{Feature-level residual}
\State $\bar{e}_{\text{pix}} = \sum \mathbf{E}_{\text{pix}} / \text{Area}(\mathbf{I}) $ \Comment{Average residuals over $\mathbf{I}$}
\State $\bar{e}_{\text{feat}} = \sum \mathbf{E}_{\text{feat}} / \text{Area}(\mathbf{I})$ 
\State $\bar{s} = \sum \mathbf{S}_{\ge 3} / \text{Area}(\mathbf{I})$ \Comment{Stereo correspondence density over $\mathbf{I}$}
\For{each mask $\mathbf{M}_k$}
    \State $e_{\text{pix},k} = \sum \mathbf{M}_k \odot \mathbf{E}_{\text{pix}} / \sum \mathbf{M}_k$ \Comment{Average residuals over $\mathbf{M}_k$}
    \State $e_{\text{feat},k} = \sum \mathbf{M}_k \odot \mathbf{E}_{\text{feat}} / \sum \mathbf{M}_k$
    \State $s = \sum \mathbf{M}_k \odot \mathbf{S}_{\ge 3} / \sum \mathbf{M}_k$ \Comment{Stereo correspondence density over $\mathbf{M}_k$}
    \If{$e_{\text{pix},k} > \bar{e}_{\text{pix}}$ and $e_{\text{feat},k} > \bar{e}_{\text{feat}}$ and $s < 0.1 \cdot \bar{s}$}
        \State Mark $\mathbf{M}_k$ as a distractor mask
    \EndIf
\EndFor
\State \Return $\mathbf{M}_h = 1 - \bigcup \{\mathbf{M}_k\}_{\text{selected}}$ \Comment{0 for distractor}
\end{algorithmic}
\end{algorithm}

Here, the stereo-based correspondence records the number of matches each pixel in the given image has, based on SIFT feature correspondences proposed in COLMAP \cite{schonberger2016structure}. A pixel is considered a valid correspondence (with the stereo correspondence map value set to true at the pixel location) if its match count exceeds a threshold, indicating it likely belongs to a static region. In contrast, distractors typically yield fewer matches due to their limited presence across images. In Algorithm \ref{alg:MAM}, $\mathbf{S}_{\ge 3}$ denotes the stereo correspondence map, where a pixel is considered a valid correspondence if it has more than three matches.

\section{Datasets and metrics}
We evaluate our method on three in-the-wild datasets with varying challenges, as shown in Table \ref{tab:datasets}. NeRF On-the-go dataset~\cite{ren_nerfonthego_2024} features indoor and outdoor sequences with consistent appearance but varying distractor ratios (5\%–30\%). RobustNeRF dataset~\cite{sabour_robustnerf_2023} provides indoor scenes with static geometry and controlled distractor placement (from single-type to 150 varied distractors), where training is done on cluttered views and testing on clean, unseen ones. We use the undistorted versions of these datasets, following the protocols of WildGaussian~\cite{kulhanek_wildgaussians_2024} and HybridGS~\cite{lin_hybridgs_2025}. The PhotoTourism dataset~\cite{jin_phototourism_2020} includes landmark scenes (Brandenburg Gate, Sacre Coeur, Trevi Fountain) captured under diverse lighting, weather, and viewpoints, with both significant appearance variation and real-world distractors. We report PSNR, SSIM, and LPIPS~\cite{zhang_lpips_2018} to assess reconstruction accuracy and perceptual quality.

\begin{table}[ht]
\centering
\caption{In-the-wild 3D reconstruction datasets.}
\label{tab:datasets}
\begin{tabular}{llllll}
\toprule
\multicolumn{1}{l}{Dataset}     & Scene            & \# Train & \# Test & Distractor     & Appear. change \\
\midrule
\multirow{6}{*}{NeRF On-the-go \cite{ren_nerfonthego_2024}} & Patio-high       & 222      & 45      & $\sim$30\%     & No                \\
                                & Spot             & 168      & 10      & $\sim$30\%     & No                \\
                                & Patio            & 98       & 26      & 15\%$\sim$20\% & No                \\
                                & Corner           & 101      & 20      & 15\%$\sim$20\% & No                \\
                                & Fountain         & 168      & 17      & 5\%$\sim$10\%  & No                \\
                                & Mountain         & 119      & 12      & 5\%$\sim$10\%  & No                \\
\midrule
\multirow{4}{*}{RobustNeRF \cite{sabour_robustnerf_2023}}     & Statue           & 255      & 19      & 1 type         & No                \\
                                & Android          & 122      & 19      & 1 type         & No                \\
                                & Yoda             & 109      & 202     & 100 types      & No                \\
                                & Crab             & 109      & 194     & 150 types      & No                \\
\midrule
\multirow{3}{*}{PhotoTourism \cite{jin_phototourism_2020}}   & Brandenburg Gate & 763      & 10      & $\sim$3.5\%          & Yes               \\
                                & Sacre Coeur      & 830      & 21      & $\sim$3.5\%          & Yes               \\
                                & Trevi Fountain   & 1689     & 19      & $\sim$3.5\%          & Yes    \\
\bottomrule
\end{tabular}
\end{table}

\section{Implementation details}
\label{sec:implement}
Our base model is built on Mip-Splatting~\cite{yu_mip-splatting_2024}. Following its default settings, we recompute the sampling rate of each Gaussian every 100 iterations, with a 2D Mip filter variance of 0.1 and a 3D smoothing filter variance of 0.2. We train for 30{,}000 iterations on NeRF On-the-go and RobustNeRF, with densification and pruning every 1{,}000 steps until iteration 15{,}000; and for 100{,}000 iterations on PhotoTourism, with densification and pruning every 1{,}000 steps until iteration 50{,}000. We omit the opacity reset and apply a 1{,}000-step warm-up before the mutual consistency regularization begins. The consistency regularization weight is set to 0.1. The learnable mask is optimized by a loss weighted $\lambda_\text{mask}=1.0$ with a learning rate of 0.1. For EMA, we use a smoothing factor of $\beta=0.8$. Semantic regions for the multi-cue adaptive mask are generated using Semantic SAM~\cite{li_segmentsam_2025} to create instance-level segmentations and apply Algorithm~\ref{alg:MAM} to select distractor regions as masks.

Additionally, we use a 32-dimensional per-view appearance embedding and a 24-dimensional per-Gaussian embedding. Color transformation is performed using a three-layer MLP with hidden size 128, outputting a scale and bias for each RGB channel. The learning rates are set to 0.001 for the per-view embedding, 0.005 for the per-Gaussian embedding, and 0.0005 for the MLP. The other 3DGS-related hyperparameters follow the setup from origin work shown in Table \ref{tab:3dgshp}. 

\begin{table}[ht]
\centering
\caption{The other 3DGS-related hyperparameters.}
\label{tab:3dgshp}
\begin{tabular}{ll}
\toprule
Parameter                 & Value            \\
\midrule
position\_lr\_init        & 0.00016          \\
position\_lr\_final       & 0.0000016        \\
position\_lr\_delay\_mult & 0.01             \\
feature\_lr               & 0.0025           \\
opacity\_lr               & 0.1              \\
scaling\_lr               & 0.005            \\
rotation\_lr              & 0.001            \\
percent\_dense            & 0.01             \\
lambda\_dssim             & 0.2              \\
densification\_interval   & 1000             \\
opacity\_reset\_interval  & No opacity reset \\
densify\_from\_iter       & 500              \\
densify\_grad\_threshold  & 0.0002 \\
\bottomrule
\end{tabular}
\end{table}

\begin{table}[ht]
\centering
\caption{The code repo and licenses.}
\label{tab:codes}
\resizebox{\linewidth}{!}{
\begin{tabular}{lll}
\toprule
Method & Link & License \\
\midrule
3DGS \cite{kerbl_3d_2023} & \url{https://github.com/graphdeco-inria/gaussian-splatting} & \href{https://github.com/graphdeco-inria/gaussian-splatting/blob/main/LICENSE.md}{Custom} \\
Mip-Splatting \cite{yu_mip-splatting_2024} & \url{https://github.com/autonomousvision/mip-splatting} & \href{https://raw.githubusercontent.com/autonomousvision/mip-splatting/refs/heads/main/LICENSE.md}{Custom} \\
WildGaussians \cite{kulhanek_wildgaussians_2024} & \url{https://github.com/jkulhanek/wild-gaussians/} & MIT License \\
NerfBaselines \cite{kulhanek_nerfbaselines_2024} & \url{https://github.com/nerfbaselines/nerfbaselines} & MIT License \\
COLMAP \cite{schonberger2016structure} & \url{https://github.com/colmap/colmap} & BSD License \\
Semantic-SAM~\cite{li_segmentsam_2025} & \url{https://github.com/UX-Decoder/Semantic-SAM} & Apache 2.0 License \\
NeRF On-the-go dataset \cite{ren_nerfonthego_2024}  & \url{https://github.com/cvg/nerf-on-the-go} & Apache 2.0 License \\
RobustNeRF dataset \cite{sabour_robustnerf_2023} & \url{https://robustnerf.github.io/} & Custom \\
PhotoTourism dataset \cite{jin_phototourism_2020} & \url{https://github.com/ubc-vision/image-matching-benchmark} & Apache 2.0 License \\
\bottomrule
\end{tabular}
}
\end{table}

\section{More results}

\begin{table}[t]
\caption{Quantitative results on the NeRF On-the-go dataset~\cite{ren_nerfonthego_2024}. The best and second-best results are highlighted in \textbf{bold} and \underline{underline}, respectively.}
\label{tab:onthego_full}
\centering
\resizebox{\linewidth}{!}{
\begin{tabular}{lllllllllllllllllll}
\toprule
Scene & \multicolumn{3}{c}{Mountain} & \multicolumn{3}{c}{Fountain} & \multicolumn{3}{c}{Corner} & \multicolumn{3}{c}{Patio} & \multicolumn{3}{c}{Spot} & \multicolumn{3}{c}{Patio-High} \\
\midrule
Method & PSNR$\uparrow$ & SSIM$\uparrow$ & LPIPS$\downarrow$ & PSNR$\uparrow$ & SSIM$\uparrow$ & LPIPS$\downarrow$ & PSNR$\uparrow$ & SSIM$\uparrow$ & LPIPS$\downarrow$ & PSNR$\uparrow$ & SSIM$\uparrow$ & LPIPS$\downarrow$ & PSNR$\uparrow$ & SSIM$\uparrow$ & LPIPS$\downarrow$ & PSNR$\uparrow$ & SSIM$\uparrow$ & LPIPS$\downarrow$ \\
\midrule
RobustNeRF \cite{sabour_robustnerf_2023} & 17.54 & 0.496 & 0.383 & 15.65 & 0.318 & 0.576 & 23.04 & 0.764 & 0.244 & 20.39 & 0.718 & 0.251 & 20.65 & 0.625 & 0.391 & 20.54 & 0.578 & 0.366 \\
NeRF On-the-go \cite{ren_nerfonthego_2024} & 20.15 & 0.644 & 0.259 & 20.11 & 0.609 & 0.314 & 24.22 & 0.806 & 0.190 & 20.78 & 0.754 & 0.219 & 23.33 & 0.787 & 0.189 & 21.41 & 0.718 & 0.235 \\
3DGS \cite{kerbl_3d_2023} & 19.40 & 0.638 & 0.213 & 19.96 & 0.659 & 0.185 & 20.90 & 0.713 & 0.241 & 17.48 & 0.704 & 0.199 & 20.77 & 0.693 & 0.316 & 17.29 & 0.604 & 0.363 \\
Mip-Splatting \cite{yu_mip-splatting_2024}  & 19.86 & 0.649 & 0.200 & 20.19 & 0.672 & 0.189 & 21.15 & 0.728 & 0.230 & 18.31 & 0.639 & 0.328 & 20.18 & 0.689 & 0.338 & 18.31 & 0.639 & 0.328 \\
WildGaussian \cite{kulhanek_wildgaussians_2024} & 20.43 & 0.653 & 0.255 & 20.81 & 0.662 & 0.215 & 24.16 & 0.822 & \textbf{0.045} & 21.44 & 0.800 & 0.138 & 23.82 & 0.816 & \underline{0.138} & 22.23 & 0.725 & 0.206 \\
SLS-mlp \cite{sabour_spotlesssplats_2024} & 19.84 & 0.580 & 0.294 & 20.19 & 0.612 & 0.258 & 24.03 & 0.795 & 0.258 & 21.55 & 0.838 & \textbf{0.065} & 23.52 & 0.756 & 0.185 & 20.31 & 0.664 & 0.259 \\
HybridGS \cite{lin_hybridgs_2025} & 21.73 & 0.693 & 0.284 & 21.11 & 0.674 & 0.252 & 25.03 & 0.847 & 0.151 & 21.98 & 0.812 & 0.169 & 24.33 & 0.794 & 0.196 & 21.77 & 0.741 & 0.211 \\
\midrule
Ours (GS-GS) & \textbf{22.00} & \textbf{0.740} & \underline{0.199} & \textbf{21.83} & \textbf{0.717} & \underline{0.180} & \textbf{26.15} & \textbf{0.885} & \underline{0.085} & \textbf{22.97} & \textbf{0.860} & 0.096 & \textbf{25.52} & \textbf{0.854} & \textbf{0.135} & \textbf{23.17} & \underline{0.796} & \underline{0.164} \\
Ours (EMA-GS) & \underline{21.93} & \underline{0.735} & \textbf{0.162} & \underline{21.61} & \underline{0.709} & \textbf{0.162} & \underline{25.77} & \underline{0.876} & 0.089 & \underline{22.87} & \underline{0.853} & \underline{0.091} & \underline{25.09} & \underline{0.839} & 0.152 & \underline{23.14} & \textbf{0.797} & \textbf{0.156} \\
\bottomrule
\end{tabular}
}
\vspace{-2mm}
\end{table}

\subsection{NeRF On-the-go and RobustNeRF}
\begin{figure}[t]
\centering
\begin{minipage}[b]{0.195\linewidth}
    \centering
    {\includegraphics[width=\linewidth]{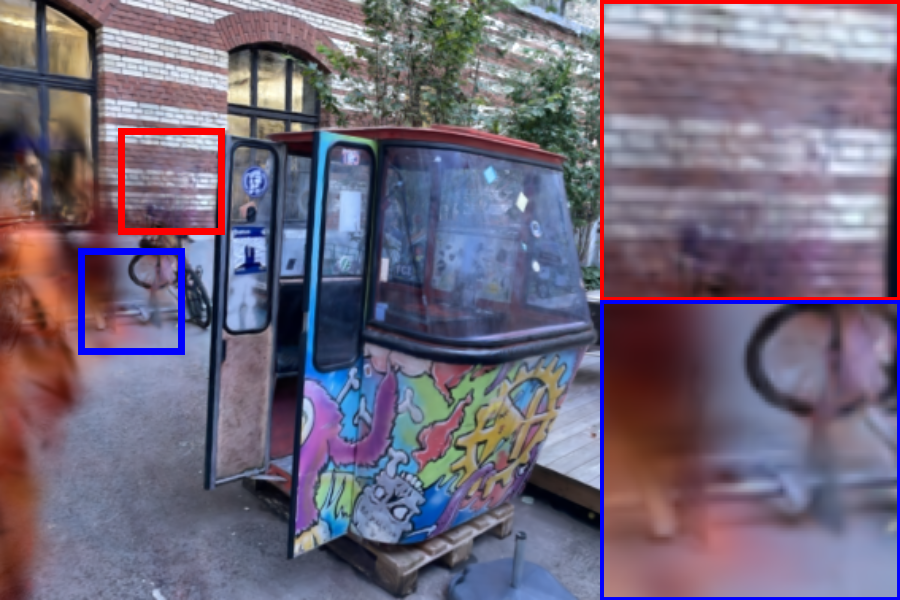}}
    {\includegraphics[width=\linewidth]{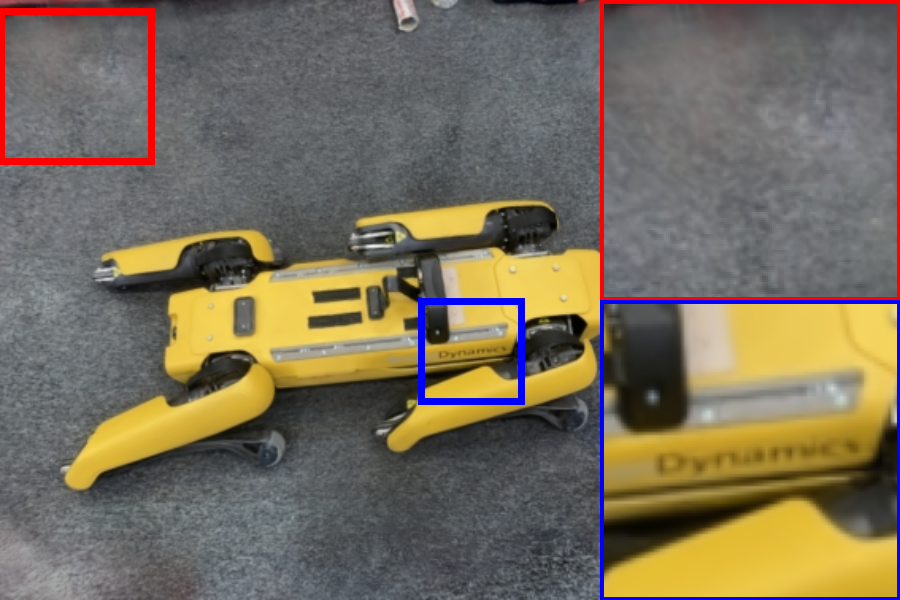}}
    {\includegraphics[width=\linewidth]{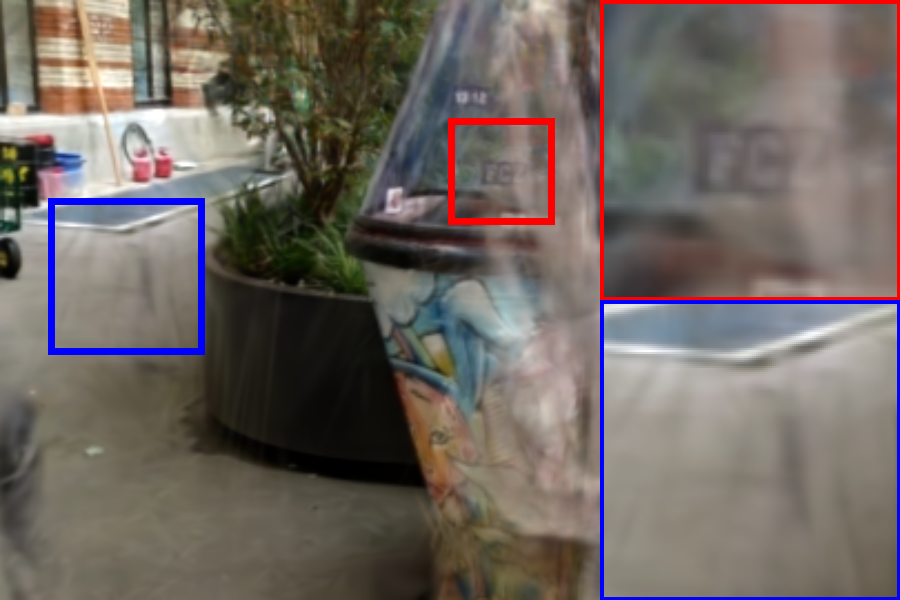}}
    {\includegraphics[width=\linewidth]{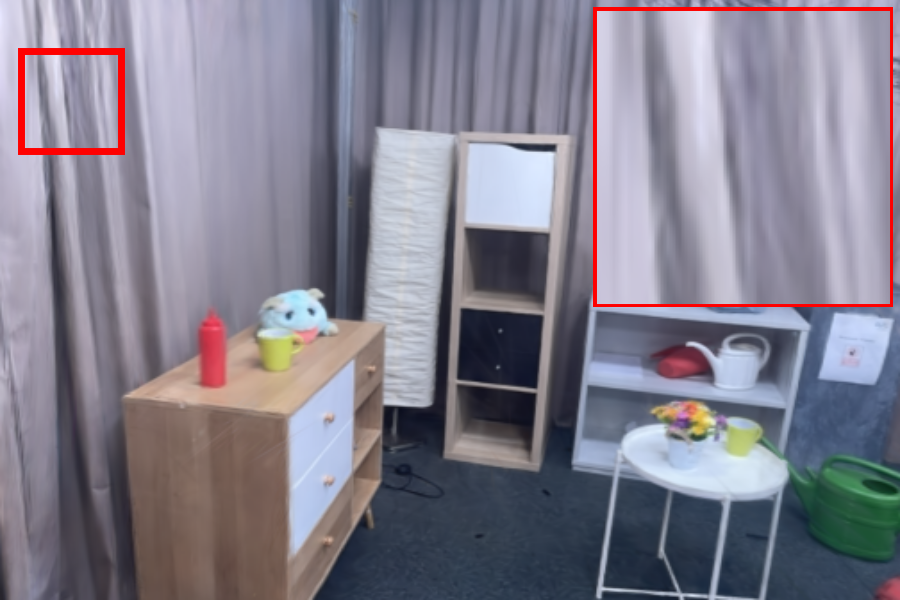}}
    {\includegraphics[width=\linewidth]{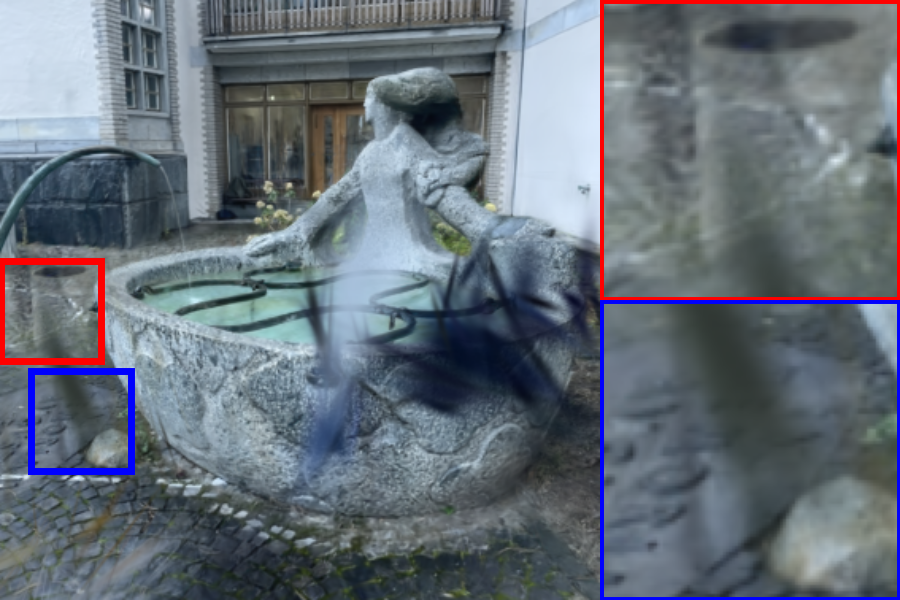}}
    {\includegraphics[width=\linewidth]{images/IMG_7940_mip_zoomed.png}}
    \centerline{Mip-Splatting \cite{yu_mip-splatting_2024}}
\end{minipage}
\hfill
\begin{minipage}[b]{0.195\linewidth}
    \centering    
    {\includegraphics[width=\linewidth]{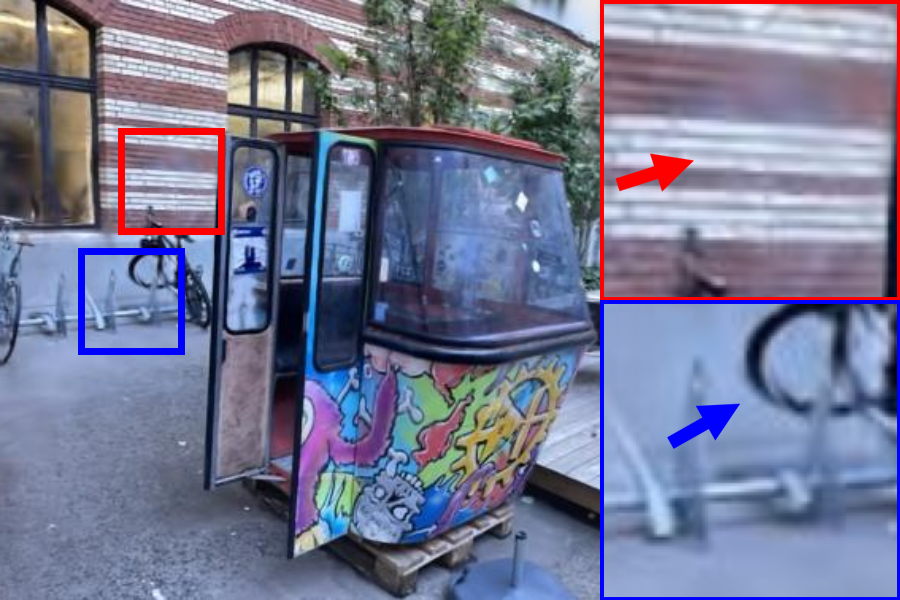}}
    {\includegraphics[width=\linewidth]{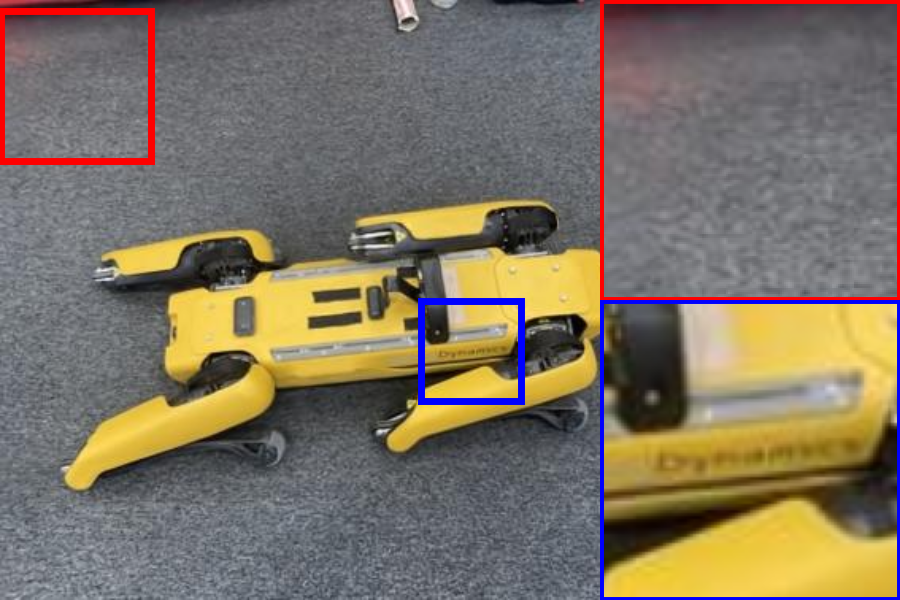}}
    {\includegraphics[width=\linewidth]{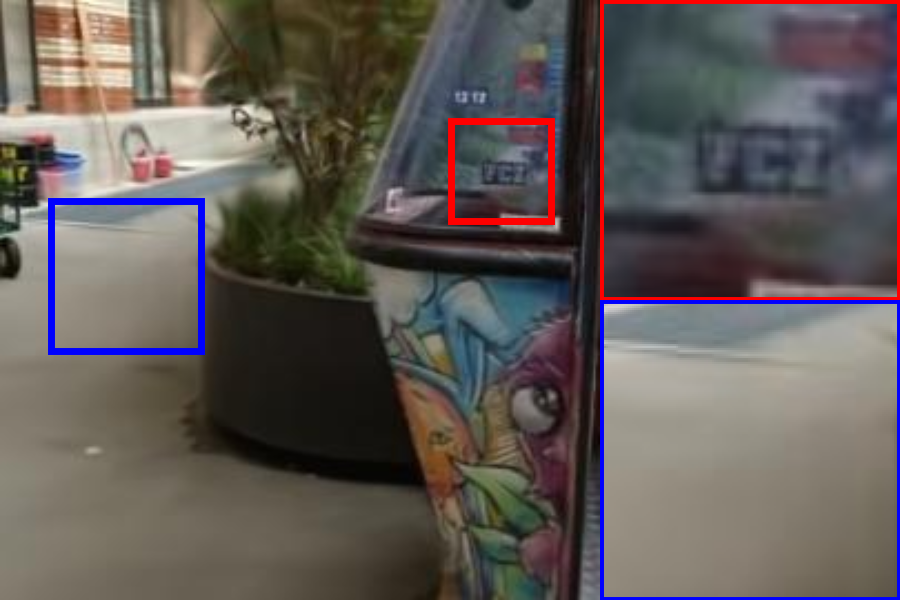}}
    {\includegraphics[width=\linewidth]{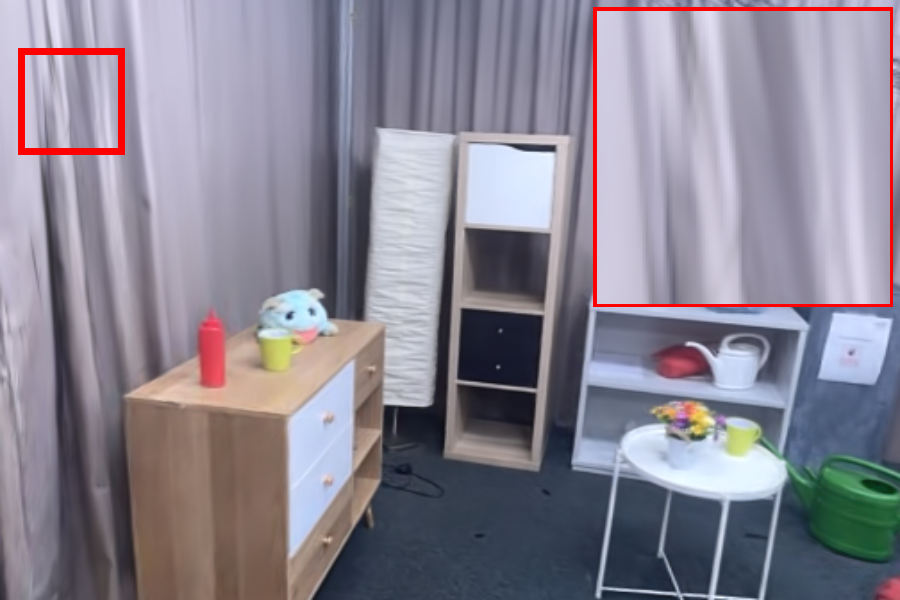}}
    {\includegraphics[width=\linewidth]{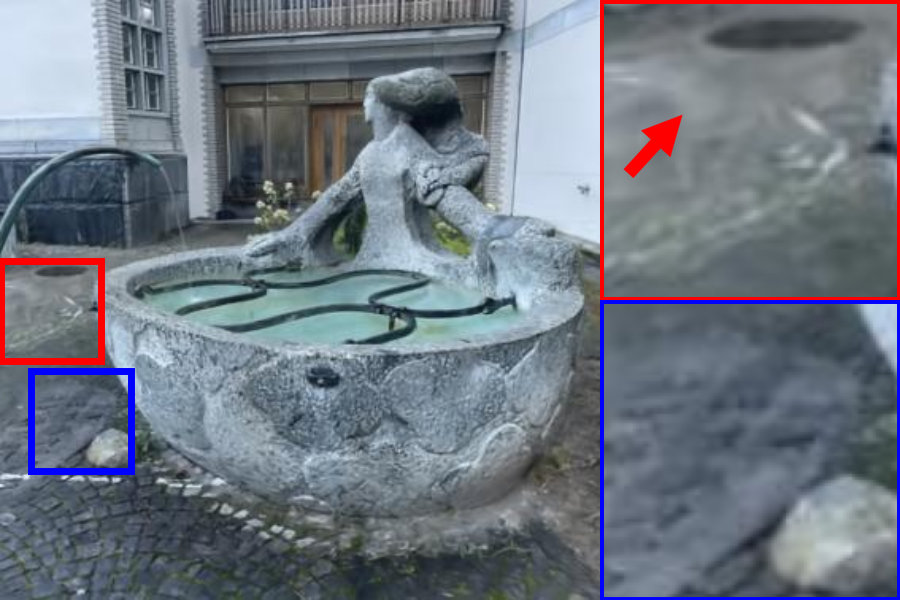}}
    {\includegraphics[width=\linewidth]{images/IMG_7940_hybridgs_zoomed.png}}
    \centerline{HybridGS \cite{lin_hybridgs_2025}}
\end{minipage}
\hfill
\begin{minipage}[b]{0.195\linewidth}
    \centering
    {\includegraphics[width=\linewidth]{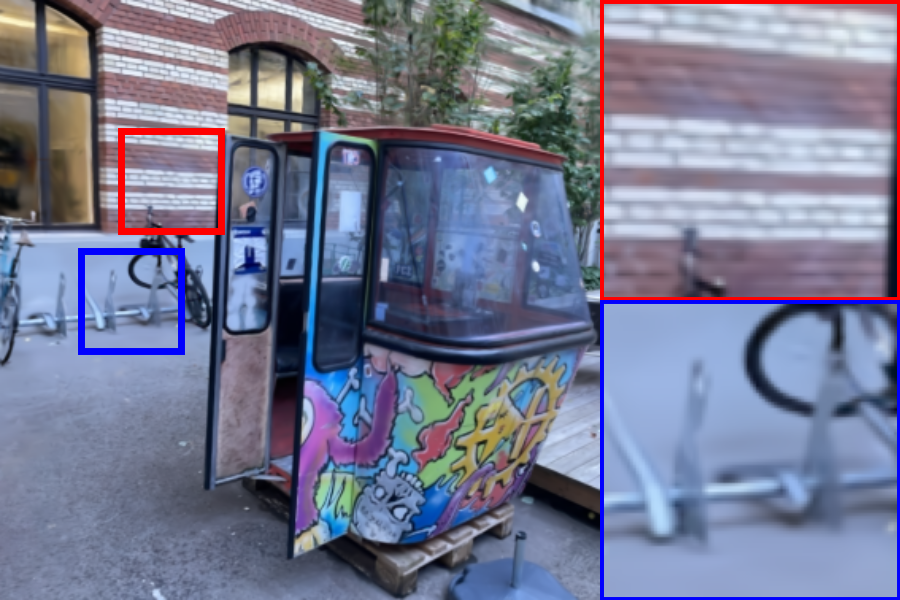}}
    {\includegraphics[width=\linewidth]{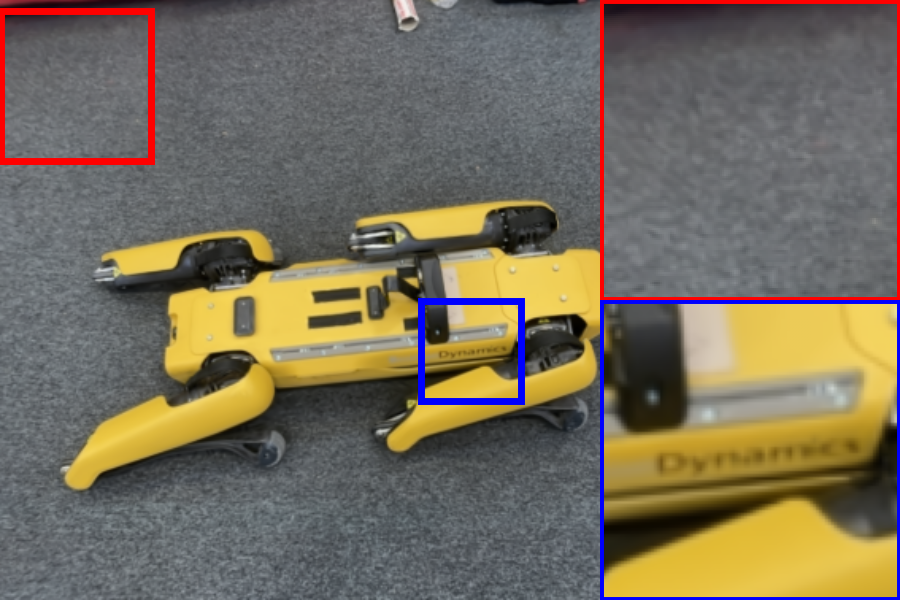}}
    {\includegraphics[width=\linewidth]{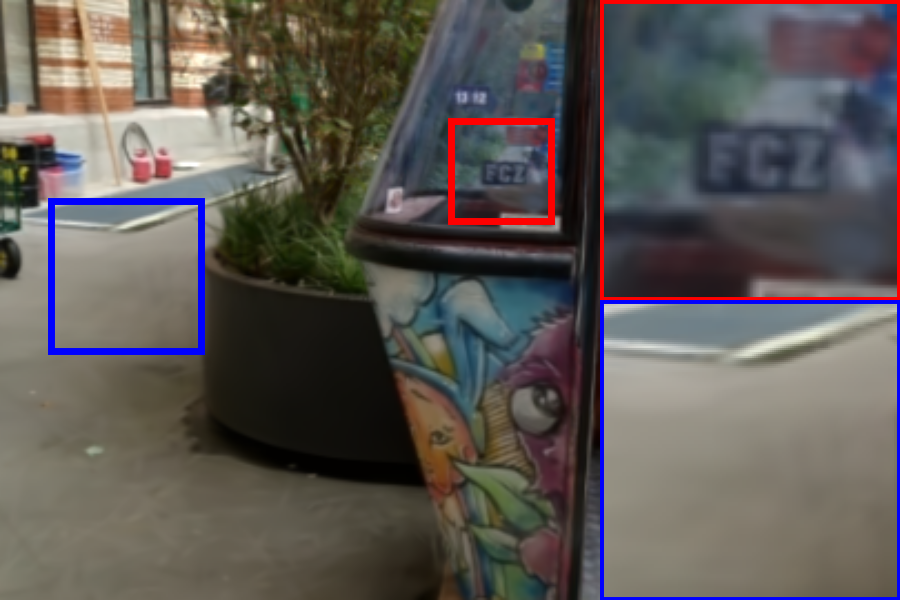}}
    {\includegraphics[width=\linewidth]{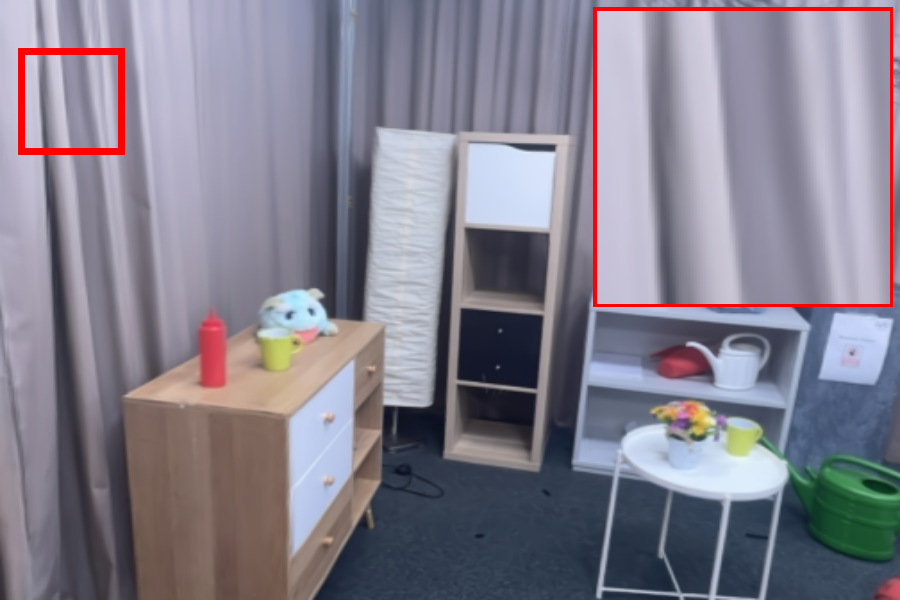}}
    {\includegraphics[width=\linewidth]{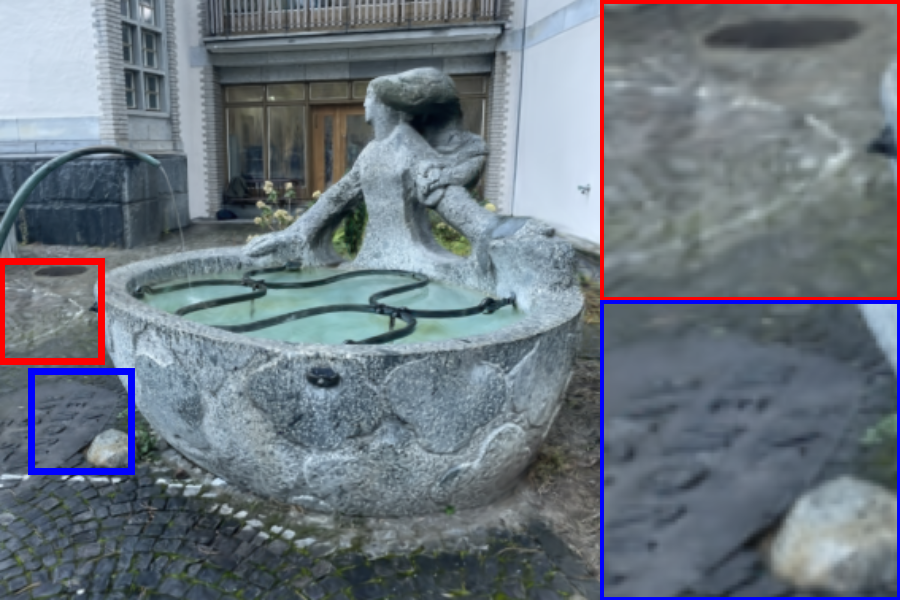}}
    {\includegraphics[width=\linewidth]{images/IMG_7940_ema_zoomed.png}}
    \centerline{Our (EMA-GS)}
\end{minipage}
\hfill
\begin{minipage}[b]{0.195\linewidth}
    \centering
    {\includegraphics[width=\linewidth]{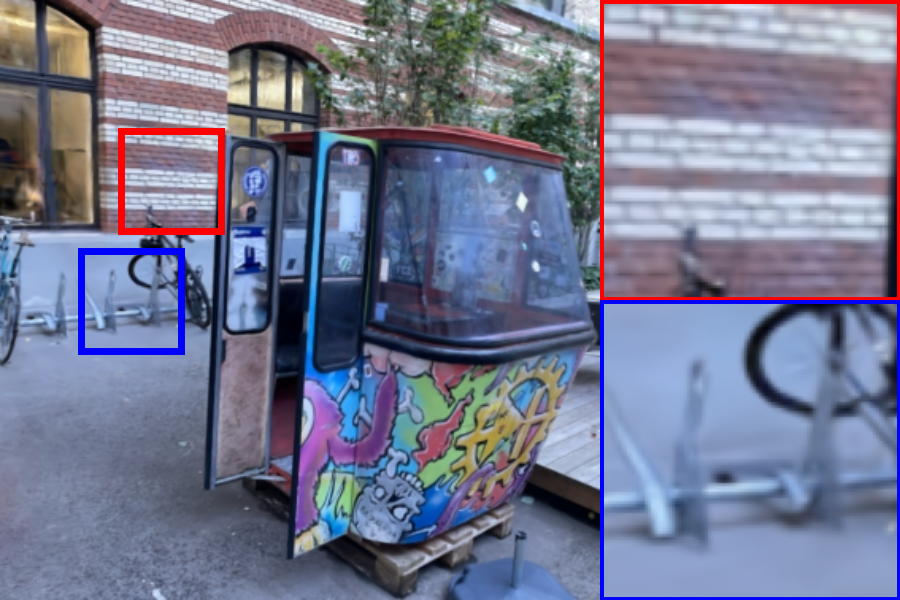}}
    {\includegraphics[width=\linewidth]{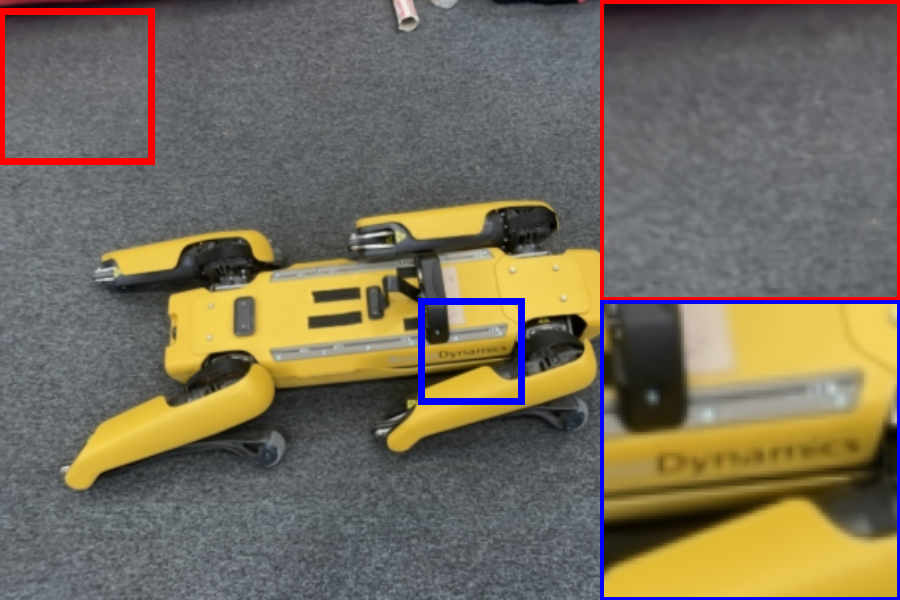}}
    {\includegraphics[width=\linewidth]{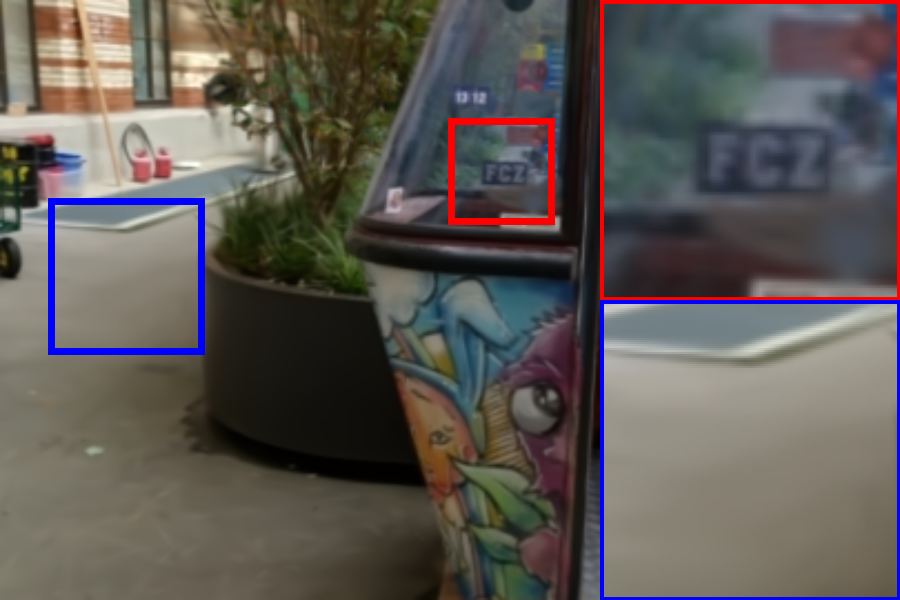}}
    {\includegraphics[width=\linewidth]{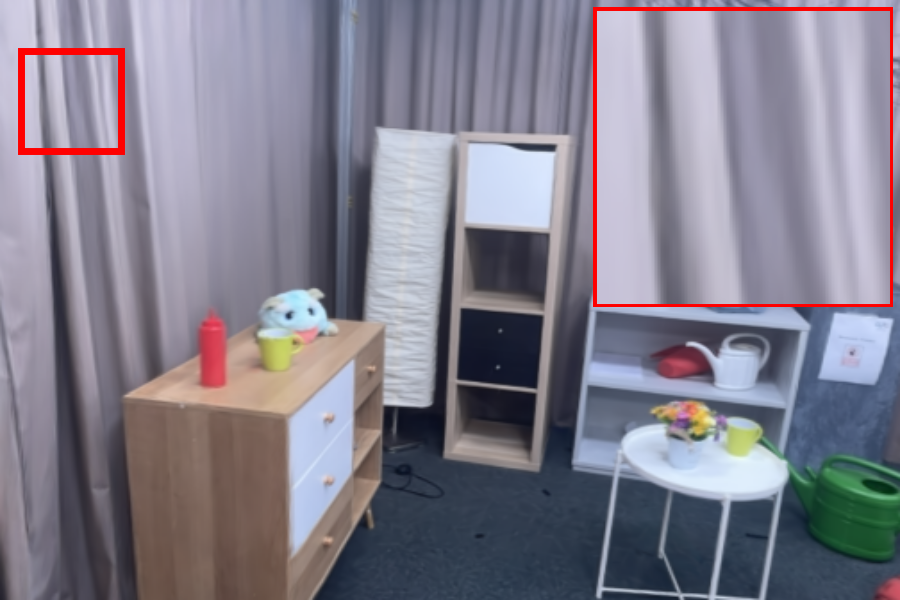}}
    {\includegraphics[width=\linewidth]{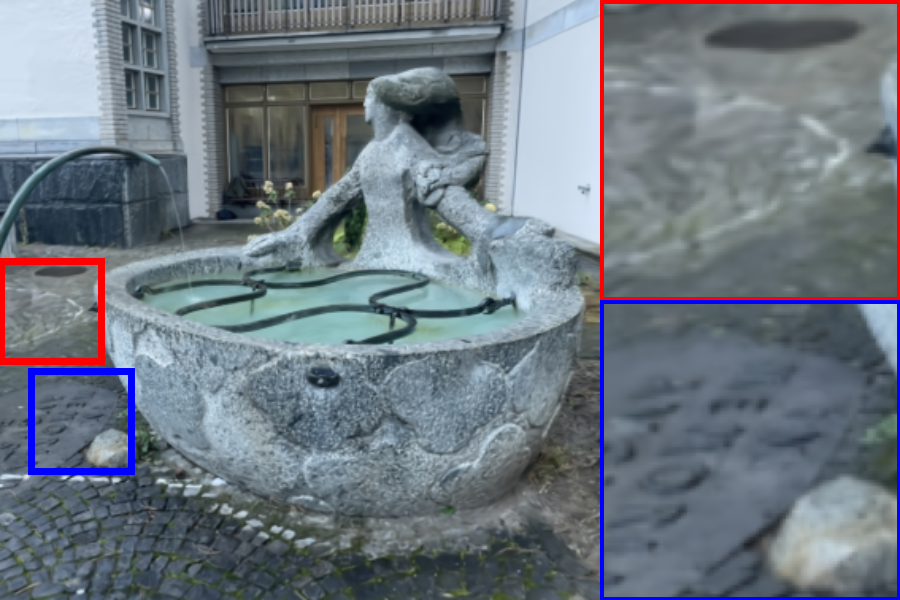}}
    {\includegraphics[width=\linewidth]{images/IMG_7940_dual_zoomed.png}}
    \centerline{Ours (GS-GS)}
\end{minipage}
\hfill
\begin{minipage}[b]{0.195\linewidth}
    \centering
    {\includegraphics[width=\linewidth]{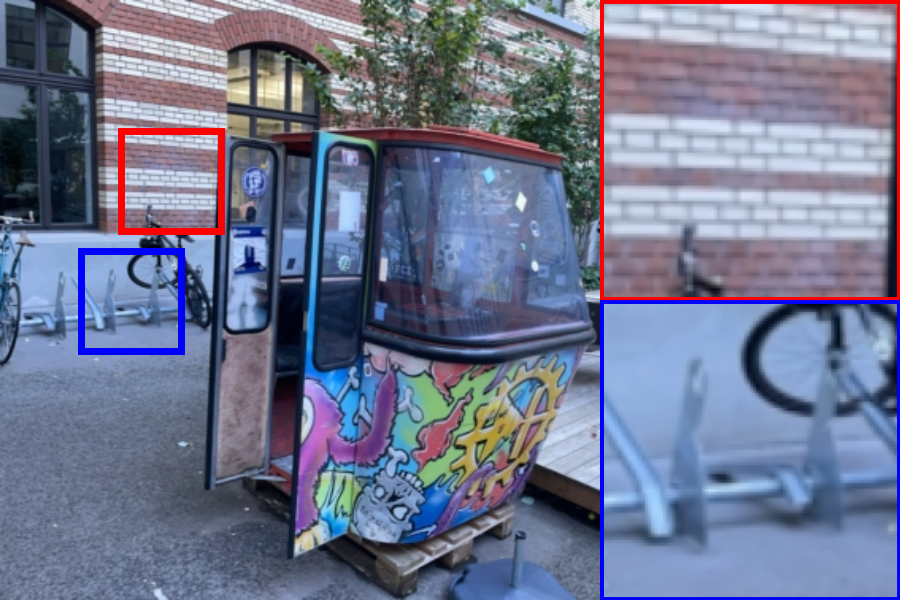}}
    {\includegraphics[width=\linewidth]{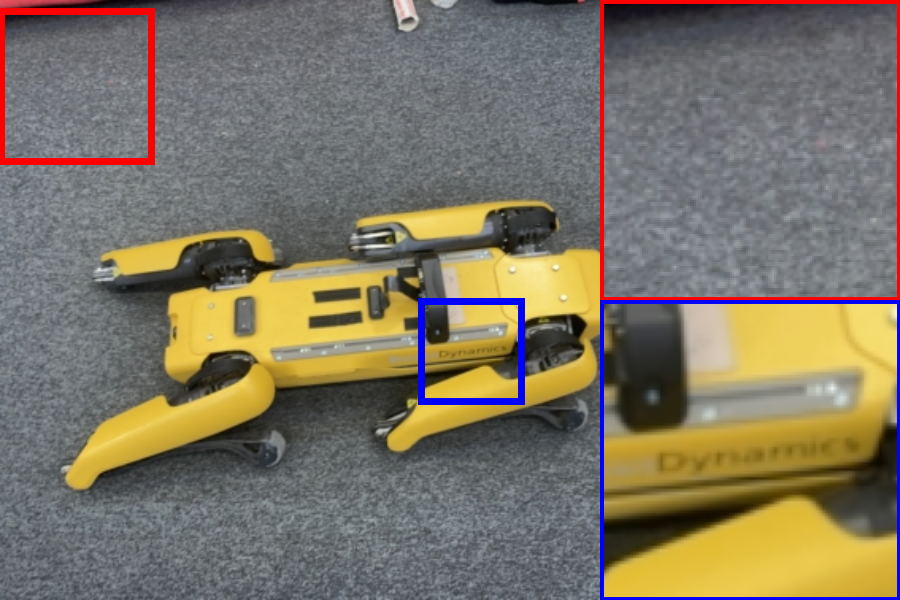}}
    {\includegraphics[width=\linewidth]{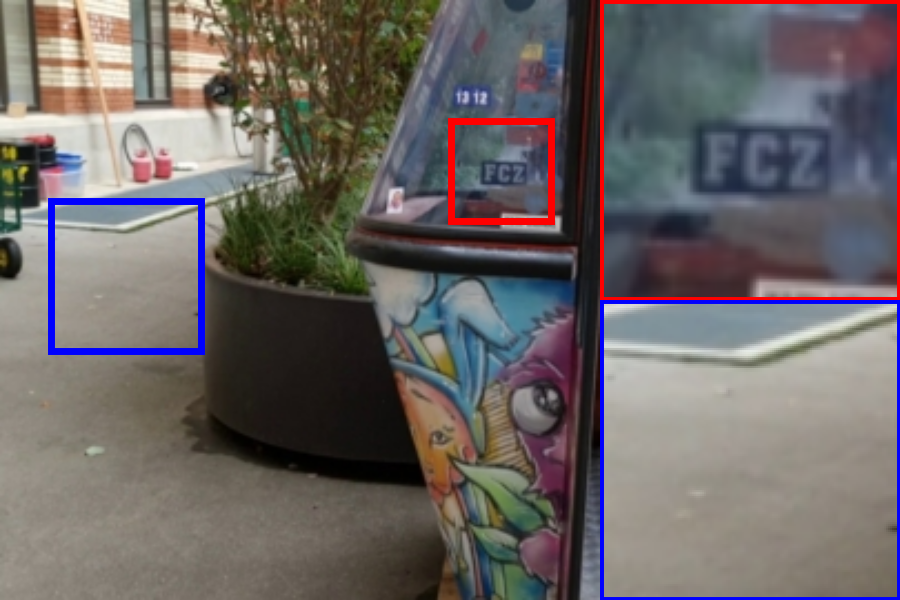}}
    {\includegraphics[width=\linewidth]{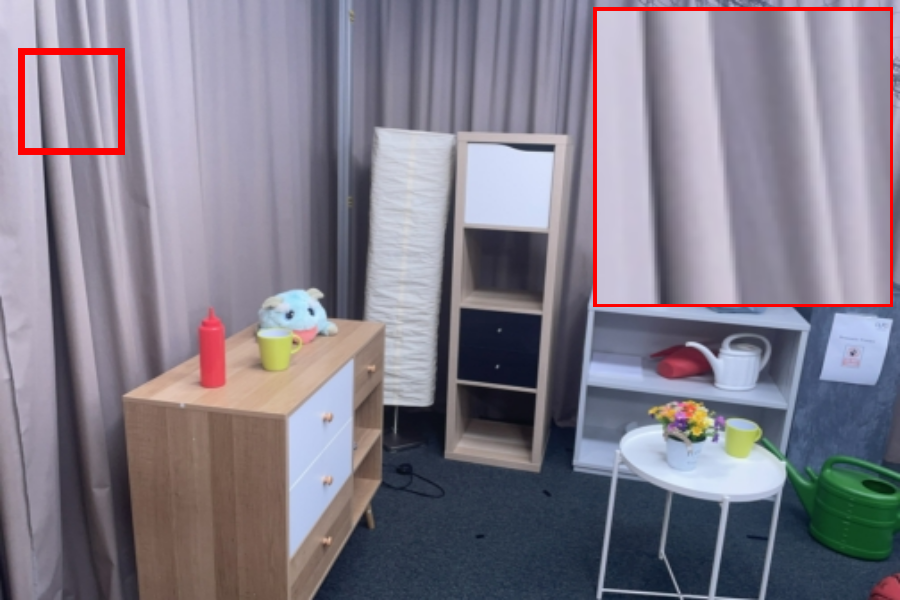}}
    {\includegraphics[width=\linewidth]{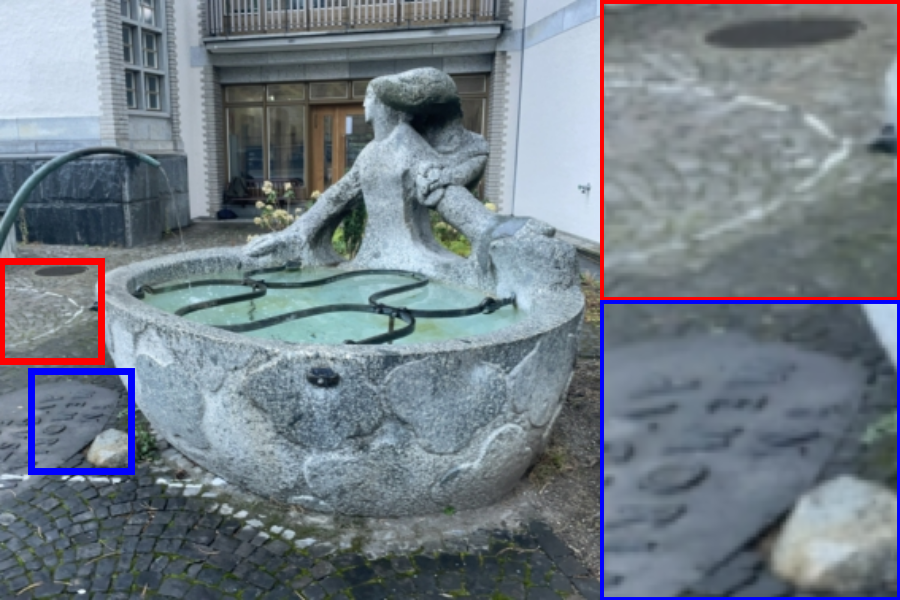}}
    {\includegraphics[width=\linewidth]{images/IMG_7940_gt_zoomed.png}}
    \centerline{Ground Truth}
\end{minipage}
\hfill
\vspace*{-5mm}
\caption{Qualitative results on the NeRF On-the-go dataset~\cite{ren_nerfonthego_2024}. The scenes shown are, from top to bottom: Patio-high (high occlusion), Spot (high occlusion), Patio (medium occlusion), Corner (medium occlusion), Mountain (low occlusion), and Fountain (low occlusion).}
\label{fig:onthego_extra}
\end{figure}

\begin{figure}[t]
\centering
\begin{minipage}[b]{0.195\linewidth}
    \centering
    {\includegraphics[width=\linewidth]{images/statue_1extra006_mip_zoomed.png}}
    {\includegraphics[width=\linewidth]{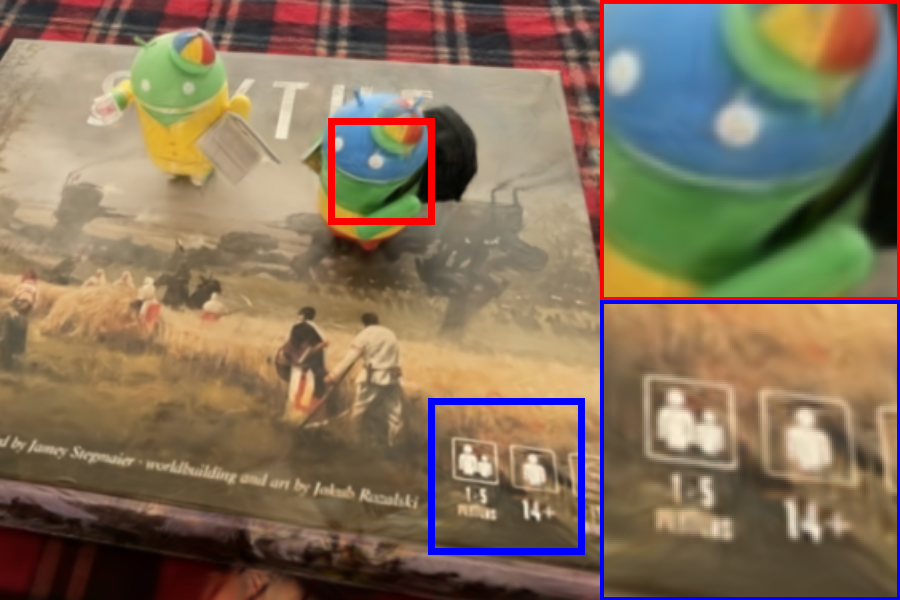}}
    {\includegraphics[width=\linewidth]{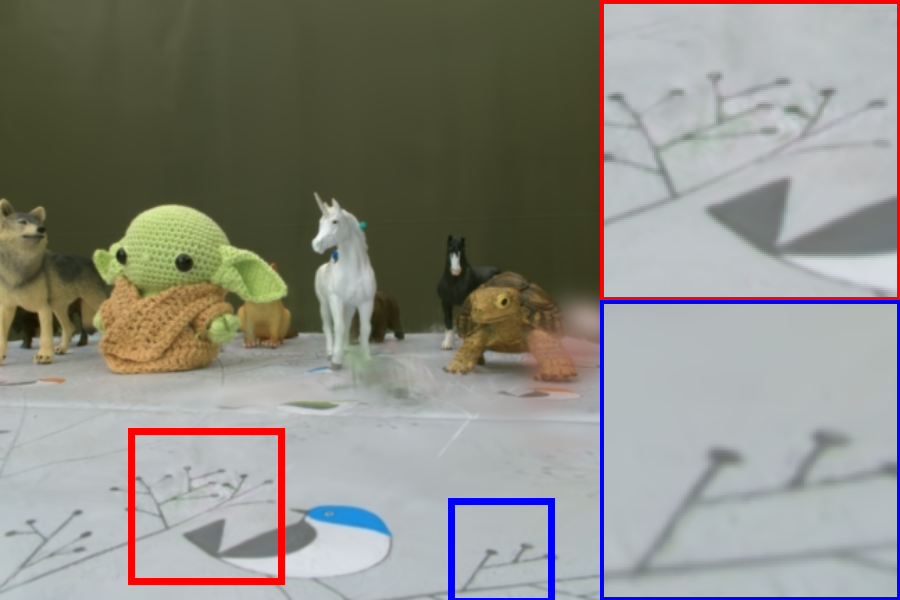}}
    {\includegraphics[width=\linewidth]{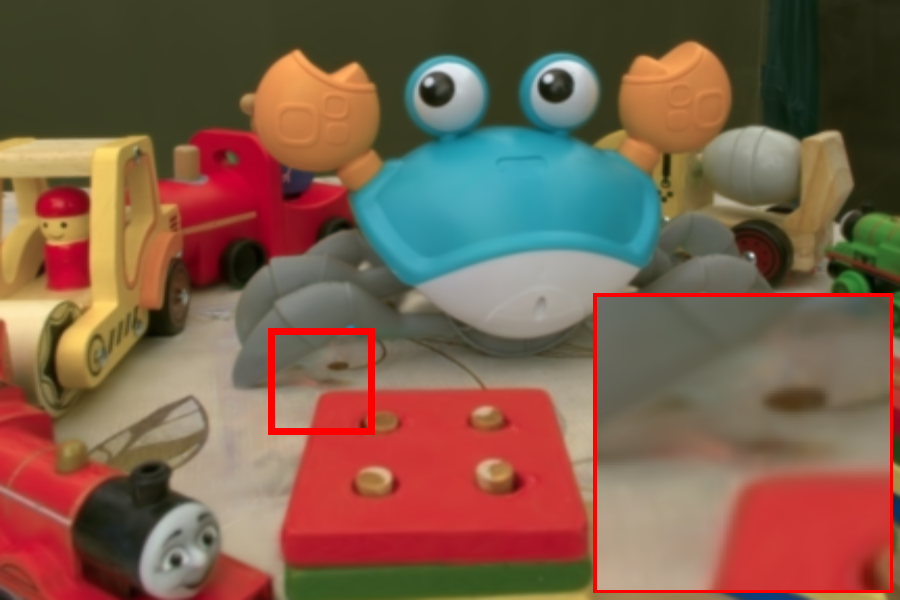}}
    \centerline{Mip-Splatting \cite{yu_mip-splatting_2024}}
\end{minipage}
\hfill
\begin{minipage}[b]{0.195\linewidth}
    \centering
    {\includegraphics[width=\linewidth]{images/statue_1extra006_hybridgs_zoomed.png}}
    {\includegraphics[width=\linewidth]{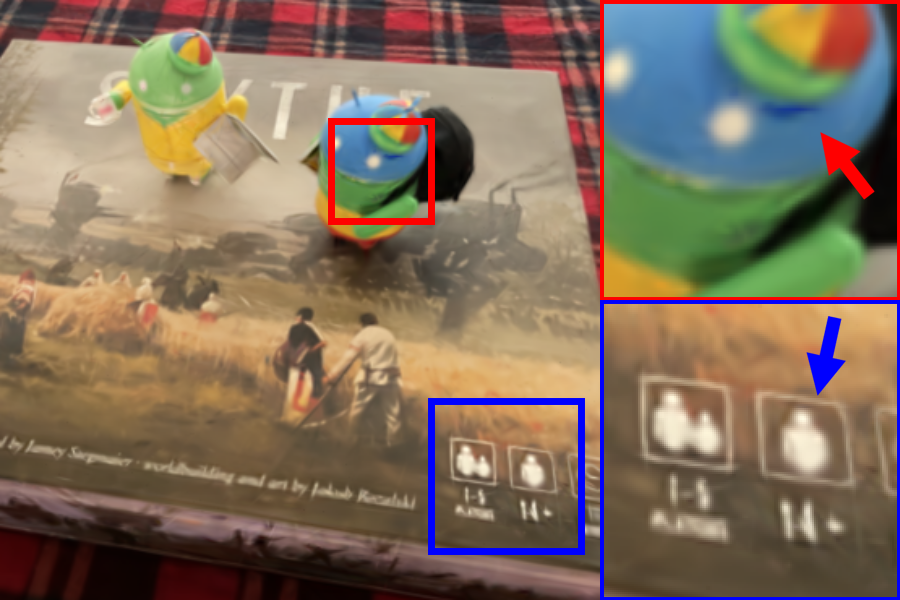}}
    {\includegraphics[width=\linewidth]{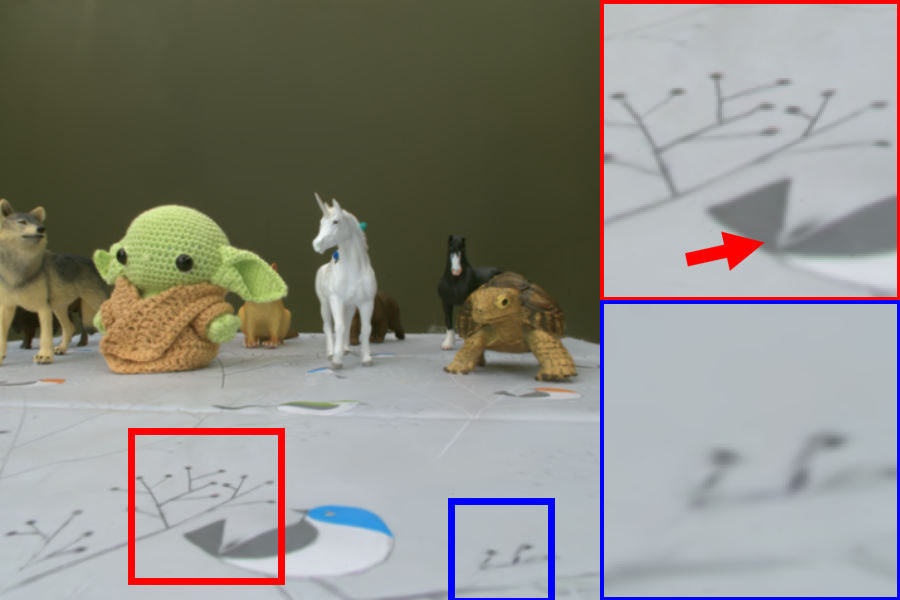}}
    {\includegraphics[width=\linewidth]{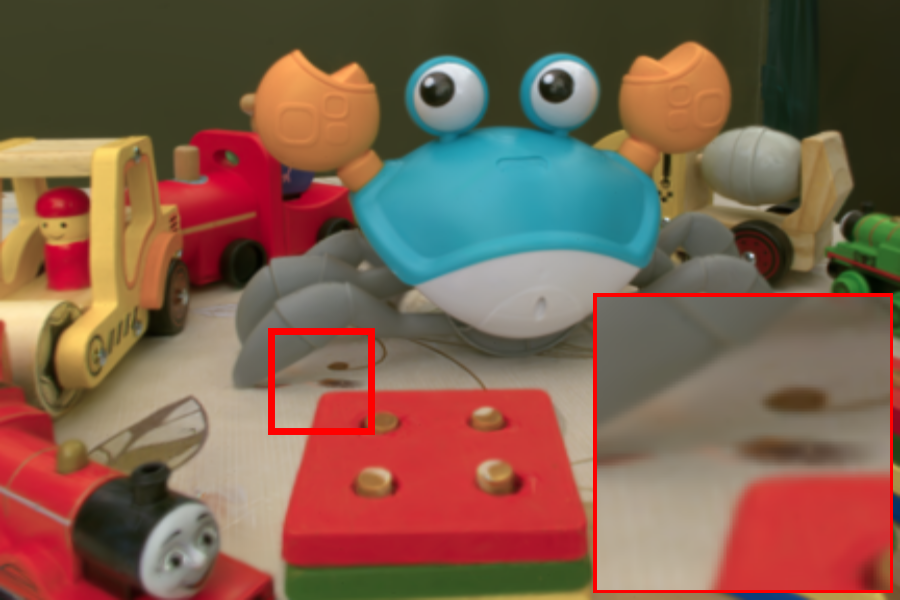}}
    \centerline{HybridGS \cite{lin_hybridgs_2025}}
\end{minipage}
\hfill
\begin{minipage}[b]{0.195\linewidth}
    \centering
    {\includegraphics[width=\linewidth]{images/statue_1extra006_ema_zoomed.png}}
    {\includegraphics[width=\linewidth]{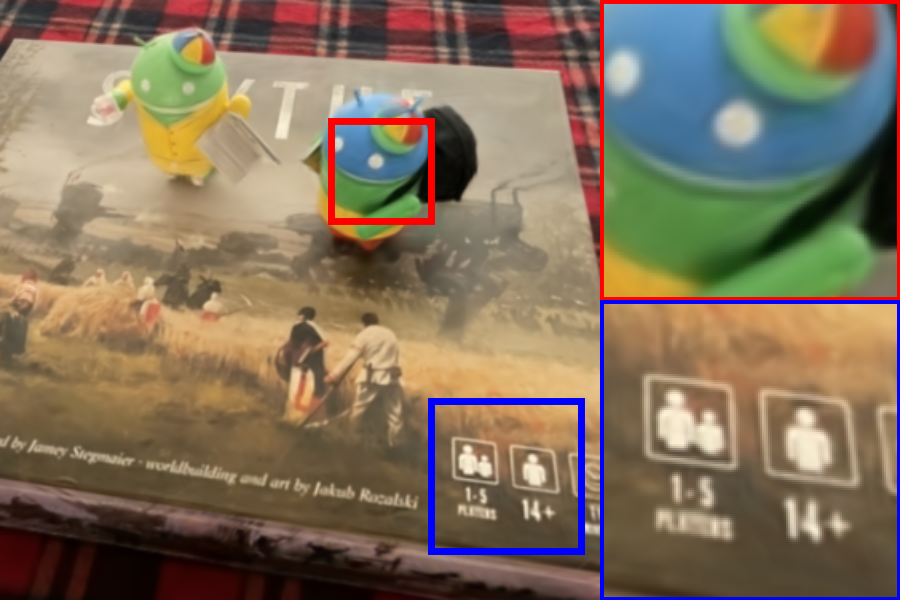}}
    {\includegraphics[width=\linewidth]{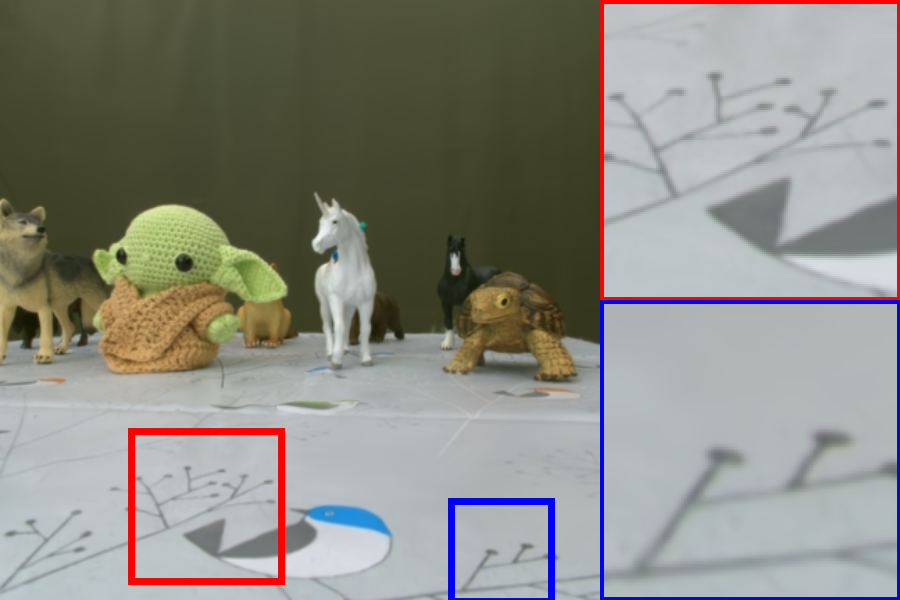}}
    {\includegraphics[width=\linewidth]{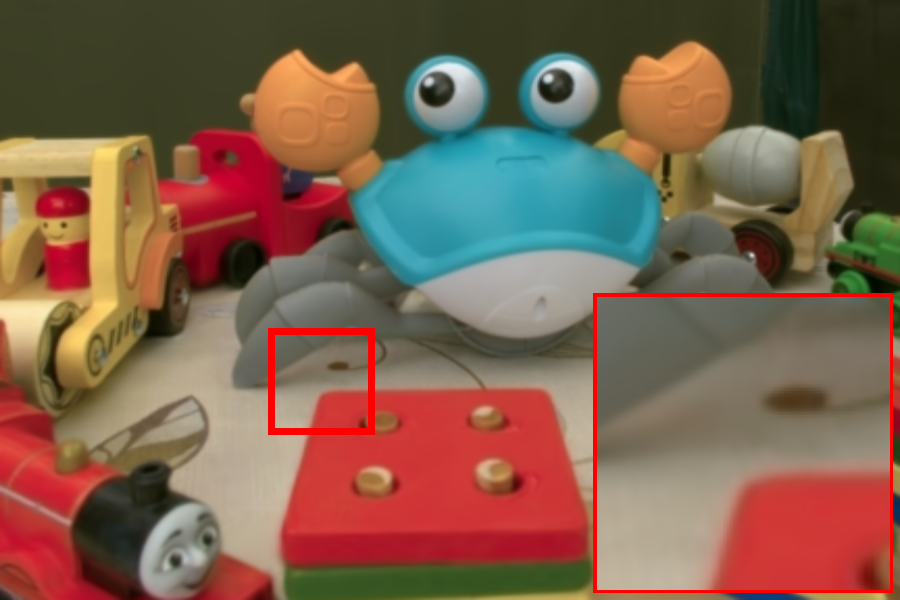}}
    \centerline{Our (EMA-GS)}
\end{minipage}
\hfill
\begin{minipage}[b]{0.195\linewidth}
    \centering
    {\includegraphics[width=\linewidth]{images/statue_1extra006_dual_zoomed.png}}
    {\includegraphics[width=\linewidth]{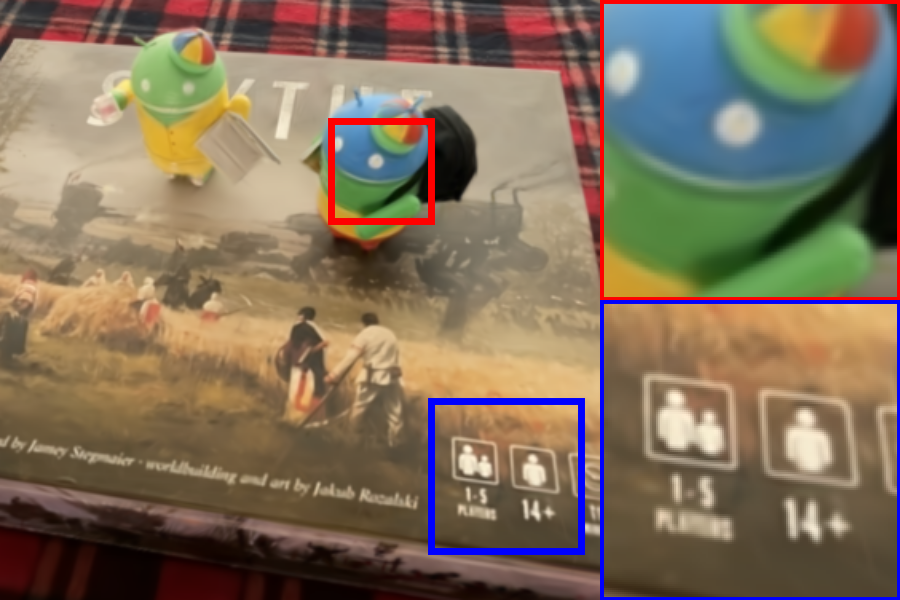}}
    {\includegraphics[width=\linewidth]{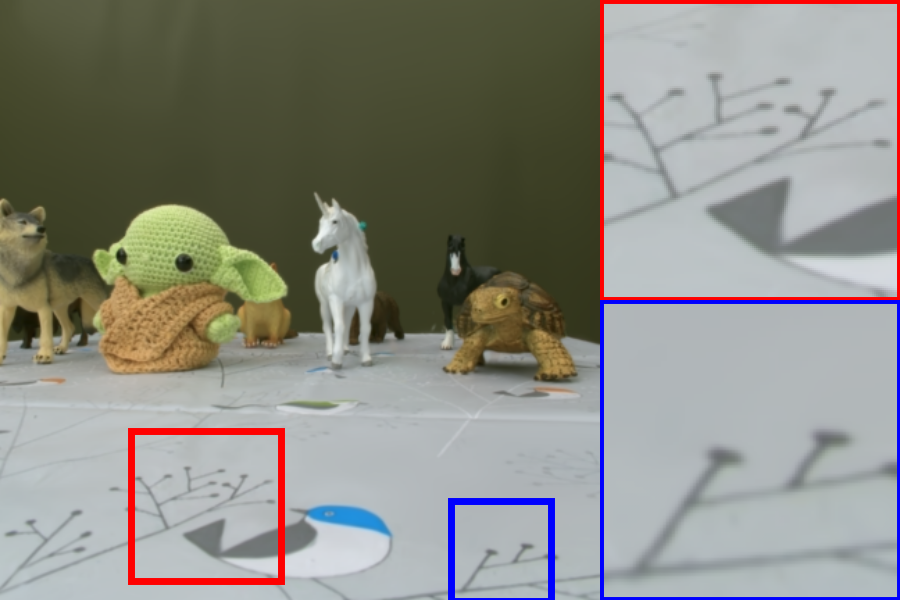}}
    {\includegraphics[width=\linewidth]{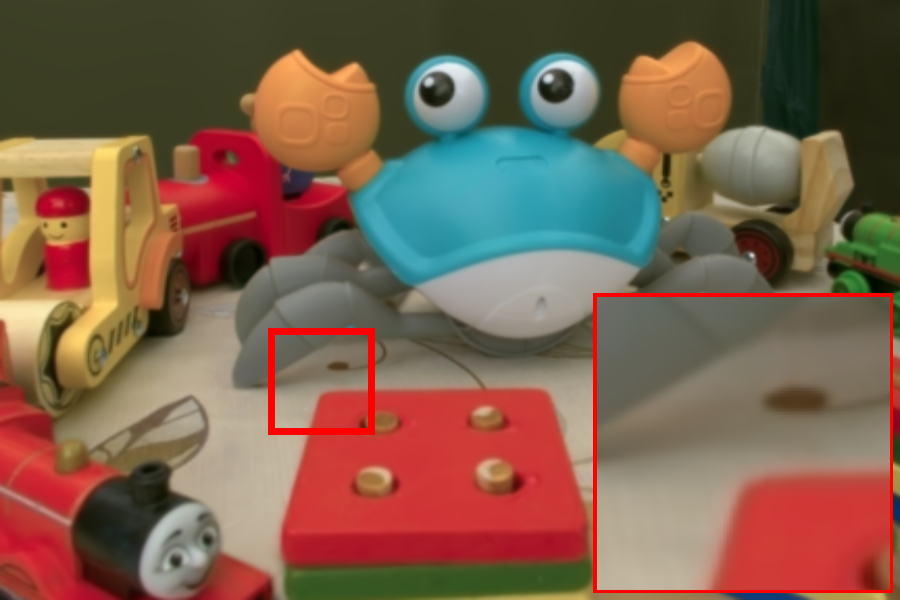}}
    \centerline{Ours (GS-GS)}
\end{minipage}
\hfill
\begin{minipage}[b]{0.195\linewidth}
    \centering
    {\includegraphics[width=\linewidth]{images/statue_1extra006_gt_zoomed.png}}
    {\includegraphics[width=\linewidth]{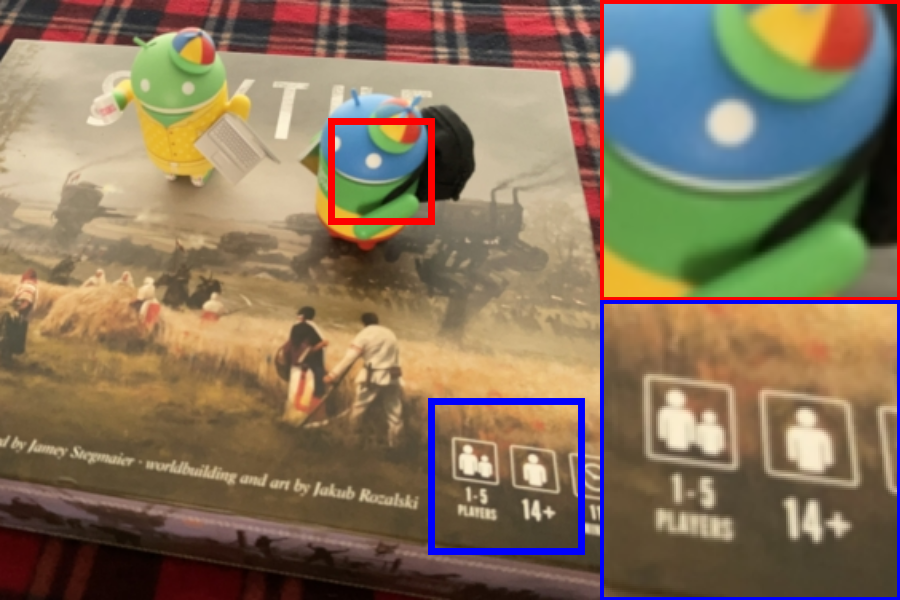}}
    {\includegraphics[width=\linewidth]{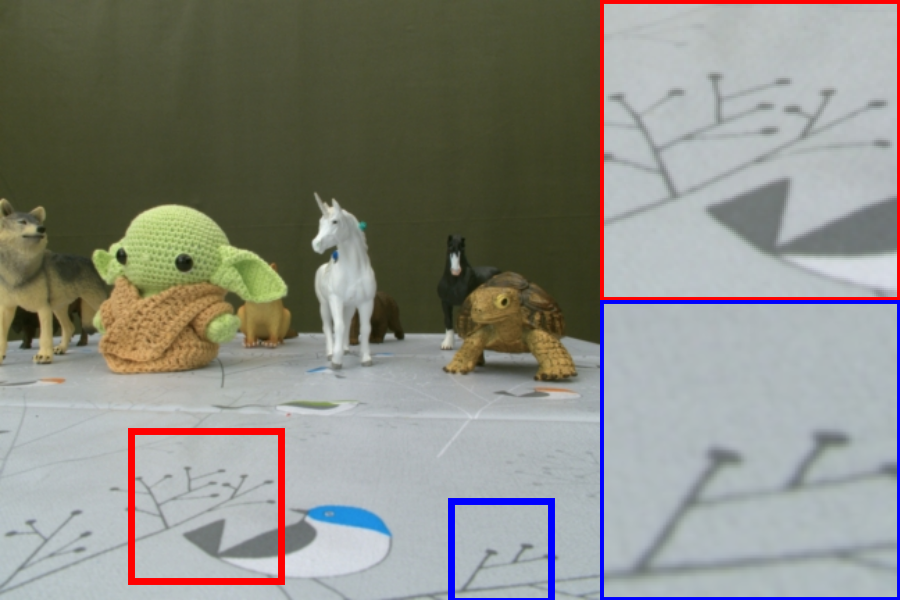}}
    {\includegraphics[width=\linewidth]{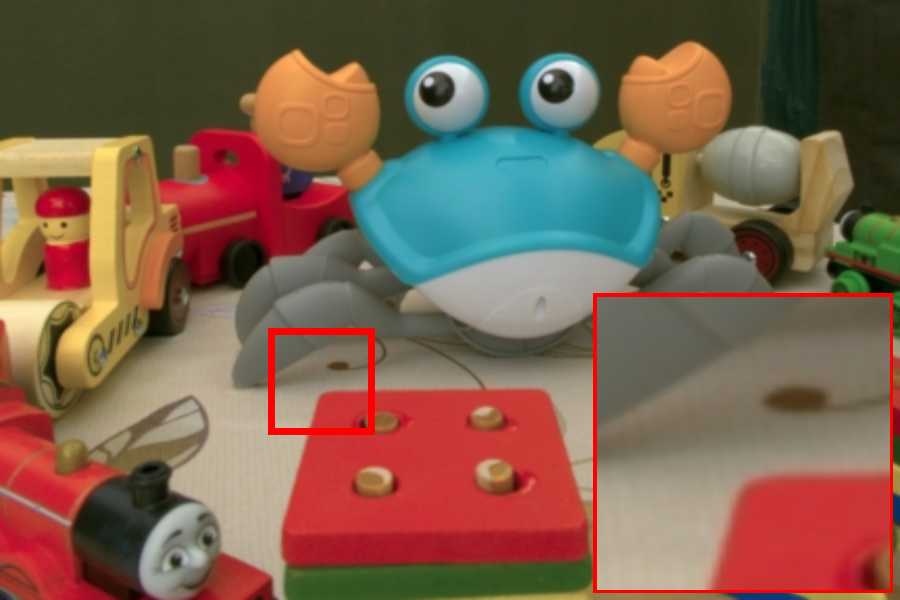}}
    \centerline{Ground Truth}
\end{minipage}
\hfill
\vspace*{-5mm}
\caption{Qualitative results on the RobustNeRF dataset~\cite{sabour_robustnerf_2023}. The scenes shown are, from top to bottom: Statue, Android, Yoda, and Crab.}
\label{fig:robustnerf_extra}
\end{figure}

In Table~\ref{tab:onthego} and Table \ref{tab:onthego_full}, our method (GS-GS) outperforms all baseline methods by more than 1 dB in scenes with medium to high occlusion ratios. The margin is smaller in low-occlusion scenes, where 3DGS-based methods already perform well due to strong geometric priors from the initial point cloud. A similar trend is observed in Table~\ref{tab:robustnerf}: while the proposed method surpasses the SOTA by approximately 0.4 dB in simpler scenes containing a single distractor type (e.g., Statue and Android), it outperforms others by more than 1 dB in complex scenes with a large number of diverse distractors (e.g., Yoda and Crab). The rendering results in Figure~\ref{fig:onthego_extra} and~\ref{fig:robustnerf_extra} further demonstrate the superiority of our method, as competing approaches exhibit distractor remains and missing details.

\subsection{PhotoTourism}
\begin{figure}[t]
\centering
\begin{minipage}[b]{0.195\linewidth}
    \centering
    {\includegraphics[width=\linewidth]{images/25927611_9367586008_mip_zoomed.png}}
    {\includegraphics[width=\linewidth]{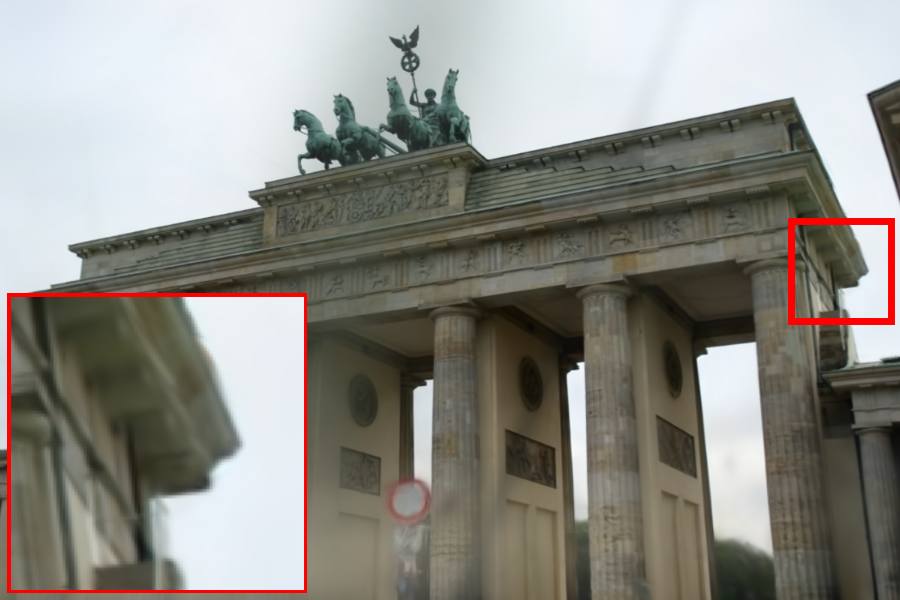}}
    {\includegraphics[width=\linewidth]{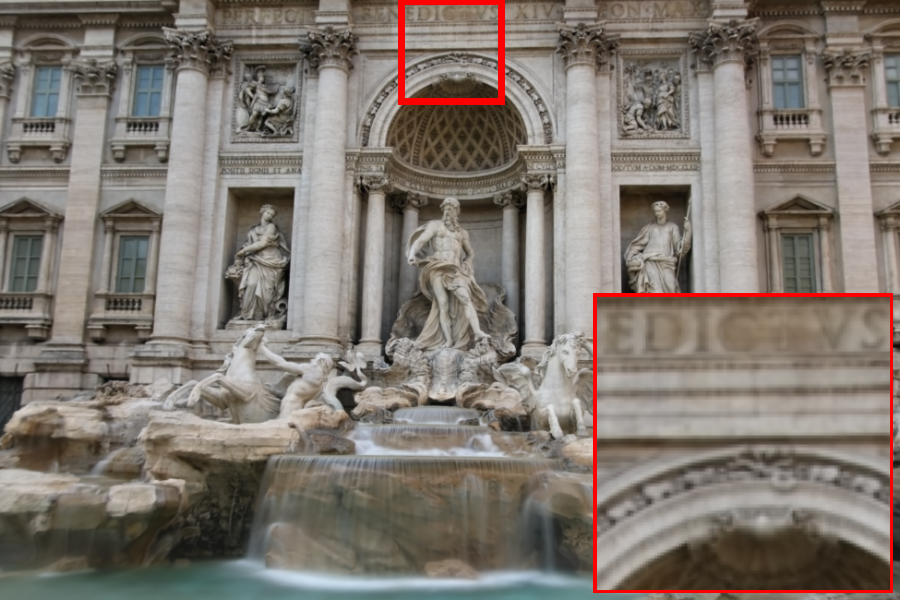}}
    \centerline{Mip-Splatting \cite{yu_mip-splatting_2024}}
\end{minipage}
\hfill
\begin{minipage}[b]{0.195\linewidth}
    \centering
    {\includegraphics[width=\linewidth]{images/25927611_9367586008_wild_zoomed.png}}
    {\includegraphics[width=\linewidth]{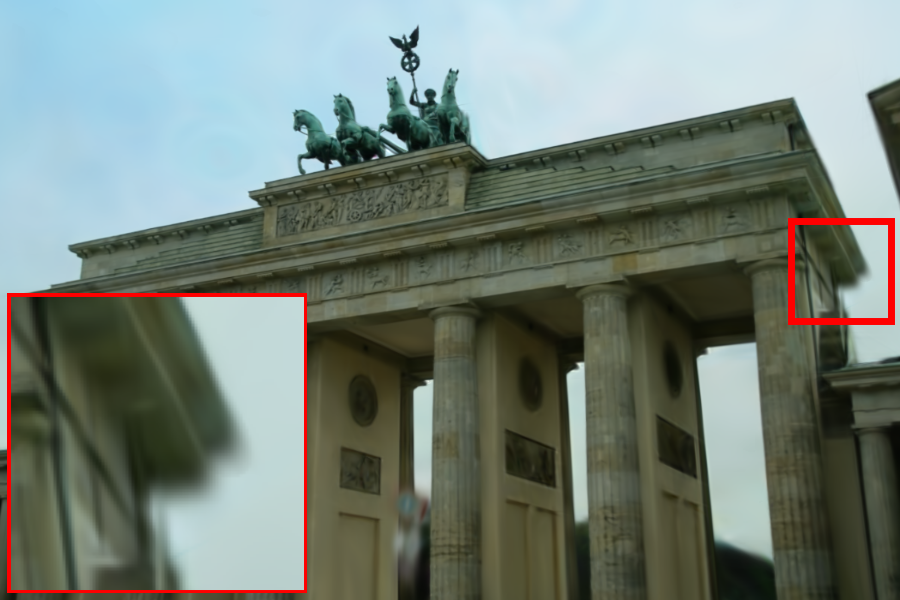}}
    {\includegraphics[width=\linewidth]{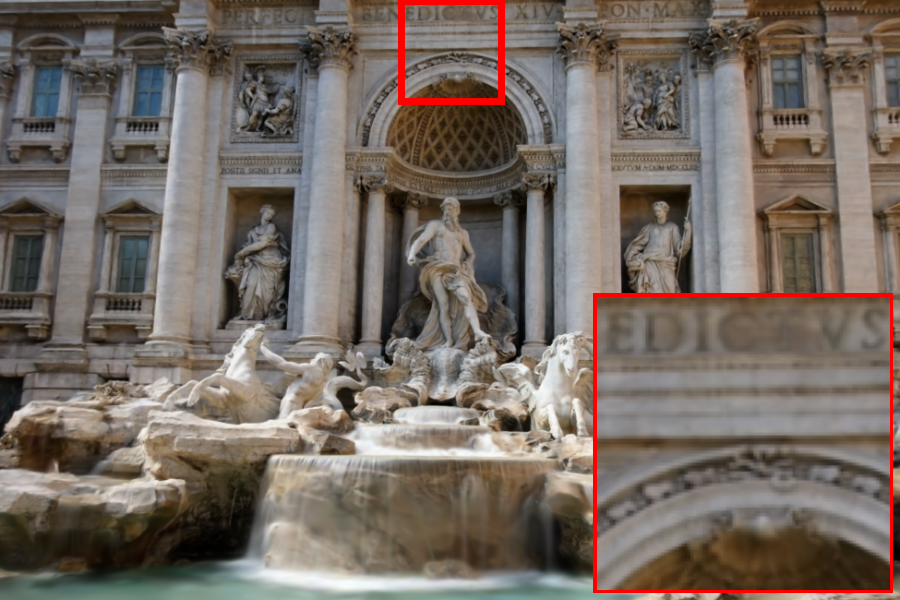}}
    \centerline{WildGaussian \cite{kulhanek_wildgaussians_2024}}
\end{minipage}
\hfill
\begin{minipage}[b]{0.195\linewidth}
    \centering
    {\includegraphics[width=\linewidth]{images/25927611_9367586008_ema_zoomed.png}}
    {\includegraphics[width=\linewidth]{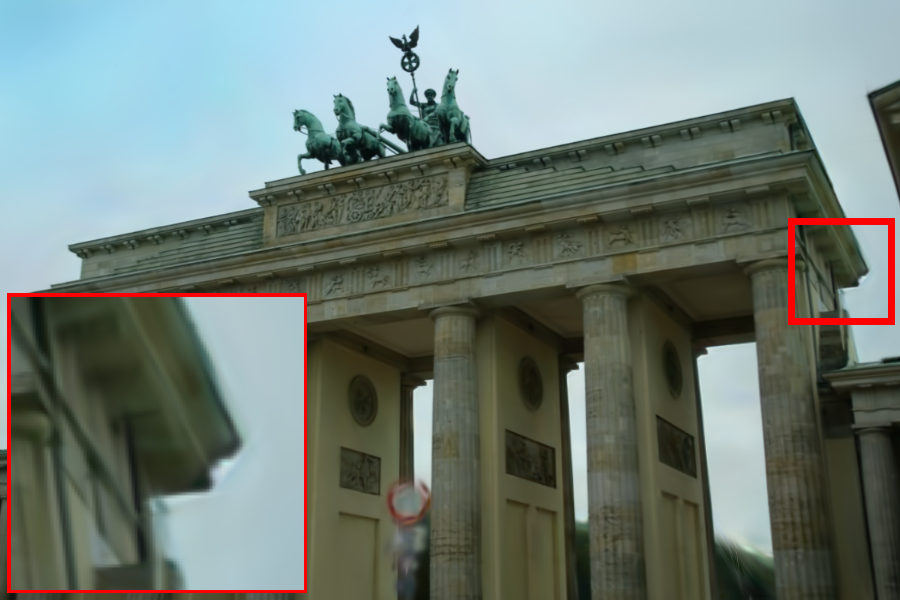}}
    {\includegraphics[width=\linewidth]{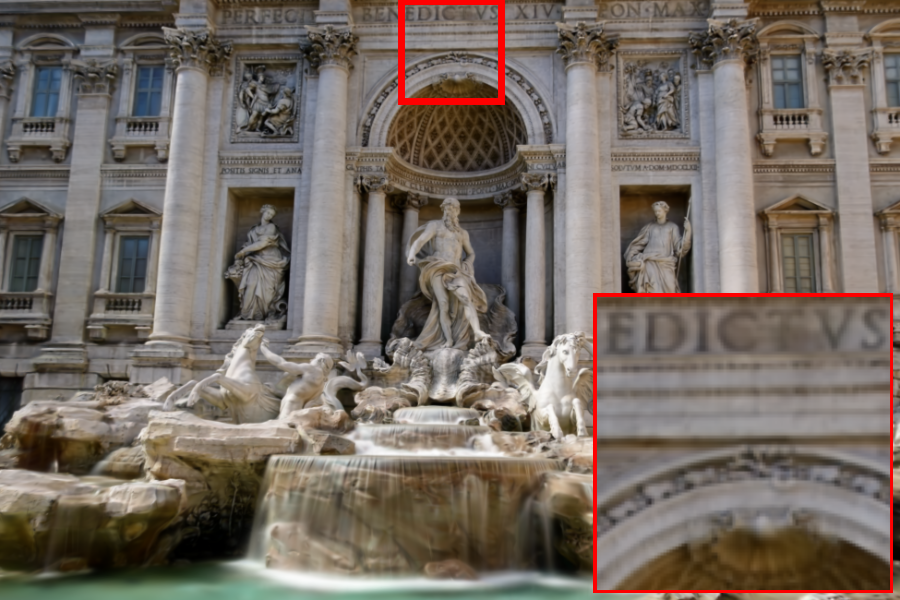}}
    \centerline{Our (EMA-GS)}
\end{minipage}
\hfill
\begin{minipage}[b]{0.195\linewidth}
    \centering
    {\includegraphics[width=\linewidth]{images/25927611_9367586008_dual_zoomed.png}}
    {\includegraphics[width=\linewidth]{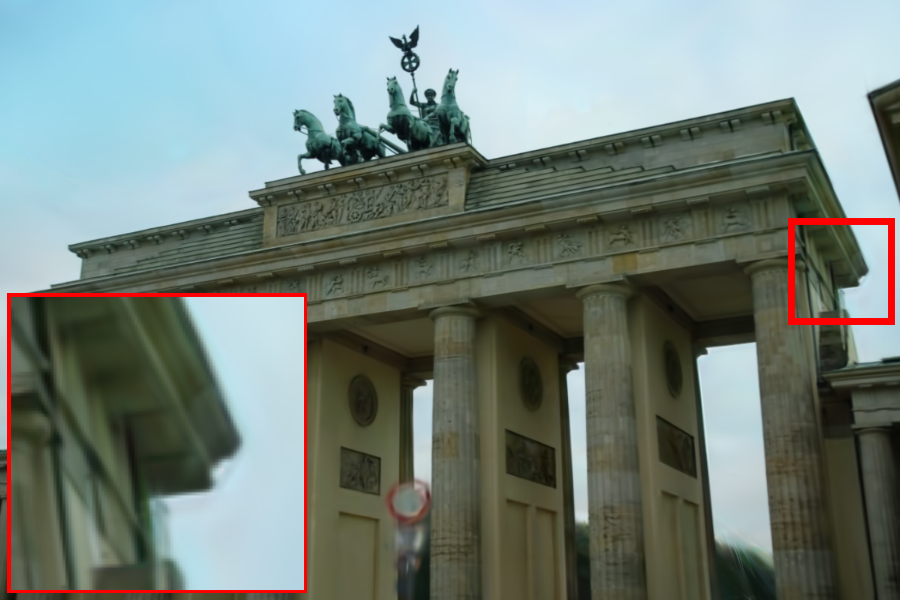}}
    {\includegraphics[width=\linewidth]{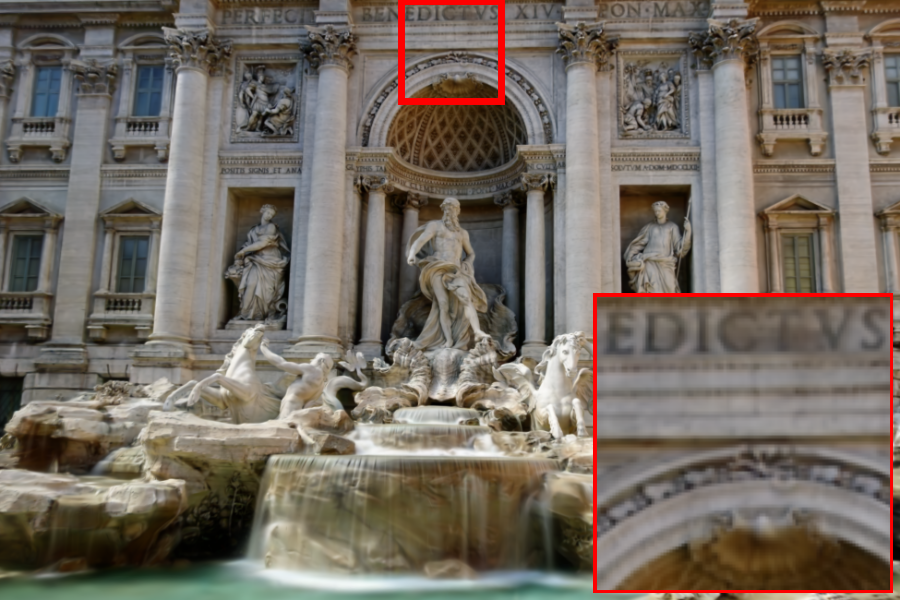}}
    \centerline{Ours (GS-GS)}
\end{minipage}
\hfill
\begin{minipage}[b]{0.195\linewidth}
    \centering
    {\includegraphics[width=\linewidth]{images/25927611_9367586008_gt_zoomed.png}}
    {\includegraphics[width=\linewidth]{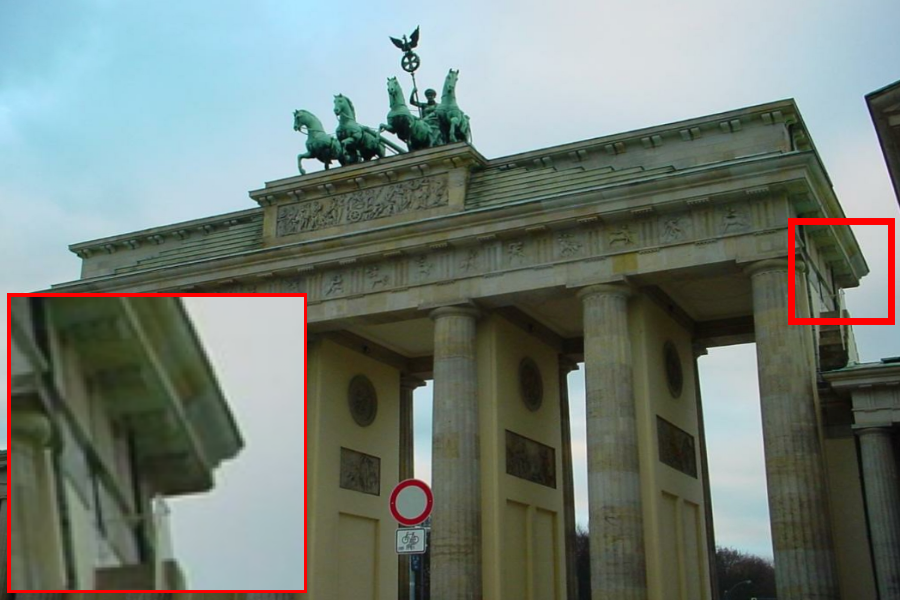}}
    {\includegraphics[width=\linewidth]{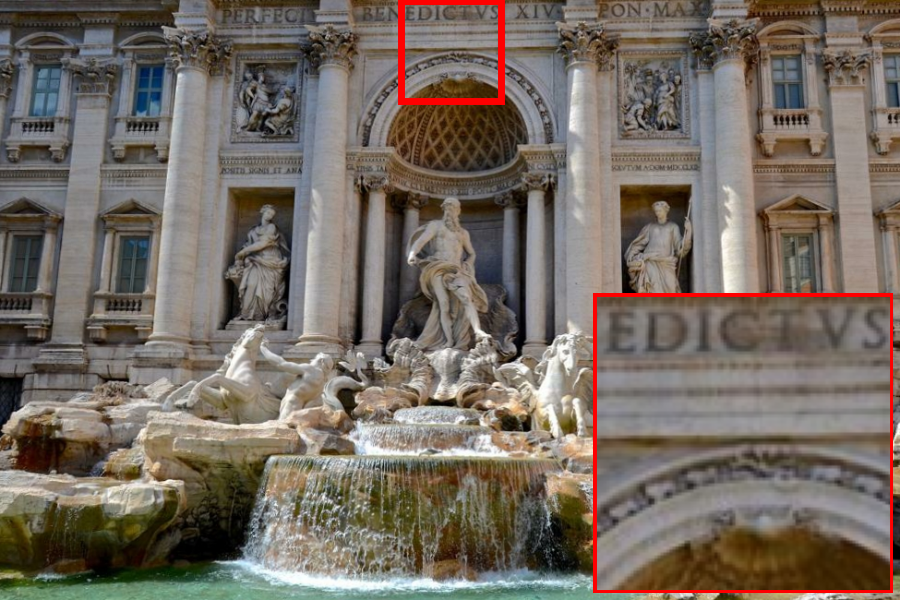}}
    \centerline{Ground Truth}
\end{minipage}
\hfill
\vspace*{-3mm}
\caption{Qualitative results on the PhotoTourism dataset~\cite{jin_phototourism_2020}. The scenes shown are, from top to bottom: Sacre Coeur, Brandenburg Gate, and Trevi Fountain.}
\vspace{-2mm}
\label{fig:phototourism_extra}
\end{figure}

The Asymmetric Dual 3DGS achieves an average improvement of 0.8 dB on the PhotoTourism dataset (Table~\ref{tab:phototourism}), demonstrating its effectiveness under challenging appearance variations. 
Furthermore, proper appearance modeling is essential for handling in-the-wild data with diverse visual conditions. This is supported by a significant performance gap of more than 4 dB between methods with and without appearance modeling, as shown in Table~\ref{tab:phototourism}, and further illustrated by the visual differences in Figure~\ref{fig:phototourism_extra}. Therefore, we apply appearance modeling for the PhotoTourism dataset by default. 
As the importance of appearance modeling is addressed here, we omit further discussion in the following ablation section and apply appearance modeling by default for the PhotoTourism dataset.

\subsection{Statistical significance of the main result}
\begin{table}[h!]
\caption{Quantitative results on the NeRF On-the-go dataset. Each experiment is repeated five times, and we report the mean and standard deviation.}
\label{tab:onthego_multirun}
\resizebox{\linewidth}{!}{
\begin{tabular}{lllllll}
\toprule
Setting & \multicolumn{3}{c}{GS-GS} & \multicolumn{3}{c}{EMA-GS} \\
\midrule
Scene   & PSNR    & SSIM   & LPIPS  & PSNR    & SSIM    & LPIPS  \\
\midrule
High Occlusion   & $24.36 \pm 0.02$   & $0.823 \pm 0.001$  & $0.151 \pm 0.001$  & $24.11 \pm 0.05$   & $0.819 \pm 0.002$   & $0.152 \pm 0.004$  \\
Medium Occlusion & $24.52 \pm 0.06$   & $0.871 \pm 0.001$  & $0.090 \pm 0.001$  & $24.26 \pm 0.08$   & $0.864 \pm 0.001$   & $0.092 \pm 0.002$  \\
Low Occlusion    & $21.99 \pm 0.04$   & $0.730 \pm 0.001$  & $0.184 \pm 0.004$  & $21.81 \pm 0.09$   & $0.723 \pm 0.002$   & $0.166 \pm 0.007$  \\
\bottomrule
\end{tabular}
}
\end{table}

\begin{table}[h!]
\caption{Quantitative results on the RobustNeRF dataset. Each experiment is repeated five times, and we report the mean and standard deviation.}
\label{tab:robustnerf_multirun}
\resizebox{\linewidth}{!}{
\begin{tabular}{lllllll}
\toprule
Setting & \multicolumn{3}{c}{GS-GS} & \multicolumn{3}{c}{EMA-GS} \\
\midrule
Scene   & PSNR    & SSIM   & LPIPS  & PSNR    & SSIM    & LPIPS  \\
\midrule
Statue  & $23.44 \pm 0.05$   & $0.893 \pm 0.001$  & $0.098 \pm 0.001$  & $23.46 \pm 0.06$   & $0.890 \pm 0.001$   & $0.097 \pm 0.001$  \\
Android & $25.58 \pm 0.05$   & $0.856 \pm 0.001$  & $0.070 \pm 0.003$  & $25.47 \pm 0.06$   & $0.849 \pm 0.002$   & $0.070 \pm 0.002$  \\
Yoda    & $37.12 \pm 0.09$   & $0.969 \pm 0.001$  & $0.074 \pm 0.001$  & $36.46 \pm 0.06$   & $0.967 \pm 0.001$   & $0.078 \pm 0.001$  \\
Crab    & $36.11 \pm 0.07$   & $0.963 \pm 0.001$  & $0.079 \pm 0.001$  & $35.52 \pm 0.07$   & $0.961 \pm 0.001$   & $0.080 \pm 0.001$ \\
\bottomrule
\end{tabular}
}
\end{table}

We repeated the experiment five times. Based on the results in Table \ref{tab:onthego_multirun} and \ref{tab:robustnerf_multirun}, our method shows statistically significant improvements.

\subsection{Hyperparameters} 
We perform hyperparameter tuning on the NeRF On-the-go dataset~\cite{ren_nerfonthego_2024} to optimize the performance of our method (GS-GS and EMA-GS). As shown in Table~\ref{tab:beta_tuning}, we tune the EMA smoothing factor $\beta$ and find that $\beta = 0.8$ yields the highest PSNR and SSIM with the lowest LPIPS. In Table~\ref{tab:densification_tuning}, we evaluate different densification intervals and observe that an interval of 1000 offers the best overall performance. Similarly, Table~\ref{tab:warmup_tuning} presents the results of tuning the warm-up interval, where 1000 again emerges as the optimal choice, outperforming both shorter and longer intervals. Lastly, Table~\ref{tab:opacity_reset_tuning} shows that removing opacity reset improves reconstruction quality, suggesting that preserving learned opacity leads to more stable and effective training.

\begin{table}[ht]
\caption{Tuning the EMA smoothing factor according to the average performance on the NeRF On-the-go dataset~\cite{ren_nerfonthego_2024}.}
\label{tab:beta_tuning}
\centering
\begin{tabular}{llll}
\toprule
$\beta$ & PSNR$\uparrow$  & SSIM$\uparrow$  & LPIPS$\downarrow$ \\
\midrule
0.5 & 22.80 & 0.797 & 0.136 \\
0.6 & 22.93 & 0.797 & 0.137 \\
0.7 & 23.12 & 0.799 & 0.136 \\
0.8 & 23.40 & 0.801 & 0.135 \\
0.9 & 23.05 & 0.798 & 0.136 \\
\bottomrule
\end{tabular}
\end{table}

\begin{table}[ht]
\caption{Tuning the densification interval according to the average performance on the NeRF On-the-go dataset~\cite{ren_nerfonthego_2024}.}
\label{tab:densification_tuning}
\centering
\begin{tabular}{lllllll}
\toprule
Setting & \multicolumn{3}{c}{GS-GS} & \multicolumn{3}{c}{EMA-GS} \\
\midrule
Densification Interval & PSNR$\uparrow$  & SSIM$\uparrow$  & LPIPS$\downarrow$ & PSNR$\uparrow$  & SSIM$\uparrow$  & LPIPS$\downarrow$ \\
\midrule
500  & 23.60 & 0.810 & 0.129 & 23.00 & 0.797 & 0.134 \\
1000 & 23.61 & 0.810 & 0.135 & 23.40 & 0.801 & 0.135 \\
1500 & 23.58 & 0.807 & 0.146 & 23.15 & 0.796 & 0.143 \\
2000 & 23.56 & 0.806 & 0.152 & 22.96 & 0.797 & 0.145 \\
\bottomrule
\end{tabular}
\end{table}

\begin{table}[ht]
\caption{Tuning the warm-up interval according to the average performance on the NeRF On-the-go dataset~\cite{ren_nerfonthego_2024}.}
\label{tab:warmup_tuning}
\centering
\begin{tabular}{lllllll}
\toprule
Setting & \multicolumn{3}{c}{GS-GS} & \multicolumn{3}{c}{EMA-GS} \\
\midrule
Warm-up Interval & PSNR$\uparrow$  & SSIM$\uparrow$  & LPIPS$\downarrow$ & PSNR$\uparrow$  & SSIM$\uparrow$  & LPIPS$\downarrow$ \\
\midrule
0    & 23.55 & 0.808 & 0.137 & 22.96 & 0.798 & 0.135 \\
500  & 23.55 & 0.809 & 0.137 & 23.08 & 0.799 & 0.134 \\
1000 & 23.61 & 0.810 & 0.135 & 23.40 & 0.801 & 0.135 \\
1500 & 23.58 & 0.809 & 0.137 & 23.10 & 0.799 & 0.135 \\
2000 & 23.60 & 0.810 & 0.135 & 22.88 & 0.798 & 0.136 \\
\bottomrule
\end{tabular}
\end{table}

\begin{table}[ht]
\caption{Impact of opacity reset on reconstruction quality, evaluated on the NeRF On-the-go dataset~\cite{ren_nerfonthego_2024}.}
\label{tab:opacity_reset_tuning}
\centering
\begin{tabular}{lllllll}
\toprule
Setting & \multicolumn{3}{c}{GS-GS} & \multicolumn{3}{c}{EMA-GS} \\
\midrule
Opacity Reset & PSNR$\uparrow$  & SSIM$\uparrow$  & LPIPS$\downarrow$ & PSNR$\uparrow$  & SSIM$\uparrow$  & LPIPS$\downarrow$ \\
\midrule
w/o  & 23.61 & 0.810 & 0.135 & 23.40 & 0.801 & 0.135 \\
w/ & 22.87 & 0.790 & 0.176 & 22.43 & 0.786 & 0.158 \\
\bottomrule
\end{tabular}
\end{table}

\begin{table}[h!]
\caption{Performance under varying weights of the mutual consistency loss, evaluated on the NeRF On-the-go dataset~\cite{ren_nerfonthego_2024}.}
\label{tab:mul_loss_weight}
\centering
\begin{tabular}{lllllllll}
\toprule
  & \multicolumn{3}{c}{GS-GS} & & \multicolumn{3}{c}{EMA-GS} \\
\midrule
$\lambda_{m}$   & PSNR    & SSIM   & LPIPS  & $\lambda_{m}$   & PSNR    & SSIM    & LPIPS  \\
\midrule
0.0 & 23.13   & 0.808  & 0.135  & 0.0 & 23.10   & 0.801  & 0.132  \\
0.5 & 23.66   & 0.810  & 0.130  & 0.05 & 23.39   & 0.803  & 0.136  \\
1.0 & 23.61   & 0.810  & 0.135  & 0.1 & 23.40   & 0.801  & 0.135  \\
1.5 & 23.54   & 0.807  & 0.142  & 0.2 & 23.43   & 0.805  & 0.133  \\
2.0 & 23.44   & 0.803  & 0.149  & 0.3 & 23.47   & 0.805  & 0.134  \\
\bottomrule
\end{tabular}
\end{table}

\begin{table}[h!]
\caption{Performance under varying weights of the learnable mask loss, evaluated on the NeRF On-the-go dataset~\cite{ren_nerfonthego_2024}.}
\label{tab:mask_loss_weight}
\centering
\begin{tabular}{lllllll}
\toprule
Setting & \multicolumn{3}{c}{GS-GS} & \multicolumn{3}{c}{EMA-GS} \\
\midrule
$\lambda_{mask}$   & PSNR    & SSIM   & LPIPS   & PSNR    & SSIM    & LPIPS  \\
\midrule
0.5 & 23.62   & 0.809  & 0.136  & 23.33   & 0.802  & 0.134  \\
1.0 & 23.61   & 0.810  & 0.135  & 23.40   & 0.801  & 0.135  \\
1.5 & 23.59   & 0.809  & 0.137  & 23.41   & 0.804  & 0.135  \\
2.0 & 23.63   & 0.811  & 0.135  & 23.32   & 0.802  & 0.134  \\
\bottomrule
\end{tabular}
\end{table}

In Table \ref{tab:mul_loss_weight} and \ref{tab:mask_loss_weight}, although the best performance is generally achieved at our default setting ($\lambda_m = 1.0$ and $\lambda_{mask} = 1.0$ for GS-GS; $\lambda_m = 0.1$ and $\lambda_{mask} = 1.0$ for EMA-GS), the differences across settings are minimal (less than 0.1 dB). This indicates that the performance is not highly sensitive to the values of $\lambda_m$ and $\lambda_{mask}$.

\section{Limitations}
We adopt the appearance modeling approach from WildGaussian \cite{kulhanek_wildgaussians_2024}, using a per-view appearance embedding to control global appearance and a per-Gaussian embedding to model the appearance of individual Gaussian primitives. However, this model struggles to capture fine-grained effects such as object highlights. A likely reason is the limited diversity in training data. To address this, we plan to introduce data augmentation with randomized illumination variations. 

\section{Social impact}
Notre-Dame de Paris suffered a devastating fire in 2019. Although the building was severely damaged, restoration was aided by a 3D model originally created for a video game, highlighting the importance of preserving 3D models of cultural landmarks. However, such sites are often crowded with people, and photos taken at different times may exhibit varying lighting conditions. This highlights the broader societal benefit of accessible and robust 3D scene reconstruction technologies. Our method contributes positively by enabling the creation of high-quality 3D models from in-the-wild images, which are often affected by distractors and lighting variations. By making it feasible to reconstruct cultural landmarks from everyday photos, our approach supports digital preservation, education, and historical restoration efforts.

There are potential negative impacts, such as misuse in surveillance or privacy-invading applications. In particular, in-the-wild image collections often contain individuals who are unintentionally captured. To mitigate this risk, we recommend removing or anonymizing identifiable information, such as faces or bodies, from the reconstructed scenes. This can be achieved through automated segmentation or masking techniques applied before or during training.

\end{document}